\documentclass{elsarticle}
\usepackage{algorithm}
\usepackage{algpseudocode}
\usepackage{amsmath}
\usepackage{amssymb}
\usepackage{bm}
\usepackage{booktabs}
\usepackage{bigstrut}
\usepackage{color}
\usepackage{epsfig}
\usepackage{epstopdf}
\usepackage{flafter}
\usepackage{float}
\usepackage{ifthen}
\usepackage{multirow}
\usepackage{pifont}
\usepackage{subfigure}
\usepackage{times}
\usepackage{threeparttable}
\usepackage{url}  % Formatting web addresses  % Conditional
\usepackage{subfigure,graphicx}
\usepackage{float}
\usepackage{color,multirow}
\usepackage{url}  % Formatting web addresses
\usepackage{ifthen}  % Conditional
\usepackage[T1]{fontenc}
\usepackage{lineno}

\usepackage[left=1.25in,right=1.25in,top=1.25in,bottom=1.25in,letterpaper]{geometry}
\newtheorem{theorem}{\textbf{Theorem}}
\newtheorem{myDef}{\textbf{Definition}}

\renewcommand{\algorithmicensure}{\textbf{Output:}} % Use Output in the format

\modulolinenumbers[5]
\journal{Neurocomputing}

\begin{document}

\begin{frontmatter}

\title{Deep Plug-and-Play Prior for Low-Rank Tensor Completion\tnoteref{mytitlenote}}
\tnotetext[mytitlenote]{This work is supported by the National Natural Science Foundation of China (61876203, 61772003, and 11971374), HKRGC GRF (12306616, 12200317, 12300218, and 12300519), HKU Grant
(104005583), China Postdoctoral Science Foundation  (2017M610628 and 2018T111031), and the State Key Laboratory of Robotics (2019-O06).}

\author[uestc]{Xi-Le Zhao}
\ead{xlzhao122003@163.com}
\author[uestc]{Wen-Hao Xu}
\ead{seanxwh@gmail.com}
\author[swufe]{Tai-Xiang Jiang\corref{cor}}
\cortext[cor]{Corresponding author.}
\ead{taixiangjiang@gmail.com}
\author[xju,zky]{Yao Wang}
\ead{yao.s.wang@gmail.com}
\author[hku]{Michael K. Ng}
\ead{mng@maths.hku.hk}

\address[uestc]{School of Mathematical Sciences/Resrarch Center for Image and Vision Computing, University of Electronic Science and Technology of China, Chengdu, Sichuan, 611731, China}
\address[swufe]{FinTech Innovation Center, Financial Intelligence and Financial Engineering Research Key Laboratory of Sichuan province, School of Economic Information Engineering, Southwestern University of Finance and Economics, Chengdu, Sichuan, 611130, China}
\address[xju]{School of Management, Xi'an Jiaotong University, Xi'an 710049, China}
\address[zky]{State Key Laboratory of Robotics, Shenyang Institute of Automation, Chinese Academy of Sciences, Shenyang 110016, China}
\address[hku]{Department of Mathematics, The University of Hong Kong, Pokfulam, Hong Kong}

\begin{abstract}
Multi-dimensional images, such as color images and multi-spectral images, are highly correlated and contain abundant spatial and spectral information. However, real-world multi-dimensional images are usually corrupted by missing entries.   By integrating deterministic low-rankness prior to the data-driven deep  prior, we suggest  a novel regularized tensor completion model for multi-dimensional image completion. In the objective function, we adopt the newly emerged tensor nuclear norm (TNN) to characterize the global low-rankness prior of the multi-dimensional images. We also formulate an implicit regularizer by plugging into a denoising neural network (termed as deep denoiser), which is convinced to express the deep image prior learned from a large number of natural images. The resulting model can be  solved by the alternating directional method of multipliers algorithm under the plug-and-play (PnP) framework. Experimental results on color images, videos, and multi-spectral images demonstrate that the proposed method can recover both the global structure and fine details very well and achieve superior performance over competing methods in terms of quality metrics and visual effects.
\end{abstract}


\end{frontmatter}

%\linenumbers

\section{Introduction}
\label{sec:Int}
%Due to sensor malfunction and poor atmospheric conditions, there are usually missing values in multi-dimensional images.
The image completion problem aims to estimate missing entries from the partially observed entries, which is a fundamental problem in low-level computer vision and computational imaging \cite{zhuang2018fast,wang2018hyperspectral, chang2015anisotropic, yang2020remote, zhao2015novel, luo2015nonnegative,luo2019non,li2017modified}.
\begin{figure}[!t]
\footnotesize\centering
\setlength{\tabcolsep}{1.2pt}
\begin{tabular}{cc}
\includegraphics[width=0.25\linewidth]{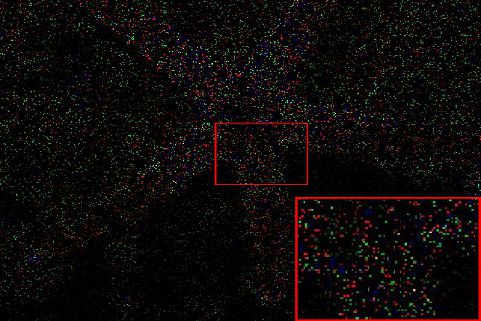}&
\includegraphics[width=0.25\linewidth]{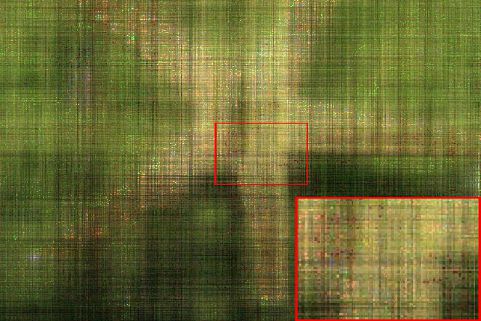}\\
 Observed (PSNR 2.03 dB) & SNN (PSNR 15.47 dB)\\
\includegraphics[width=0.25\linewidth]{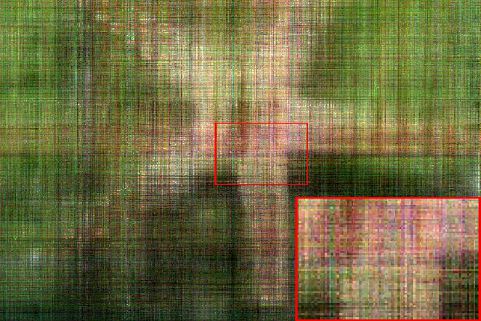}&
\includegraphics[width=0.25\linewidth]{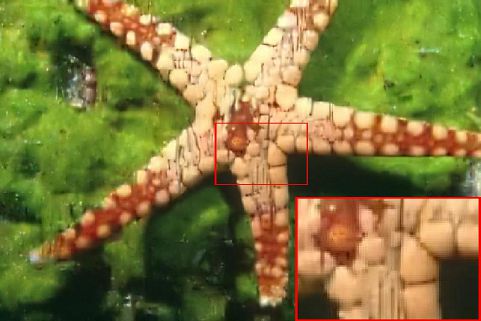}\\
TNN (PSNR 15.73 dB)& DP3LRTC (PSNR 23.77 dB)
  \end{tabular}
  \caption{The recovered results by SNN, TNN, and the poposed DP3LRTC on color image \textit{Starfish} with the sampling rate $5\%$ .}
  \label{fig:highlight}
\end{figure}
By exploiting the redundancy of natural images, the image completion problem can be formulated as the following low-rank matrix completion (LRMC) model:
 \begin{equation}\label{equ:lrmc}
  \arg\min_{{\mathbf X}}\   \text{rank}({\mathbf X}) \quad \text{s.t.}\ \mathcal{P}_{\Omega}({\mathbf X}) = \mathcal{P}_{\Omega}({\mathbf O}),
\end{equation}
where ${\mathbf X}$ is the underlying matrix, $\mathcal{\mathbf O}$ is the observed incomplete matrix, $\Omega$ is the index set corresponding to the observed entries, $\mathcal{P}_{\Omega}(\cdot)$ is the projection function that keeps the entries  in $\Omega$ while setting others be zeros, {and the elements of the observation $\mathbf O$ in the set $\Omega$ are given while the remaining elements are missing.}
For gray images, the low-rankness is characterized by the matrix rank or its convex envelope, i.e., the nuclear norm. For multi-dimensional images, the image elements are usually reordered into a matrix, which is known as matricization or unfolding.

However, matricization will inevitably destroy the intrinsic structure of multi-dimensional images. As the higher-order extension of the matrix, the tensor can provide a more natural and compact representation for multi-dimensional images. Thus we can formulate the image completion problem as the following low-rank tensor completion (LRTC) model:
\begin{equation}\label{equ:lrtc}
  \arg\min_{\mathcal{X}}\   \text{rank}(\mathcal{X}) \quad \text{s.t.}\ \mathcal{P}_{\Omega}(\mathcal{X}) = \mathcal{P}_{\Omega}(\mathcal{O}),
\end{equation}
where $\mathcal{X}$ is the underlying tensor and $\mathcal{O}$ is the observed incomplete tensor (as shown in the top-left of Fig. \ref{fig:highlight}). Here, we generally consider the  element-wise sampling in the model (\ref{equ:lrtc}), which can be extended to the structured sampling, e.g., inpainting \cite{zhuang2018fast} and demosaicing \cite{zhuang2018hy}.

 %Generally, it has been widely  recognized that tensor-based models show superiority over the matrix-based ones \cite{liu2013tensor,sidiropoulos2017tensor}.

Different from the matrix case, there is not an exact (or unique) definition for tensor rank. Thus, how to define the tensor rank becomes an important task. Many research efforts have been devoted to this topic \cite{kolda2009tensor,wang2017tensor,LONG2019tensor,Sidiropoulos2017tensor}, such as the CANDECOMP/PARAFAC (CP) rank and the Tucker rank.
The CP rank is defined based on the CP decomposition, where an $n$-th order tensor is decomposed as the sum of rank-one tensors \cite{che2017neural,wang2019neural}, i.e., the outer product of $n$ vectors. The CP rank is defined as the minimal number of the rank-one tensors required to express the target tensor \cite{Zhao2015Bayesian}. However, its related optimization is difficult, and even determining the CP rank of a given tensor is NP-hard \cite{hillar2013most}.
The Tucker rank is based on the Tucker decomposition which decomposes a tensor into a core tensor multiplied by a matrix along each mode \cite{liu2013tensor,gandy2011tensor,xie2018kronecker,tan2013low}. The Tucker rank is defined as the vector consisting of the ranks of unfolding matrices. The Tucker rank has been considered in the LRTC problem  by minimizing its convex surrogate, i.e., the sum of the nuclear norm (SNN) \cite{liu2013tensor}, or the non-convex surrogates\cite{ji2017non,cao2015folded}. The top-right of Fig. \ref{fig:highlight} exhibits the recovered result by the SNN-based model. However, unfolding the tensor along each mode will also destroy the intrinsic structures of the tensor.

The tensor singular value decomposition (t-SVD), based on the tensor-tensor product (t-product), has been emerged as a powerful tool for preserving the intrinsic structures of the tensor \cite{kilmer2011factorization,kilmer2013third}.
Although the t-SVD is originally suggested for third-order tensors,  it has been extended to $n$-th order tensors ($n\geq3$) \cite{martin2013order,zheng2018tensor}. Based on the t-SVD, the corresponding multi-rank and tubal-rank have received considerable attention. The tensor nuclear norm (TNN) \cite{zhang2014novel,lu2016tensor,jiang2020multi,zheng2019mixed,wang2019noisy} is suggested as a convex surrogate of the multi-rank.  The TNN-based LRTC model is given as follows:
\begin{equation}\label{equ:tnnmodel}
  \arg\min_{\mathcal{X}}   \left\|\mathcal{X}\right\|_\text{TNN} \quad \text{s.t.}\quad \mathcal{P}_{\Omega}(\mathcal{X}) = \mathcal{P}_{\Omega}(\mathcal{O}),
\end{equation}
where $\mathcal{X}\in\mathbb{R}^{n_1\times n_2\times n_3}$, $\left\|\mathcal{X}\right\|_\text{TNN}=\sum_{i=1}^{n_{3}} \left\|\bar{\mathbf{X}}^{(i)} \right \|_{\ast}$, $\bar{\mathbf{X}}^{(i)}$ is the $i$-th frontal slice of $\mathcal{\bar{X}}$, and  $\bar{\mathcal{X}}$ is the tensor generated by performing discrete Fourier transformation (DFT) along the mode-3 fibers (tubes) of $\mathcal{X}$, i.e., $\mathcal{\bar{X}}={\tt{fft}}(\mathcal{X},[],3)$.
The TNN regularizer has the advantage of preserving the global structure of multi-dimensional images. We can observe from Fig. \ref{fig:highlight} that the recovered result by the TNN-based method (bottom-left) is slightly better than that by the SNN-based method (top-right).

%A recently emerged definition of tensor rank called multi-rank was proposed in based on tensor singular value decomposition (t-SVD). Briefly speaking, for a third-order tensor $\mathcal{X} \in \mathbb{R}^{n_{1}\times n_{2}\times n_{3}}$, t-SVD firstly performs one-dimensional discrete Fourier transformation along the tubes and gets $\bar{\mathcal{X}}$. Then each frontal slice of $\bar{\mathcal{X}}$, denoted as $\bar{\mathbf{X}}^{(i)}$ $(i=1,\cdots,n_3)$ , is decomposed in the matrix SVD format, and t-SVD finally performs the inverse Fourier transform (see Algorithm \ref{alg:tsvd}).

 %or simply vectorizing the elements beyond the third-order, e.g., reshaping a tensor of size ${n_1\times n_2\times n_3\times n_4\times n_5}$ as a tensor of size ${n_1\times n_2\times n_3 n_4 n_5}$.
Although promoting the low-rankness of the underlying data has shown its effectiveness for tensor completion, these methods suffer from two drawbacks. {\color{red}First, many real-world multi-dimensional images contain abundant fine details. This makes these multi-dimensional images are not strictly low-rank in mathematics.
Second, when the sampling rate is extremely low or structurally sampled, the rank of these observed entries is already low.  The low-rank regularization is insufficient for a fancy recovery of the underlying tensor.} This phenomena can be observed in Fig. \thinspace \ref{fig:highlight}, where results by the SNN-based model and  TNN-based model are  low-quality when the sampling rate is 5\%.\\

As a compensation, many LRTC methods introduce {\color{red}additional regularizers, which express other types of image priors, for a better solution of this ill-posed inverse problem.} For example, the total variation (TV) regularizer, framelet regularizer, and nonlocal regularizer have received much attention \cite{li2017low, zheng2019low, li2019low, jiang2018matrix}. In particular, the TV regularizer was incorporated into the TNN-based LRTC model by exploiting the local smoothness \cite{jiang2018anisotropic}. Recently, deep learning-based methods have been developed to learn data-driven priors from a large number of natural images \cite{zhang2017beyond, zhang2018ffdnet, zhang2017image} and have shown promising performance on extensive application. {\color{red}Therefore, we consider to leverage the power of deep learning-based methods.}\\

{\color{red}However, from the literatures, convolutional neural networks (CNNs) are successfully applied for the recovery of 2-D gray level images or the color images. There is not a general CNN structure suitable for various types of multi-dimensional imaging data. Fortunately, the multi-dimensional imaging data can be viewed as consisting of the typical 2-D gray level images. For example, each frame of a video is indeed a 2-D image and each band of a multi-spectral image looks also like a 2-D image. Thus, we believe that the CNNs designed for 2-D images can be utilized in LRTC.\\}

In this work, we attempt to exploit the respective strengths of the deterministic low-rankness prior and the data-driven deep image prior. By integrating the low-rankness prior with deep image prior expressed by an implicit regularizer, we suggest the novel LRTC model as follows:
\begin{equation}\label{equ:model}
     \arg\min_{\mathcal{X}} \left\| \mathcal{X}\right\|_\text{TNN}+\lambda \Phi(\mathcal{X}) \quad \text{s.t.}\quad \mathcal{P}_{\Omega}(\mathcal{X}) = \mathcal{P}_{\Omega}(\mathcal{O}),
\end{equation}
where $\Phi(\mathcal{X})$ is an implicit regularizer to plug in the denoising CNNs (termed as deep denoisers). In (\ref{equ:model}), the two regularizers are organically combined and benefit from each other. {\color{red}On the one hand, the implicit regularizer can bring in deep image priors to each 2-D spatial slice of the multi-dimensional imaging data, well characterizing the fine details which can hardly be captured by the TNN regularizer. On the other hand, TNN can enforce the inner global correlations of the multi-dimensional imaging data, avoiding the lackness of lack inherent relations that the slices, which are separately handled by the CNN, . $\left\| \mathcal{X}\right\|_\text{TNN}$ and $\Phi(\mathcal{X})$ respectively regularize the coarse structure and the fine details, being complementary to each other.}
To efficiently solve the proposed model, we develop the alternating direction method of multipliers (ADMM)  \cite{boyd2011distributed} under the highly flexible Plug-and-Play (PnP) framework  \cite{sreehari2016plug,venkatakrishnan2013plug,chan2016plug}, which allows us to plug in the stat-of-the-art deep denoisers. We term our method
as \textbf{dee}p \textbf{p}lug-and-\textbf{p}lay \textbf{p}rior for \textbf{l}ow-\textbf{r}ank \textbf{t}ensor \textbf{c}ompletion (DP3LRTC). From the bottom-right of Fig.\thinspace\ref{fig:highlight}, we can observe that the deterministic  low-rankness prior and data-driven deep image prior contribute to the superior performance of the proposed DP3LRTC.

The rest of this paper is organized as follows. Section \ref{sec:Bac} presents minimal preliminaries necessary for the subsequent discussion. Section \ref{sec:Mod} gives the corresponding solving algorithm to tackle the proposed model. Section \ref{sec:Exp} gives the experimental results and discusses details about the DP3LRTC.
Section \ref{sec:Con} concludes this paper.

\section{Preliminaries}\label{sec:Bac}
%, we introduce some basic preliminaries of tensor notations, the t-SVD scheme, and the plug-and-play framework with CNNs designed for denoising.
In this section, we introduce minimal and necessary preliminaries of  the modular parts, i.e., TNN and  PnP, for the subsequent discussion.

\subsection{Notations} \label{subsec:Not}
In this subsection, we give the basic notations and briefly introduce some definitions. We denote vectors as bold lowercase letters (e.g., $\mathbf{x}$), matrices as uppercase letters (e.g., $\mathbf{X}$), and tensors as calligraphic letters (e.g., $\mathcal{X}$). For a third-order tensor $\mathcal{X}\in \mathbb{R}^{n_1\times n_2\times n_3}$, with the MATLAB notation, we denote its $(i,j,k)$-th element as $\mathcal{X}(i,j,k)$ or $\mathcal{X}_{i,j,k}$. Its $(i,j)$-th mode-1, mode-2, and mode-3 fibers are denoted as $\mathcal{X}(:,i,j)$, $\mathcal{X}(i,:,j)$, and $\mathcal{X}(i,j,:)$, respectively. We use $ \mathcal{X}(i,:,:)$, $\mathcal{X}(:,i,:)$, and $\mathcal{X}(:,:,i)$ to denote the $i$-th horizontal, lateral, and frontal slices of $\mathcal{X}$, respectively.
More compactly, we also use $\mathbf{X}^{(i)}$ to represent the $i$-th frontal slices $\mathcal{X}(:,:,i)$.
The Frobenius norm of $\mathcal{ X}$ is defined as $\|\mathcal{X}\|_F:=\sqrt{\sum_{i,j,k}|\mathcal{X}(i,j,k)|^2}$.

\subsection{T-SVD and TNN} \label{subsec:tSVD}
%For a third-order tensor $\mathcal{X}\in \mathbb{R}^{n_1\times n_2\times n_3}$,
%the block circulation operation \cite{kilmer2013third} is defined as
%\begin{equation*}
%\setlength{\arraycolsep}{0.6pt}
%\begin{aligned}
%&\mathbf{\tt bcirc}(\mathcal{X}):=
%\begin{pmatrix}
%\mathbf{X}^{(1)} & \mathbf{X}^{(n_{3})} &  \cdots & \mathbf{X}^{(2)} \\
%    \mathbf{X}^{(2)} & \mathbf{X}^{(1)} &  \cdots & \mathbf{X}^{(3)} \\
%    \vdots      & \vdots      &  \ddots &  \vdots             \\
%    \mathbf{X}^{(n_{3})} & \mathbf{X}^{(n_{3}-1)} &  \cdots & \mathbf{X}^{(1)} \\
%\end{pmatrix}\in \mathbb{R}^{n_1n_3\times n_2n_3}.\\
%\end{aligned}
%\end{equation*}

%The block diagonalization operation and its inverse operation are defined as
%\begin{equation*}
%\setlength{\arraycolsep}{0.6pt}
%\begin{aligned}
%&\mathbf{\tt bdiag}(\mathcal{X}):=
%\begin{pmatrix}
%\mathbf{X}^{(1)} &  &  &  \\
%      & \mathbf{X}^{(2)} &  & \\
%      &  &  \ddots &          \\
%      &  &  & \mathbf{X}^{(n_{3})} \\
%\end{pmatrix}\in \mathbb{R}^{n_1n_3\times n_2n_3},\\
%\end{aligned}
%\end{equation*}
%\begin{equation*}
%  \mathbf{\tt unbdiag}\big(\mathbf{\tt bdiag}(\mathcal{X})\big):=\mathcal{X}.\\
%\end{equation*}
%
%The block vectorization operation and its inverse operation are defined as
%\begin{equation*}
%\setlength{\arraycolsep}{0.3pt}
%\begin{aligned}
%\mathbf{\tt bvec}&(\mathcal{X}):=
%\begin{pmatrix}
%(X^{(1)})^{\text{T}},&
%(X^{(2)})^{\text{T}},&
%\cdots,&
%(X^{(n_3)})^{\text{T}}
%\end{pmatrix}^{\text{T}}\in \mathbb{R}^{n_1n_3 \times n_2},~~\mathbf{\tt bvfold}\big(\mathbf{\tt bvec}(\mathcal{X})\big):=\mathcal{X}.
%\end{aligned}
%\end{equation*}
Recently, Kilmer have introduced the novel tensor-tensor product as follows:
\begin{myDef}[t-product \cite{kilmer2013third}]
\emph{Let $\mathcal{A}$ be $n_1\times n_2 \times n_3$ and $\mathcal{B}$ be $n_2 \times n_4 \times n_3$. Then the tensor-tensor product (t-product) is the
$n_1\times n_4 \times n_3$ tensor $\mathcal{C}=\mathcal{A}*\mathcal{B}$ whose $(i,j)$th tube is given by
%\begin{equation*}
%\mathcal{C}=\mathcal{A}*\mathcal{B}\Leftrightarrow \mathcal{C}(i,j,:)=\sum_{t=1}^{n_3}\mathcal{A}(i,t,:)\star \mathcal{B}(t,j,:)
%\end{equation*}
\begin{equation}\mathcal{C}(i,j,:)=\sum_{k=1}^{n_2}\mathcal{B}(i,k,:)\star\mathcal{C}(k,j,:),\end{equation}
where ``$\star$" denotes circular convolution of two vectors.}
\end{myDef}

%\emph{The t-product between two three-way tensors $\mathcal{X}\in \mathbb{R}^{n_1\times n_2\times n_3}$ and $\mathcal{Y}\in \mathbb{R}^{n_2\times n_4\times n_3}$ is defined as
%$$\mathcal{X}\ast \mathcal{Y}:=\mathrm{\tt bvfold}\big(\mathrm{\tt bcirc}(\mathcal{X}) \mathrm{\tt bvec}(\mathcal{Y})\big)\in \mathbb{R}^{n_1\times n_4\times n_3}.$$}
%\end{myDef}
%Indeed, the t-product can be regarded as a matrix-matrix multiplication, except that the multiplication operation between scalars is replaced by circular convolution between the tubes, i.e.,\\
%\begin{equation*}
%\qquad \mathcal{F}=\mathcal{X} * \mathcal{Y} \Leftrightarrow \mathcal{F}(i, j,:)=\sum_{t=1}^{n_{2}} \mathcal{X}(i, t,:) \star \mathcal{Y}(t, j,:),
%\end{equation*}
%
%where $\star$ denotes the circular convolution between two tubes. Note that the circular convolution in the spatial domain is equivalent to the multiplication in the Fourier domain, the t-product between two tensors
%$\mathcal{F}=\mathcal{X} * \mathcal{Y} $ { is equivalent to } \\
%\begin{equation*}
%\overline{\mathcal{F}}=\text { bdfold }(\text{bdiag}(\overline{\mathcal{X}}) \text{bdiag}(\overline{\mathcal{Y}})) .
%\end{equation*}

\begin{myDef}[special tensors \cite{kilmer2013third}]\emph{The \textbf{conjugate transpose} of a third-order tensor $\mathcal{X}\in \mathbb{R}^{n_1\times n_2\times n_3}$, denote as $\mathcal{X}^{\mathrm{H}}$, is the tensor obtained by conjugate transposing each of the frontal slices and then reversing the order of transposed frontal slices 2 through $n_3$.} \emph{The \textbf{identity tensor} $\mathcal{I} \in \mathbb{R}^{n_1\times n_2\times n_3}$ is the tensor whose first frontal slice is the identity matrix, and other frontal slices are all zeros.} \emph{A third-order tensor $\mathcal{Q}$ is \textbf{orthogonal} if $\mathcal{Q}\ast \mathcal{Q}^{\mathrm{H}}=\mathcal{Q}^{\mathrm{H}}\ast \mathcal{Q}=\mathcal{I}.$}
\emph{A third-order tensor $\mathcal{S}$ is \textbf{f-diagonal} if each of its frontal slices is a diagonal matrix.}
\end{myDef}

Based on t-product,  the tensor singular value decomposition (t-SVD) is suggested.
\begin{theorem}[t-SVD \cite{kilmer2013third}]
\emph{Let $\mathcal{X}\in \mathbb{R}^{n_1\times n_2\times n_3}$ be a third-order tensor, then it can be decomposed as
\begin{equation}\mathcal{X}=\mathcal{U}\ast \mathcal{S}\ast \mathcal{V}^{\mathrm{H}},\end{equation}
where $\mathcal{U}\in \mathbb{R}^{n_1\times n_1\times n_3}$ and $\mathcal{V}\in \mathbb{R}^{n_2\times n_2\times n_3}$ are the orthogonal tensors, and $\mathcal{S}\in \mathbb{R}^{n_1\times n_2\times n_3}$ is a f-diagonal tensor whose  frontal slices are  diagonal matrices.}
\end{theorem}

The t-SVD can be efficiently obtained by computing a series of matrix SVDs in the Fourier domain; see Algorithm \ref{alg:tsvd} for more details.\\
\begin{algorithm}[h]
\renewcommand\arraystretch{0.8}
\caption{The t-SVD for a third-order tensor.}
\begin{algorithmic}[1]
\renewcommand{\algorithmicrequire}{\textbf{Input:}} % Use Input in the format of Algorithm
\renewcommand{\algorithmicensure}{\textbf{Output:}}
\Require
$\mathcal{X}\in\mathbb{R}^{n_1\times n_2\times n_3}$.
\State $\bar{\mathcal{X}}\leftarrow$ $\tt{fft}$$(\mathcal{X},[],3)$.
\For {$i=1$ to $n_3$}
\State $[ U, S, V]=$ $\tt{svd}$$(\bar{X}^{(i)})$.
\State $\bar{U}^{(i)}\leftarrow U$; $\bar{S}^{(i)}\leftarrow S$;
$\bar{V}^{(i)}\leftarrow V$.
\EndFor
\State
$\mathcal{U}\leftarrow$ $\tt{ifft}$$(\bar{\mathcal{U}},[],3)$.\\
$\mathcal{S}\leftarrow$ $\tt{ifft}$$(\bar{\mathcal{S}},[],3)$.\\
$\mathcal{V}\leftarrow$ $\tt{ifft}$$(\bar{\mathcal{V}},[],3)$.%\\
%\Return $\mathcal{Y}$;\\
%\Return $\mathbf{\mathcal{U}}$, $\mathbf{\mathcal{S}}$ and $\mathbf{\mathcal{V}}$
\Ensure
$\mathcal{U}$,
$\mathcal{S}$,
$\mathcal{V}$.
\end{algorithmic}
\label{alg:tsvd}
\end{algorithm}

 Based on t-SVD, we have the corresponding definitions of  tensor   multi-rank and tubal-rank.
\begin{myDef}[tensor tubal-rank and multi-rank \cite{zhang2014novel}]
\label{def:rank}
\emph{Let $\mathcal{X}\in\mathbb{R}^{n_1\times n_2\times n_3}$ be a third-order tensor, the tensor multi-rank, denoted as $\text{rank}_{\text{m}}(\mathcal{X}) \in\mathbb{R}^{n_3}$, is a vector whose $i$-th element is the rank of the $i$-th frontal slice of $\bar{\mathcal{X}}$, where $\bar{\mathcal{X}}=\mathrm{\tt fft}(\mathcal{X},[],3)$.
The tubal-rank of $\mathcal{X}$, denote as $\text{rank}_{\text{t}}(\mathcal{X})$, is defined as the number of non-zero tubes of $\mathcal{S}$, where  $\mathcal{X}=\mathcal{U}\ast \mathcal{S}\ast \mathcal{V}^{\mathrm{H}}$.}
\end{myDef}

The relationship between the tubal-rank and multi-rank is $\text{rank}_{\text{t}}(\mathcal{X})=\max\big(\text{rank}_{\text{m}}(\mathcal{X})\big)$. The convex surrogate of the multi-rank is suggested as follows:
\begin{myDef}[TNN \cite{zhang2014novel}]
\label{def:tnn}
\emph{The tensor nuclear norm of a tensor $\mathcal{X}\in \mathbb{R}^{n_{1}\times n_2\times n_{3}}$, denoted as $\|\mathcal{X}\|_{\text{TNN}}$, is defined as the sum of singular values of all the frontal slices of $\bar{\mathcal{X}}$, i.e.,}
\begin{equation}
\left\|\mathcal{X}\right\|_\text{TNN}:=\sum_{i=1}^{n_3}\|\bar{\mathbf{X}}^{(i)}\|_{*}.
\end{equation}
\emph{where $\bar{\mathbf{X}}^{(i)}$ is the $i$-th frontal slice of $\mathcal{\bar{X}}$, and $\bar{\mathcal{X}}=\mathrm{\tt fft}(\mathcal{X},[],3)$.}
\end{myDef}

\subsection{Plug-and-Play (PnP) Framework} \label{subsec:FFDNet}

PnP is a highly flexible framework \cite{sreehari2016plug,venkatakrishnan2013plug,chan2016plug} that leverages the power of the state-of-the-art denoisers in the ADMM or other proximal algorithms. After variable splitting technique, the optimization problem is decoupled into easier subproblems, one of which is the proximal operator of regularization. The proximal operator of regularization $\textrm{prox}_{\Phi}: \mathbb{R}^{n}\rightarrow \mathbb{R}^{n}$  is  defined as
 \begin{equation}\label{prox}
\textrm{prox}_{\Phi}(y)=\textrm{arg}\mathop{\textrm{min}}_{x}\left\lbrace \Phi(x) + \frac{\rho}{2}\left\|x-y\right\|^2\right\rbrace,
\end{equation}
which maps the input $y$ to the minimizer of (\ref{prox}).  Under the PnP framework, the proximal operator of regularization
is replaced by  the denoising algorithm (termed as denoiser), which maps the noisy image to the clean image. Here the denoiser serves as an implicit regularizer to express the deep image prior.
Regularization by Denoising (RED) \cite{romano2017little,reehorst2018regularization}, which uses the denoising engine in defining the regularization of the inverse problem, is also suggested as an alternative framework.

The state-of-the-art denoisers can be flexibly plugged in as a modular part under the PnP framework. In the past decades, the hand-crafted denoisers, which exploit the deterministic priors of the nature images, have been dominantly used, e.g., TV denoiser \cite{rudin1992nonlinear}, framelet denoiser \cite{cai2012framelet},  BM3D  denoiser \cite{dabov2007video}, WNNM denoiser \cite{gu2014weighted}, and ITS denoiser \cite{Xie2016multispectral}. Recently, deep learning-based denoisers have been rapidly developed to learn data-driven image priors from a large number of natural images \cite{zhang2017beyond,zhang2018ffdnet,zhang2017image} and have shown promising performance.
The PnP framework, which integrates modern denoiser, has shown great empirical success on diverse applications including denoising \cite{,he2018non,zhuang2018fast}, restoration\cite{zhang2017learning}, tomography, super-resolution, and fusion.

However, even the most basic question of convergence is still a challenging problem. The convergence of the PnP framework has been considered from the fixed-point viewpoint in \cite{chan2016plug}. Very recently, the convergence of the PnP framework with properly trained denoisers has been considered in \cite{ryu2019plug}.

\section{The Proposed Model and Algorithm}\label{sec:Mod}
{\color{red}As previously discussed, given an incomplete observation $\mathcal{O}\in\mathbb{R}^{n_1\times n_2\times n_3}$ with its support $\Omega$ (namely, the index set corresponding to the observed entries), our DP3LRTC model is formulated as:
\begin{equation}\label{equ:model2}
     \arg\min_{\mathcal{X}} \left\| \mathcal{X}\right\|_\text{TNN}+\lambda \Phi(\mathcal{X}) \quad \text{s.t.}\quad \mathcal{P}_{\Omega}(\mathcal{X}) = \mathcal{P}_{\Omega}(\mathcal{O}).
\end{equation}
In our model, $\Phi(\mathcal{X})$ is an implicit reuglarizer, whose related subproblem will be solved by a CNN within the PnP framework. We make some remarks for our model. When using CNNs, a straightforward idea is to design a network structure and conduct end-to-end training. However, this may lacks flexibility and generalization ability for different kinds of data and different kinds of degradation, which is corresponding to different sampling situations in the tensor completion problem. Also, training a 3-D CNN also costs much more time than training a 2-D CNN. Therefore, in our model, we directly plug in a 2-D denoising CNN, which is trained on the natural image dataset, to solve the $\Phi(\mathcal{X})$ related subproblem. Specifically, the denoising CNN hereof acts on the spatial slices of the multi-dimensional data, for example bands of MSIs and frames of videos, to preserve the spatial details. Meanwhile, the whole structure of the data is captured by the low-rank term $\left\| \mathcal{X}\right\|_\text{TNN}$, which would dominate the correlation along the temporal or spectral direction. Thus, our model can deal with different types of multi-dimensional data and various sampling situations in a flexible and economic manner.}\\

Then, we develop the ADMM \cite{boyd2011distributed} to tackle the minimization problem  \eqref{equ:model2} under the PnP framework. First, we denote the indicator function as
\begin{equation}
1_{\mathbb{S}}(\mathcal{X})=\left\{\begin{array}{ll}{0,} & {\text { if } \mathcal{X} \in \mathbb{S}}, \\ {\infty,} & {\text { otherwise, }}\end{array}\right.
\end{equation}
where $\mathbb{S} :=\left\{\mathcal{X} \in \mathbb{R}^{n_{1} \times n_{2} \times n_{3}}, \mathcal{P}_{\Omega}(\mathcal{X}) = \mathcal{P}_{\Omega}(\mathcal{O})\right\}$ and introduce two auxiliary variables $\mathcal{Y}$ and $\mathcal{Z}$.
Then, we reformulate \eqref{equ:model}  as a constrained optimization problem, i.e.,
\begin{equation}\label{equ:model2}
    \begin{aligned}
     \arg\min_{\mathcal{X,Y,Z}}\ &\left\| \mathcal{Y}\right\|_\text{TNN}+\lambda \Phi(\mathcal{Z})+1_{\mathbb{S}}(\mathcal{X})\\ \text{s.t.}\quad &\mathcal{Y}=\mathcal{X},\ \mathcal{Z}=\mathcal{X}.
 \end{aligned}
\end{equation}

The augmented Lagrangian function of (\ref{equ:model2}) is
\begin{equation}\label{lagrangian}
    \begin{aligned}
    L= \left\| \mathcal{Y}\right\|_\text{TNN}+\lambda \Phi(\mathcal{Z})+1_{\mathbb{S}}(\mathcal{X})+\langle\mathcal{X}-\mathcal{Y},\Lambda_{1}\rangle+\frac{\beta}{2}\left\|\mathcal{X}-\mathcal{Y}\right\|_{F}^{2}
     +\langle\mathcal{X}-\mathcal{Z},\Lambda_{2}\rangle+\frac{\beta}{2}\left\|\mathcal{X}-\mathcal{Z}\right\|_{F}^{2},
\end{aligned}
\end{equation}
where $\Lambda_1$ and $\Lambda_2$ are the Lagrangian multipliers, and $\beta$ is a nonnegative penalty parameter.

According to the ADMM framework, the solution of (\ref{equ:model}) can be found by solving a sequence of subproblems.

In \textbf{Step 1}, we need to solve the $[\mathcal{Y},\mathcal{Z}]$-subproblem. Since the variables $\mathcal{Y}$ and $\mathcal{Z}$ are decoupled, their optimal solutions can be calculated separately.

1) The $\mathcal{Y}$-subproblem is
\begin{equation}\label{equ:ypro}
     \arg\min_{\mathcal{Y}} \left\| \mathcal{Y}\right\|_\text{TNN}+\frac{\beta}{2}\left\|\mathcal{X}^l-\mathcal{Y}+\Lambda_1^l/\beta\right\|_{F}^{2}.
\end{equation}

Let the t-SVD of $(\mathcal{X}^l+\Lambda_1^l/\beta)$ be $\mathcal{U} \ast \mathcal{S} \ast \mathcal{V}^{\mathrm{H}}$, the closed-form solution of  the $\mathcal{Y}$-subproblem can be exactly calculated via singular value thresholding (SVT)  \cite{cai2010singular,lu2018tensor} as
\begin{equation}\label{equ:ysolution}
  \mathcal{Y}^{l+1} = \mathcal{U} \ast \mathcal{D} \ast \mathcal{V}^{\mathrm{H}},
\end{equation}
where $ \mathcal{D}$ is an $f$-diagonal tensor whose each frontal slice in the Fourier domain is $ \bar{\mathcal{D}}(i,i,k) = \max\{\bar{\mathcal{S}}(i,i,k) - \frac{1}{\beta} ,0\}$. The complexity of computing $\mathcal { Y }$ is ${O}(n_3 \operatorname{min} ({n_1}^{2}n_2, n_1{n_2}^{2})+n_1n_2n_3 \operatorname{log} (n_3))$.

%$\bar{\mathcal{S}}$ be the result of DFT of $\mathcal{S}$ along the tubes. Then each element of the singular tubes of $\bar{\mathcal{Y}}^{l+1}$ is the result of multiplying every entry $\bar{\mathcal{S}}(i,i,k)$ with $[1-\frac{1/\beta}{\bar{\mathcal{S}}(i,i,k)}]_{+}$ \cite{cai2010singular}, where ``$[\cdot]_+$'' denotes keeping the positive part. In other word, the closed form solution of problem (\ref{equ:ypro}) can be obatined by the singular value thresholding (SVT) operator as

2) The $\mathcal{Z}$-subproblem is
\begin{equation}\label{equ:zpro1}
  \arg\min_{\mathcal{Z}} \lambda\Phi(\mathcal{Z})+\frac{\beta}{2}\left\|\mathcal{X}^l-\mathcal{Z}+\Lambda_2^l/\beta\right\|_{F}^{2}.
\end{equation}

 Let $\sigma = \sqrt{\lambda/\beta}$, then, (\ref{equ:zpro1}) can be rewritten as
\begin{equation}\label{equ:zpro2}
  \textrm{prox}_{\Phi}(\mathcal{Z})=  \arg\min_{\mathcal{Z}} \Phi(\mathcal{Z}) + \frac{1}{2\sigma^{2}}\left\|\mathcal{Z}-\mathcal{X}^{l}-\Lambda_2^l/\beta\right\|_{F}^{2}.
\end{equation}
%Treating $\mathcal{X}^{l}+\Lambda_2^l/\beta$ as a ``noisy image'', (\ref{equ:zpro2}) can be regarded as minimizing the residual of ``noisy image'' and the ``clean image'' $\mathcal{Z}$ using the prior expressing by the regularizer $\Phi(\mathcal{Z})$.
%With this idea, in PnP framework \cite{sreehari2016plug, buzzard2018plug, reehorst2018regularization}, we utilize the FFDNet as the denoiser to solve the related $\mathcal{Z}$ subproblem.
Under the PnP
framework, the proximal operator of  regularization $\textrm{prox}_{\Phi}: \mathbb{R}^{n_1\times n_2\times n_3}\rightarrow \mathbb{R}^{n_1\times n_2\times n_3}$   is replaced by the deep learning-based denoiser,  which maps the noisy image to the clean image. Here we consider the fast and flexible denoising convolutional neural network, namely FFDNet  \cite{zhang2018ffdnet}, as the deep denoiser.
Feeding $\mathcal{X}^{l}+\Lambda_2^l/\beta$ into the denoiser FFDNet, we obtain the solution of the $\mathcal{Z}$-subproblem as
\begin{equation}\label{equ:zsolution}
  \mathcal{Z}^{l+1} = {\tt FFDNet} (\mathcal{X}^{l}+\Lambda_2^l/\beta,\sigma).
\end{equation}
In FFDNet the parameter $\sigma$ is related to the noise level, but here $\sigma$  is related to the  error level between the estimation and ground truth.
Here we consider the pre-trained  FFDNet   denoiser which belongs to the CNN denoiser family.  We feed the color images into the the FFDNet denoiser  trained for color images and   the spatial slices of gray videos and MSIs into the the FFDNet denoiser  trained for gray images. The complexity of computing $\mathcal {Z}$ is ${O}(n_1n_2n_3n_ln_fn_k)$, where $n_l$ is the number of layers, $n_f$ is the number of features, and $n_k$  is the number  of kernel pixels.\\

In \textbf{Step 2}, we need to solve the $\mathcal{X}$-subproblem:
\begin{equation}\label{equ:xpro}
\begin{aligned}
  \arg\min_{\mathcal{X}}\  1_{\mathbb{S}}(\mathcal{X}) +\frac{\beta}{2}\left\|\mathcal{X}-\mathcal{Y}^{l+1}+\Lambda_1^l/\beta\right\|_{F}^{2}
  +\frac{\beta}{2}\left\|\mathcal{X}-\mathcal{Z}^{l+1}+\Lambda_2^l/\beta\right\|_{F}^{2}.
\end{aligned}
\end{equation}
By minimizing the $\mathcal{X}$-subproblem, we  have $1_{\mathbb{S}}(\mathcal{X}) = 0$, i.e., $\mathcal{X} \in \mathbb{S}$.  Thus, the closed-form solution of  $\mathcal{X}$-subproblem is given as follows:
\begin{equation}\label{equ:xsolution}
  \begin{cases}
    \begin{aligned}
          &\mathcal{P}_{\Omega}(\mathcal{X}^{l+1}) = \mathcal{P}_{\Omega}(\mathcal{O}),\\
          &\mathcal{P}_{\Omega^c}(\mathcal{X}^{l+1})  = \mathcal{P}_{\Omega^c}(\frac{\beta\mathcal{Y}^{l+1}+\beta\mathcal{Z}^{l+1}-\Lambda_1^l-\Lambda_2^l}{2\beta}),
    \end{aligned}
  \end{cases}
\end{equation}
where $\Omega^{c}$ denotes the complementary set of $\Omega$. The complexity of computing $\mathcal {X}$ is ${O}(n_1n_2n_3)$.

In \textbf{Step 3}, we update the multipliers $\Lambda_1$ and $\Lambda_2$ as follows:
\begin{equation}\label{equ:uplam}
  \begin{cases}
    \begin{aligned}
          \Lambda_1^{l+1} = &\Lambda_1^l + \beta(\mathcal{X}^{l+1}- \mathcal{Y}^{l+1}),\\
          \Lambda_2^{l+1} = &\Lambda_2^l+ \beta(\mathcal{X}^{l+1}- \mathcal{Z}^{l+1}).
    \end{aligned}
  \end{cases}
\end{equation}
The complexity of updating $\Lambda_1$ and $\Lambda_2$  is ${O}(n_1n_2n_3)$.

Finally, the  algorithm is summarized in Algorithm \ref{alg:opt}. The computational complexity is ${O}(n_3 \operatorname{min} ({n_1}^{2}n_2, n_1{n_2}^{2})+n_1n_2n_3 \operatorname{log} (n_3)+n_1n_2n_3n_ln_fn_k)$.  The main part of the ADMM algorithm is implemented on CPU while the FFDNet  is  evaluated on GPU.
\begin{algorithm}[htbp]
\renewcommand\arraystretch{0.8}
\caption{The ADMM algorithm for solving (\ref{equ:model}).}
\begin{algorithmic}[1]
\renewcommand{\algorithmicrequire}{\textbf{Input:}} % Use Input in the format of Algorithm
\renewcommand{\algorithmicensure}{\textbf{Output:}}
\Require
the observed tensor $\mathcal{O}$, the set index $\Omega$ of the observed entries, the parameters $\beta$ and $\sigma$, and the maximum iteration number $l_{\text{max}}$.
\State \textbf{Initialization}:   the observed tensors $\mathcal{Y},\ \mathcal{Z},$ and  $\mathcal{X}$, the zero tensors $\Lambda_1$ and $\Lambda_2$.
\While {not converged and $l\leq l_{\text{max}}$}
\State
Updating $\mathcal{Y}$ via (\ref{equ:ysolution}),
\State
Updating $\mathcal{Z}$ via (\ref{equ:zsolution}),
\State
Updating $\mathcal{X}$ via (\ref{equ:xsolution}),
\State
Updating multipliers $\Lambda_1$ and $\Lambda_2$ via (\ref{equ:uplam}).
\EndWhile
\Ensure
the recovered tensor $\mathcal{X}$.
\end{algorithmic}
\label{alg:opt}
\end{algorithm}

\section{Numerical Experiments} \label{sec:Exp}
In this section, the performance of DP3LRTC will be comprehensively evaluated by experiments on color images, videos, and multi-spectral images (MSIs). Although the FFDNet denoiser is trained for  color images, the DP3LRTC can be well generalized for gray videos and MSIs. The DP3LRTC is compared with the baseline methods: SNN \cite{liu2013tensor},  TNN  \cite{zhang2014novel}, and TNN-3DTV \cite{jiang2018anisotropic}.

The peak signal to noise ratio (PSNR) and the structural similarity index (SSIM) \cite{wang2004image} are chosen as the quality metrics. We report the mean PSNR and mean SSIM of all bands (or frames). The relative change (RelCha) is adopted as the stopping criterion of all methods, which is defined as
\begin{equation*}
\text{RelCha} = \frac{\left\|\mathcal{X}^{l+1}-\mathcal{X}^{l}\right\|_{F}}{\left\|\mathcal{X}^{l}\right\|_{F}}.
\end{equation*}
In all experiments, when RelCha is smaller than the tolerance $10^{-4}$, we stop the iterations. All parameters involved in different methods are manually selected from a candidate set $\{10^{-4}, 10^{-3}, 10^{-2}, 10^{-1}, 1, 10, 10^{2}\}$ to obtain the highest PSNR value. In this paper,  we mainly consider the  element-wisely sampled, i.e., entries are randomly sampled  for each band (or frame). For color images, we consider the element-wisely sampling, the tubal sampling, i.e., entries are sampled along all RGB channels for the spatial pixel, and the Bayer sampling, i.e., each $2\times2\times3$ cell contains two green pixels, one blue pixel, and one red pixel.
All the experiments are implemented on Windows 10 and Matlab (R2018a) with an Intel(R) Core(TM) i5-4590 CPU at 3.30 GHz, 16 GB RAM, and NVIDIA GeForce GTX 1060 6GB.
\subsection{Color Image Completion}\label{subsec:cic}
In this subsection, 8 color images\footnote{Available at \url{http://sipi.usc.edu/database/database.php}.} are selected for the color images completion experiments. Each image is rescaled to $[0,1]$, and the observed elements are element-wisely sampled with different sampling rates (SR). For color images, we use the FFDNet network trained for  color images as the PnP denoiser.

\begin{table}[hbtp]
\footnotesize
\selectfont
\setlength{\tabcolsep}{1.5pt}
\renewcommand\arraystretch{0.8}
\caption{Quantitative comparison of the results by SNN \cite{liu2013tensor},  TNN  \cite{zhang2014novel}, TNN-3DTV \cite{jiang2018anisotropic}, and the proposed method on color images. The \textbf{best} and \underline{second} best values are highlighted in bold and underlined, respectively.}
\centering
\begin{tabular}{cccccccccccccccc}
\toprule
\multicolumn{1}{c}{\multirow{2}[4]{*}{Image}}& \multicolumn{1}{c}{\multirow{2}[4]{*}{SR}} & \multicolumn{4}{c}{PSNR}     & & \multicolumn{4}{c}{SSIM}& & \multicolumn{4}{c}{Time (s)}\\
\cmidrule{3-16}\multicolumn{2}{c}{} & {\scriptsize SNN}   & {\scriptsize TNN}   & {\scriptsize TNN-3DTV} & {\scriptsize  DP3LRTC} && {\scriptsize SNN}   & {\scriptsize TNN}   & {\scriptsize TNN-3DTV} & {\scriptsize DP3LRTC} && {\scriptsize SNN}   & {\scriptsize TNN}   & {\scriptsize TNN-3DTV} & {\scriptsize DP3LRTC}\\\midrule
\multirow{3}[2]{*}{\shortstack{\textit{Starfish}\\ \\$321\times481\times3$}}
& 10\%  & 18.48  & 19.47  & \underline{22.59}  & \textbf{27.43} && 0.3617  & 0.3007  & \underline{0.6345}  & \textbf{0.8270} && \bf 6 & \underline{8}&  38 &  17 \\
& 20\%  & 22.27  & 22.55  & \underline{24.85}  & \textbf{31.25} && 0.5476  & 0.5052  & \underline{0.7519}  & \textbf{0.9143} && \bf 6 & \underline{8}&  38 &  17 \\
& 30\%  & 24.66  & 25.71  & \underline{26.52}  & \textbf{34.38} && 0.6883  & 0.6685  & \underline{0.8217}  & \textbf{0.9533} && \bf 5 & \underline{9}&  40 &  15 \\
\midrule

\multirow{3}[2]{*}{\shortstack{\textit{Airplane}\\ \\$512\times512\times3$}}
& 10\%  & 20.91  & 21.94  & \underline{23.46}  & \textbf{28.48} && 0.6706  & 0.6623  & \underline{0.8414}  & \textbf{0.9471} && \bf 10 & \underline{17}&  71 &  54 \\
& 20\%  & 24.81  & 25.17  & \underline{25.83}  & \textbf{30.57} && 0.8279  & 0.8242  & \underline{0.9220}  & \textbf{0.9709} && \bf 12 & \underline{19}&  76 &  33 \\
& 30\%  & 27.20  & 27.90  & \underline{28.45}  & \textbf{31.36} && 0.9060  & 0.9014  & \underline{0.9531}  & \textbf{0.9806} && \bf 10 & \underline{18}&  75 &  30 \\
\midrule

\multirow{3}[2]{*}{\shortstack{\textit{Baboon}\\ \\$512\times512\times3$}}
& 10\%  & 17.43  & 17.83  & \underline{19.56}  & \textbf{21.68} && 0.4077  & 0.3959  & \underline{0.5835}  & \textbf{0.7798} && \bf 11 & \underline{17}&  71 &  30 \\
& 20\%  & 19.34  & 20.07  & \underline{20.67}  & \textbf{23.44} && 0.5974  & 0.6005  & \underline{0.7271}  & \textbf{0.8433} && \bf 9 & \underline{18}&  77 &  34 \\
& 30\%  & 21.17  & 21.66  & \underline{21.93}  & \textbf{24.06} && 0.7308  & 0.7346  & \underline{0.8083}  & \textbf{0.9118} && \bf 8 & \underline{19}&  80 &  60 \\
\midrule

\multirow{3}[2]{*}{\shortstack{\textit{Fruits}\\ \\$512\times512\times3$}}
& 10\%  & 20.73  & 20.72  & \underline{24.81}  & \textbf{31.47} && 0.6046  & 0.5646  & \underline{0.8363}  & \textbf{0.9362} && \bf 10 & \underline{17}&  72 &  54 \\
& 20\%  & 24.23  & 24.23  & \underline{27.31}  & \textbf{34.90} && 0.7777  & 0.7510  & \underline{0.9122}  & \textbf{0.9621} && \bf 10 & \underline{18}&  75 &  57 \\
& 30\%  & 26.90  & 26.99  & \underline{29.22}  & \textbf{36.48} && 0.8689  & 0.8541  & \underline{0.9434}  & \textbf{0.9723} && \bf 9 & \underline{19}&  77 &  44 \\
\midrule

\multirow{3}[2]{*}{\shortstack{\textit{Lena}\\ \\$512\times512\times3$}}
& 10\%  & 21.43  & 21.89  & \underline{25.96}  & \textbf{30.93} && 0.6415  & 0.6177  & \underline{0.8396}  & \textbf{0.9241} && \bf 10 & \underline{17}&  71 &  54 \\
& 20\%  & 24.98  & 25.68  & \underline{28.41}  & \textbf{32.92} && 0.8034  & 0.7888  & \underline{0.9069}  & \textbf{0.9423} && \bf 10 & \underline{19}&  77 &  34 \\
& 30\%  & 27.71  & 28.06  & \underline{30.07}  & \textbf{33.46} && 0.8844  & 0.8719  & \underline{0.9380}  & \textbf{0.9575} && \bf 10 & \underline{18}&  75 &  54 \\
\midrule

\multirow{3}[2]{*}{\shortstack{\textit{Watch}\\ \\$768\times1024\times3$}}
& 10\%  & 22.47  & 23.01  & \underline{26.27}  & \textbf{33.96} && 0.7128  & 0.7490  & \underline{0.8863}  & \textbf{0.9825} && \bf 39 & \underline{58}&  250 &  143 \\
& 20\%  & 25.64  & 26.61  & \underline{28.46}  & \textbf{38.18} && 0.8641  & 0.8923  & \underline{0.9466}  & \textbf{0.9888} && \bf 38 & \underline{62}&  255 &  141 \\
& 30\%  & 28.37  & 29.70  & \underline{30.31}  & \textbf{40.89} && 0.9332  & 0.9502  & \underline{0.9709}  & \textbf{0.9971} && \bf 36 & \underline{62}&  262 &  140 \\
\midrule

\multirow{3}[2]{*}{\shortstack{\textit{Opera}\\ \\$586\times695\times3$}}
& 10\%  & 24.23  & 25.05  & \underline{25.56}  & \textbf{30.51} && 0.7499  & 0.7486  & \underline{0.8075}  & \textbf{0.9300} && \bf 17 & \underline{28}&  121 &  87 \\
& 20\%  & 27.55  & 28.24  & \underline{29.13}  & \textbf{33.57} && 0.8649  & 0.8734  & \underline{0.9132}  & \textbf{0.9628} && \bf 15 & \underline{29}&  124 &  89 \\
& 30\%  & 29.80  & 30.89  & \underline{31.14}  & \textbf{35.90} && 0.9188  & 0.9323  & \underline{0.9447}  & \textbf{0.9812} && \bf 15 & \underline{34}&  124 &  48 \\
\midrule

\multirow{3}[2]{*}{\shortstack{\textit{Water}\\ \\$768\times1024\times3$}}
& 10\%  & 20.20  & 21.00  & \underline{22.57}  & \textbf{26.17} && 0.5860  & 0.6126  & \underline{0.7756}  & \textbf{0.9223} && \bf 38 & \underline{64}&  254 &  180 \\
& 20\%  & 22.75  & 23.37  & \underline{24.29}  & \textbf{28.74} && 0.7790  & 0.6685  & \underline{0.8822}  & \textbf{0.9472} && \bf 34 & \underline{60}&  256 &  103 \\
& 30\%  & 24.67  & 25.76  & \underline{25.88}  & \textbf{30.65} && 0.8767  & 0.8922  & \underline{0.9302}  & \textbf{0.9806} && \bf 31 & \underline{61}&  262 &  142 \\
\midrule

\multirow{3}[2]{*}{Average}
& 10\%  & 20.74  & 21.36  & \underline{23.85}  & \textbf{28.83} && 0.5919  & 0.5814  & \underline{0.7756}  & \textbf{0.9061} && \bf 18 & \underline{28}&  119 &  81 \\
& 20\%  & 23.95  & 24.49  & \underline{26.12}  & \textbf{31.70} && 0.7578  & 0.7380  & \underline{0.8703}  & \textbf{0.9415} && \bf 17 & \underline{29}&  122 &  68 \\
& 30\%  & 26.31  & 27.08  & \underline{27.94}  & \textbf{33.40} && 0.8509  & 0.8507  & \underline{0.9138}  & \textbf{0.9668} && \bf 16 & \underline{30}&  124 &  72 \\
\bottomrule
\end{tabular}
\label{tab:imgc1}
\end{table}

Tab.\thinspace\ref{tab:imgc1} presents the PSNR and SSIM values of recovered results  by different methods. From Tab.\thinspace\ref{tab:imgc}, it can be observed that the TNN-based method obtains better performance than SNN. Since TNN-3DTV introduces additional prior knowledge, it outperforms the TNN-based method and achieves the second-best PSNR and SSIM values. The proposed DP3LRTC obtains the highest PSNR and SSIM values. Meanwhile, it is noteworthy that the average margins between the results by DP3LRTC and TNN-3DTV are more than 3.9 dB in PSNR and 0.05 in SSIM.

\begin{figure}[htbp]
\footnotesize
\setlength{\tabcolsep}{0.9pt}\renewcommand\arraystretch{0.9}
\centering
\begin{tabular}{ccccccc}
Observed & SNN & TNN & TNN-3DTV  & DP3LRTC & Ground truth\\
\includegraphics[width=0.16\textwidth]{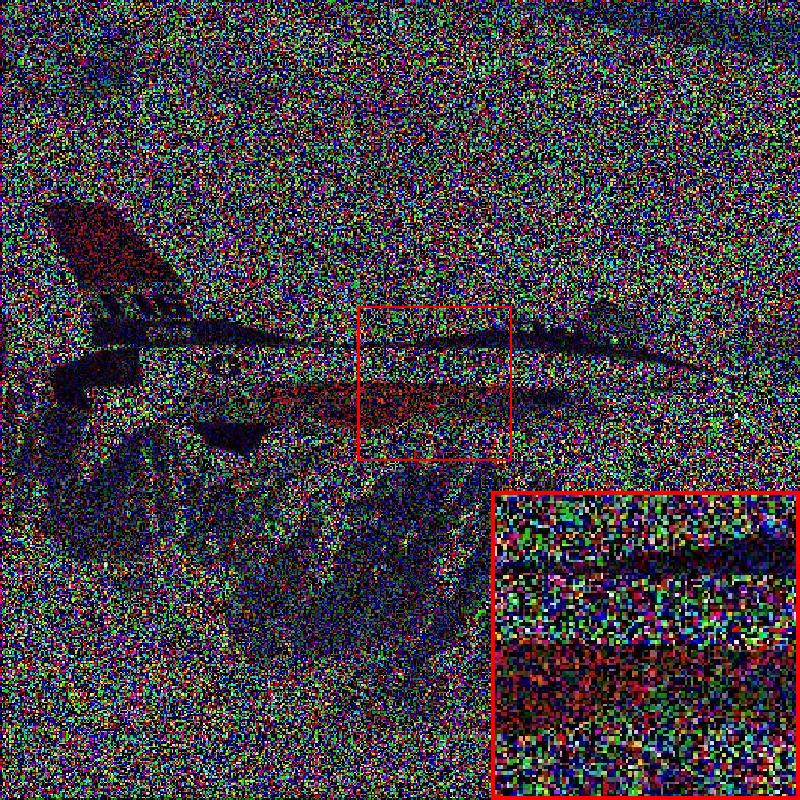}&
\includegraphics[width=0.16\textwidth]{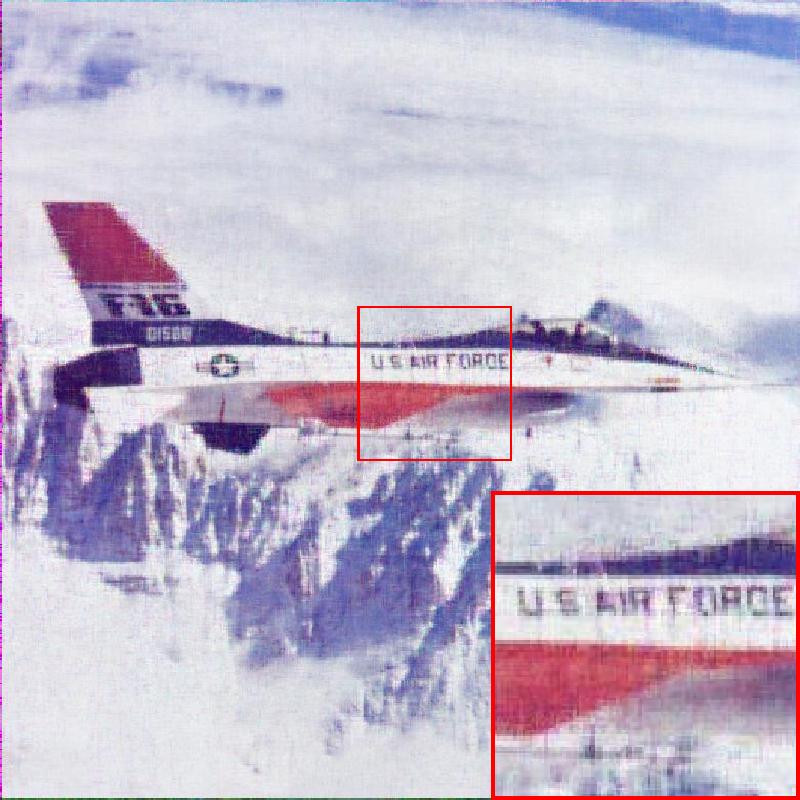}&
\includegraphics[width=0.16\textwidth]{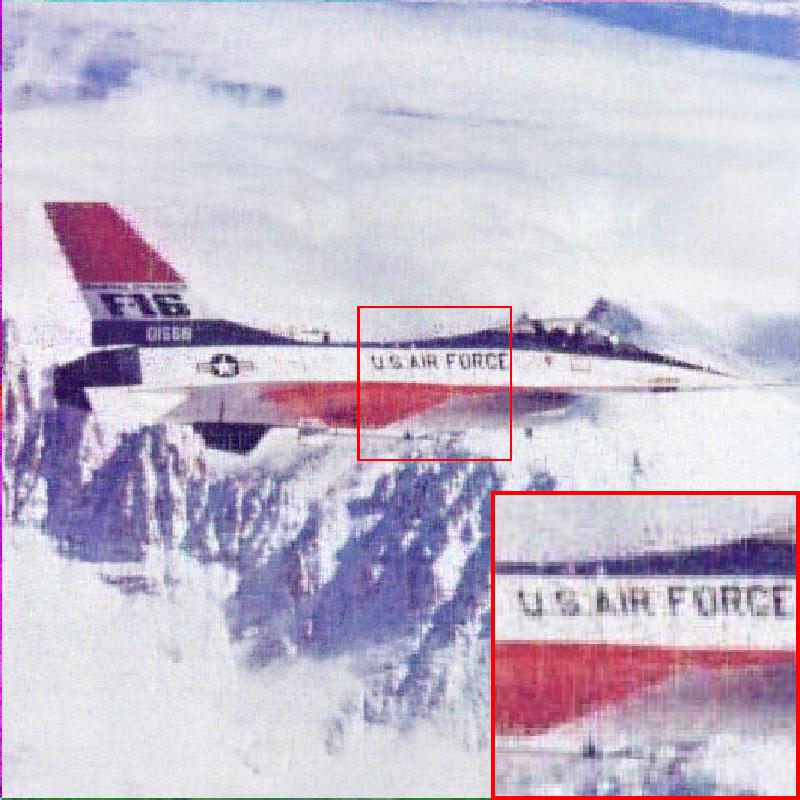}&
\includegraphics[width=0.16\textwidth]{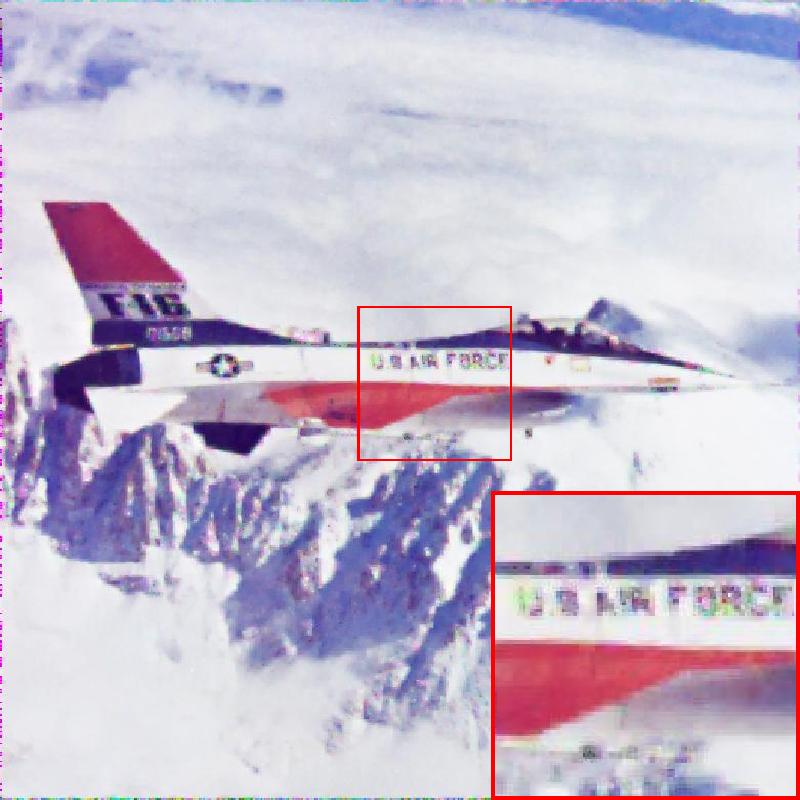}&
\includegraphics[width=0.16\textwidth]{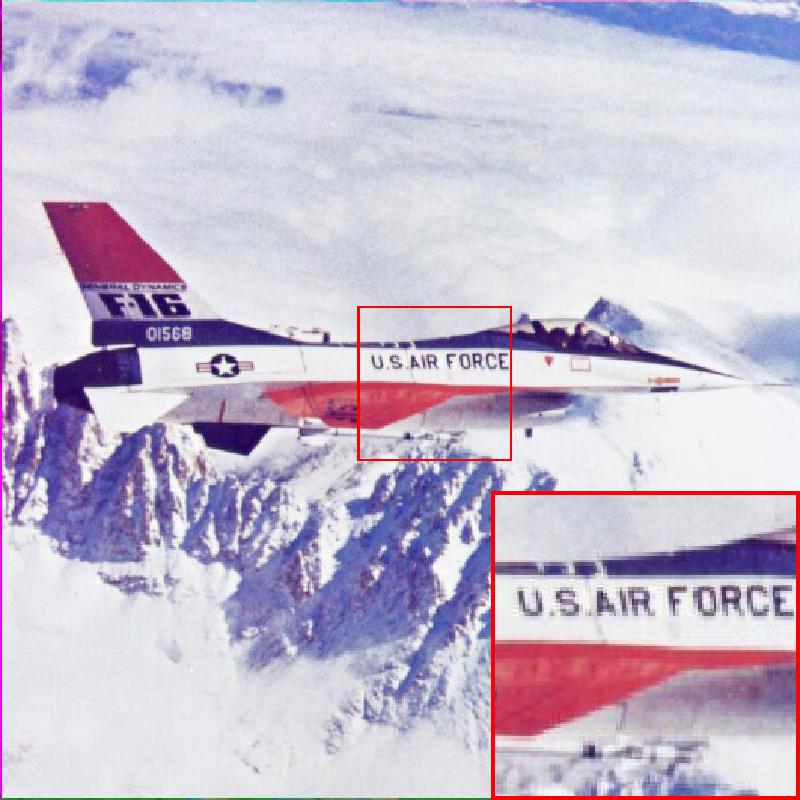}&
\includegraphics[width=0.16\textwidth]{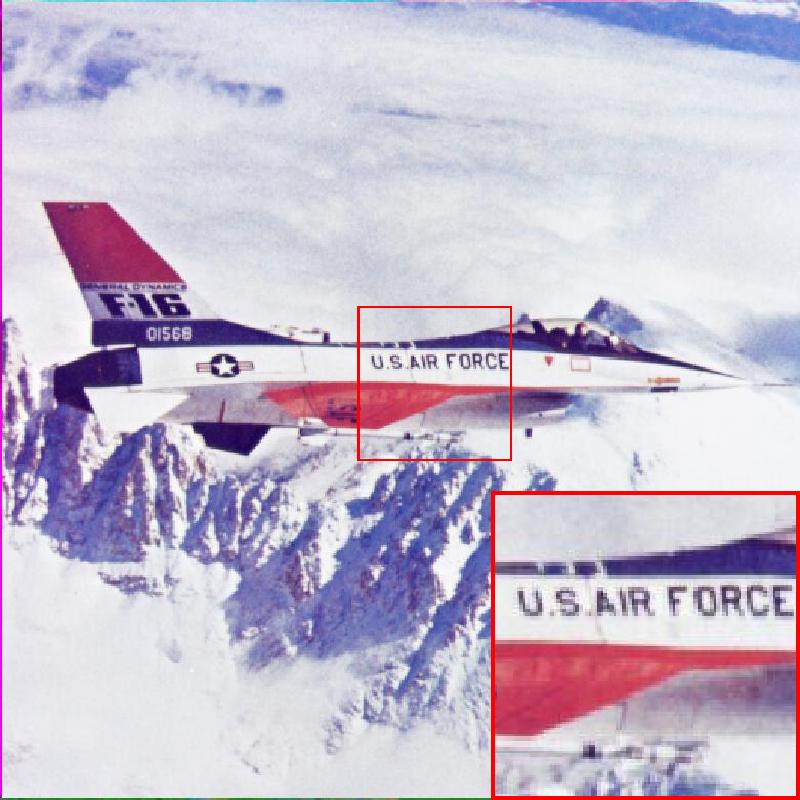}\\

\includegraphics[width=0.16\textwidth]{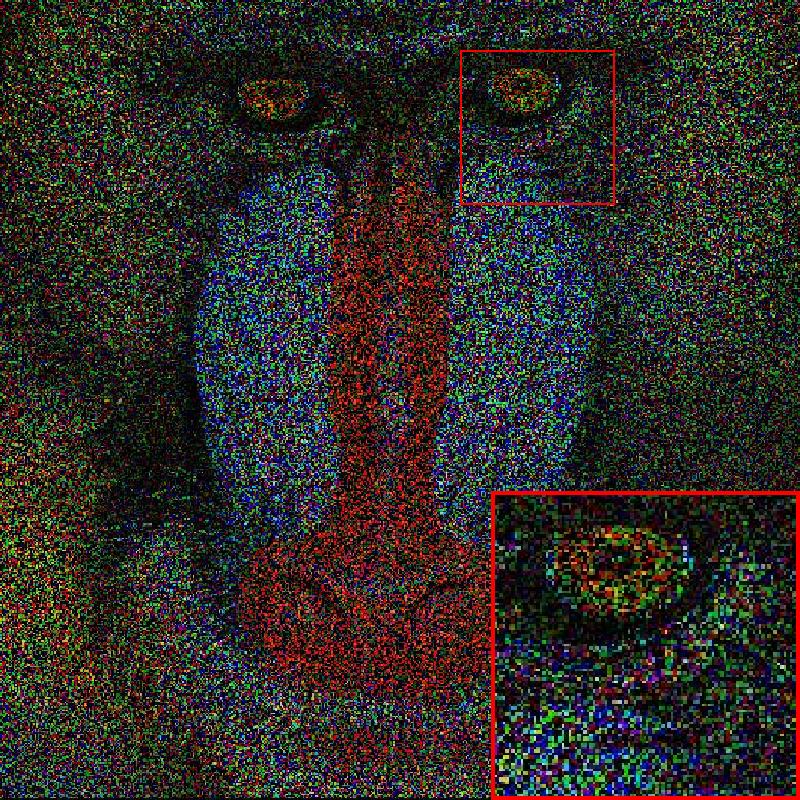}&
\includegraphics[width=0.16\textwidth]{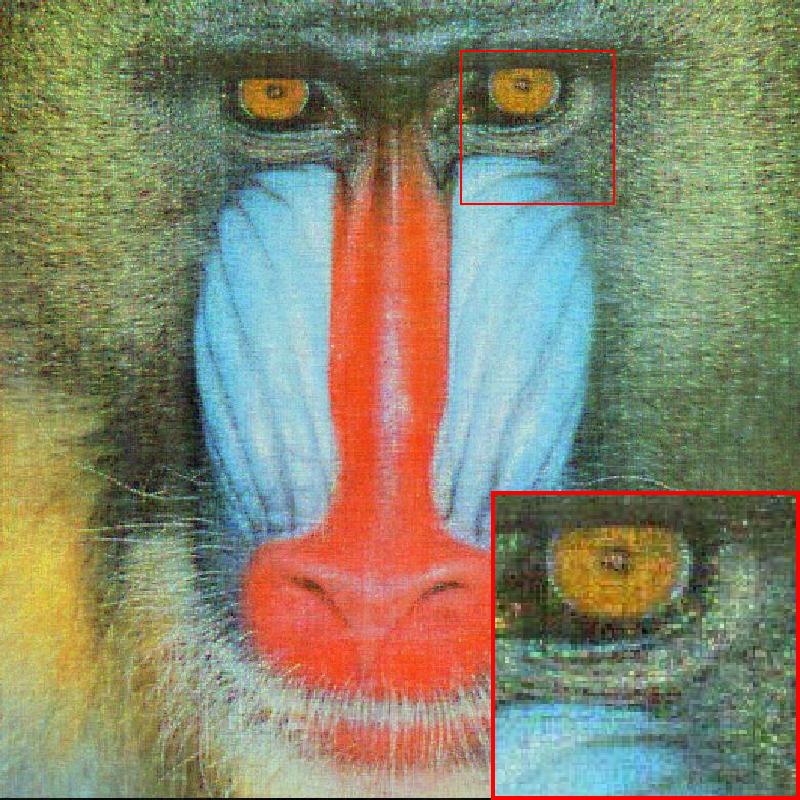}&
\includegraphics[width=0.16\textwidth]{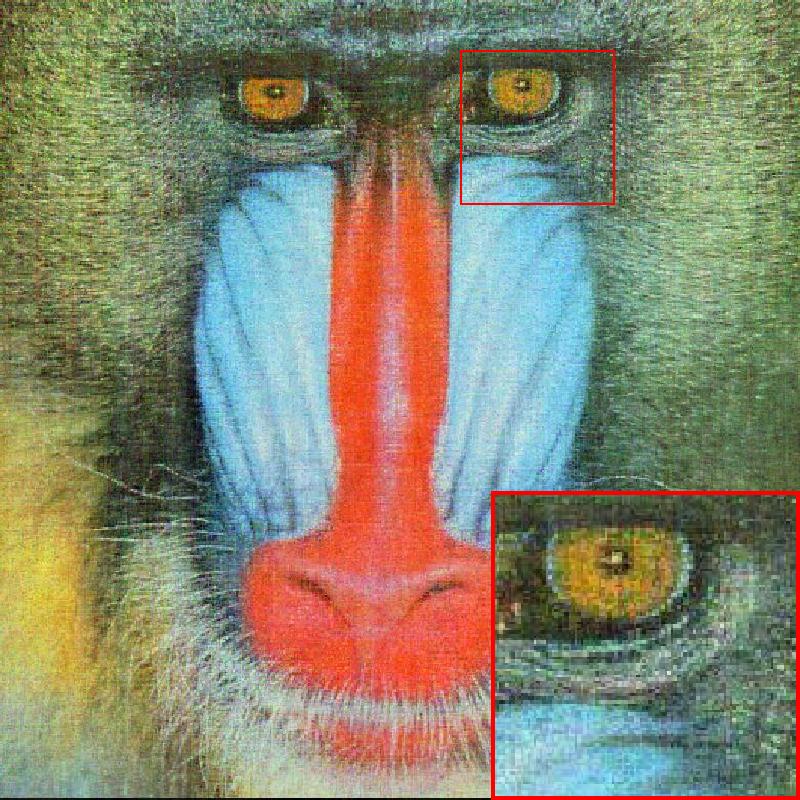}&
\includegraphics[width=0.16\textwidth]{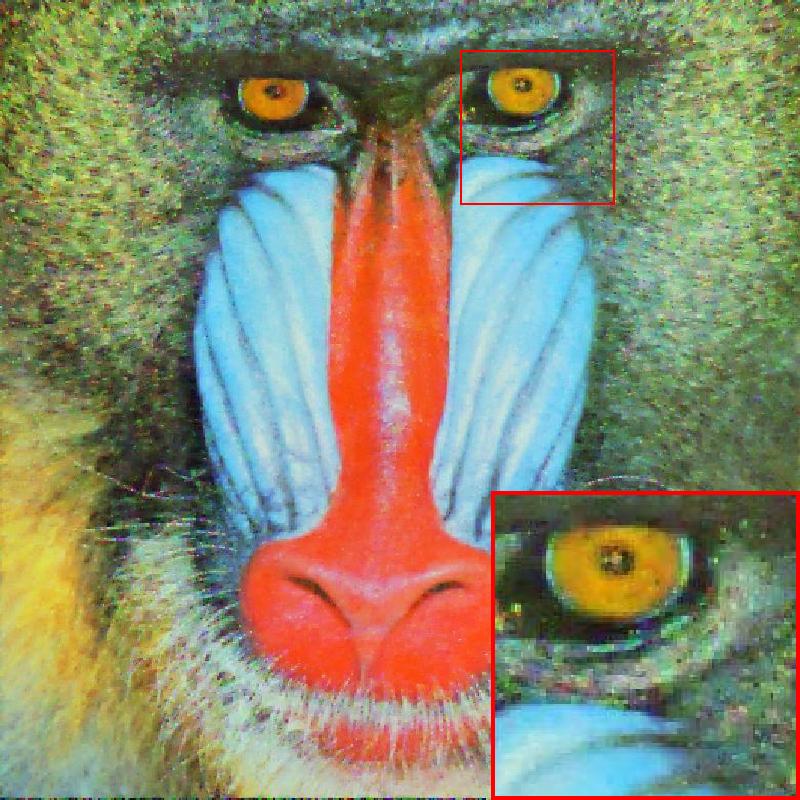}&
\includegraphics[width=0.16\textwidth]{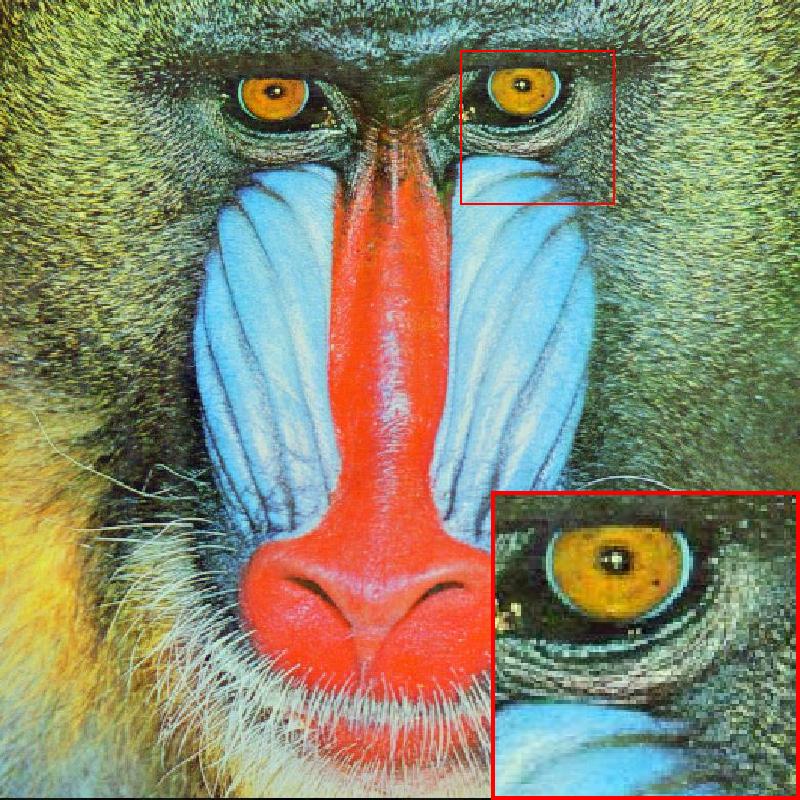}&
\includegraphics[width=0.16\textwidth]{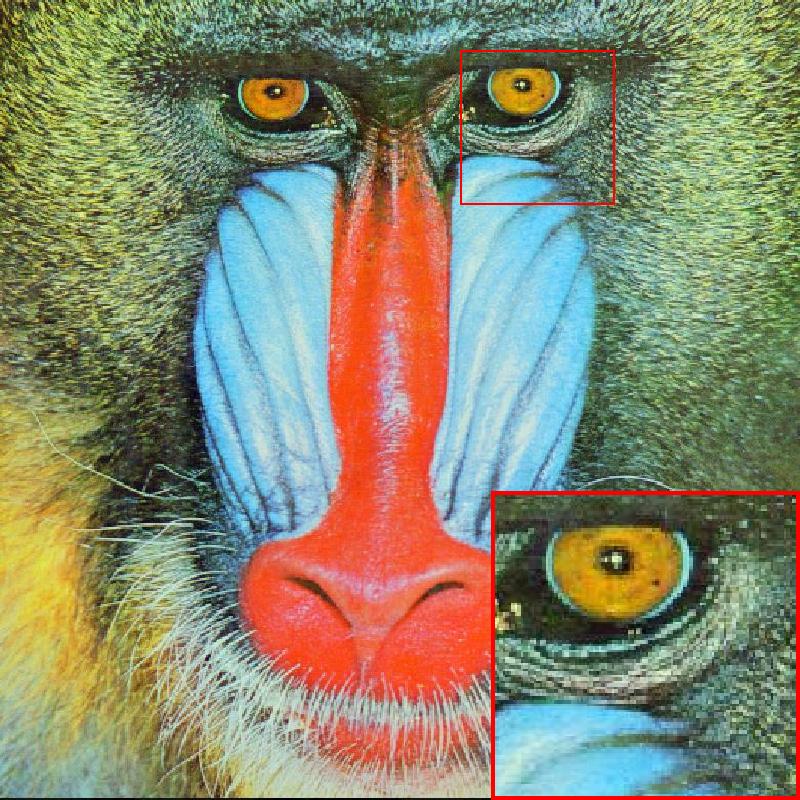}\\

\includegraphics[width=0.16\textwidth]{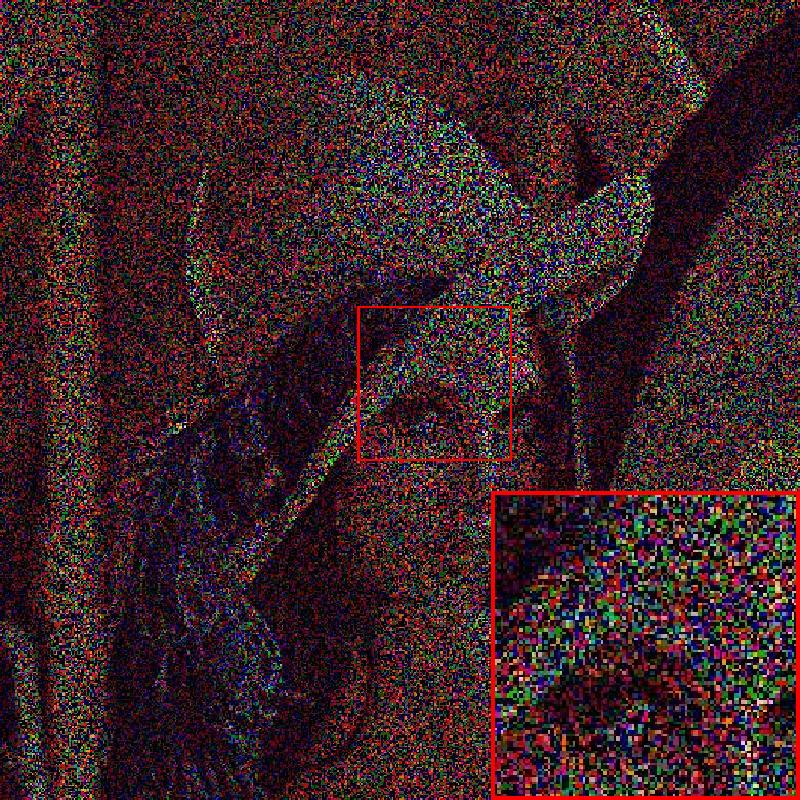}&
\includegraphics[width=0.16\textwidth]{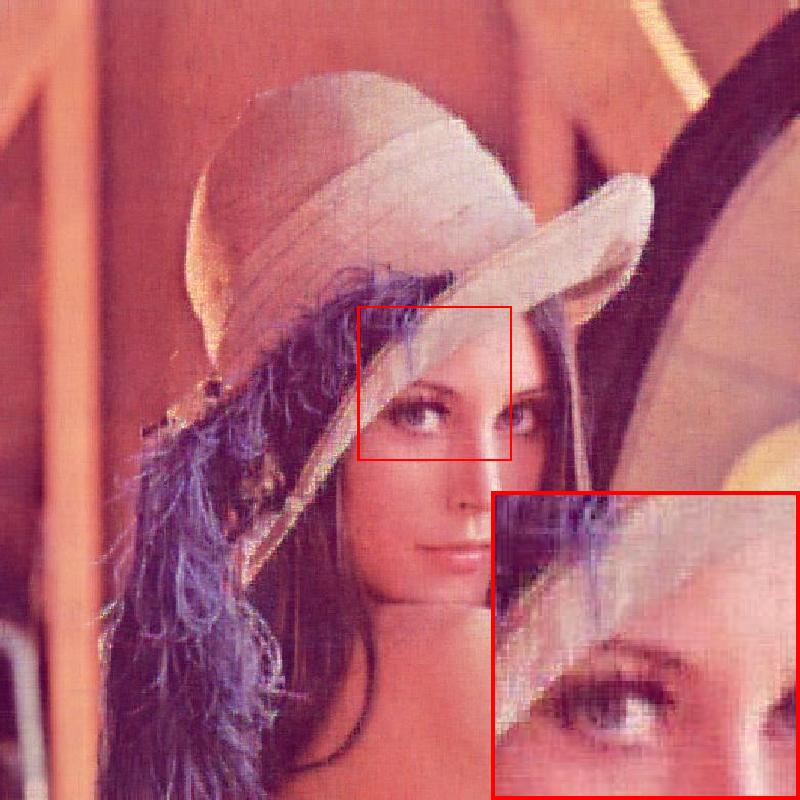}&
\includegraphics[width=0.16\textwidth]{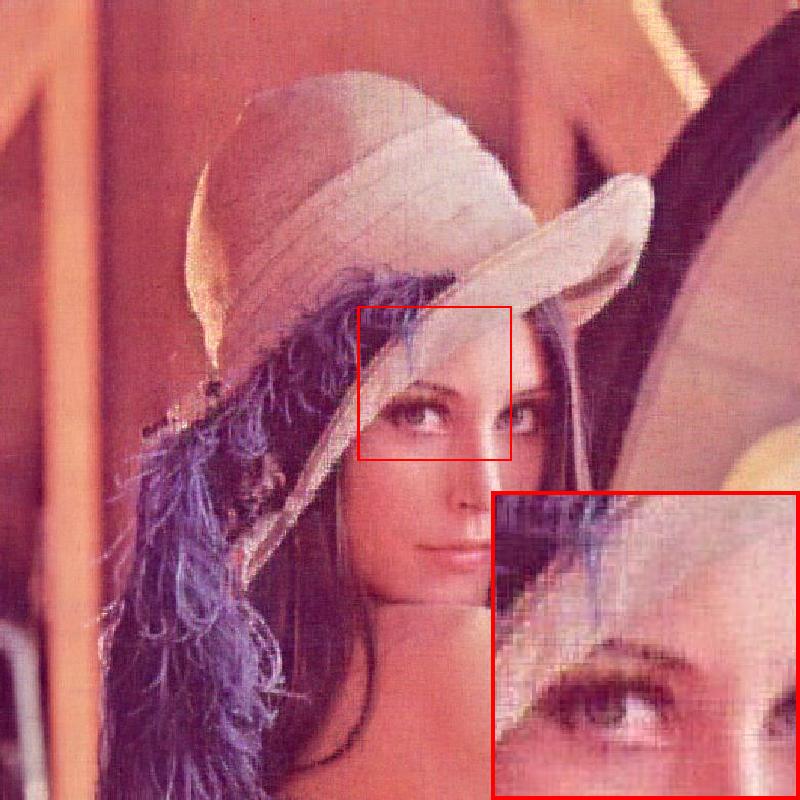}&
\includegraphics[width=0.16\textwidth]{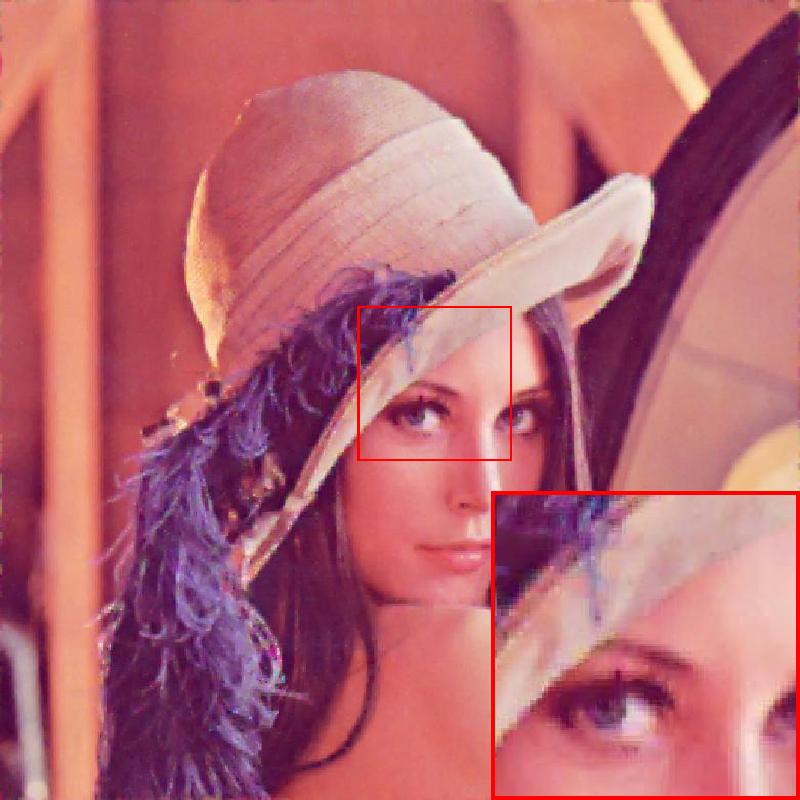}&
\includegraphics[width=0.16\textwidth]{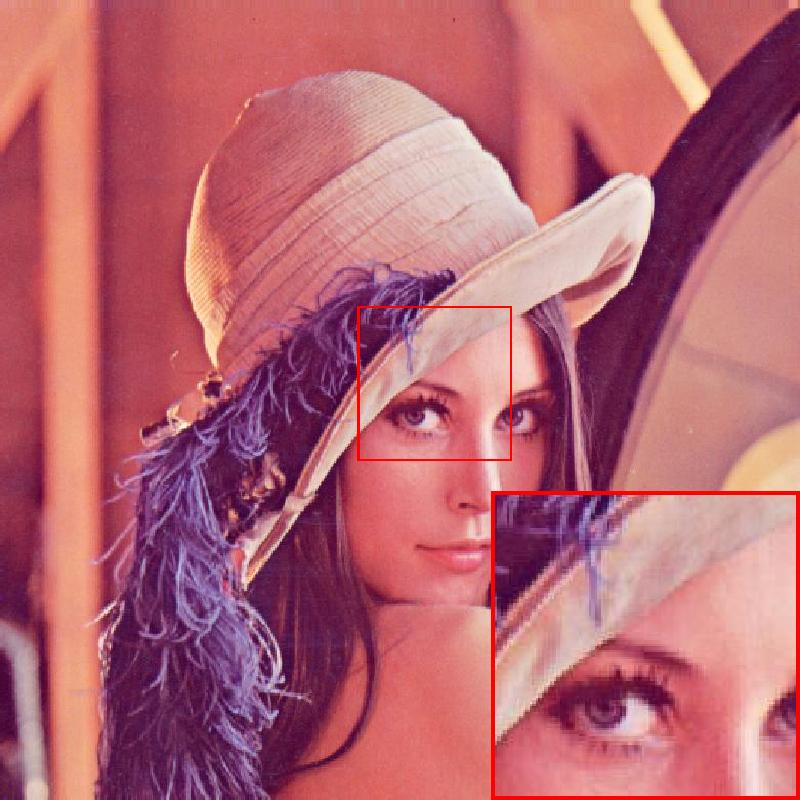}&
\includegraphics[width=0.16\textwidth]{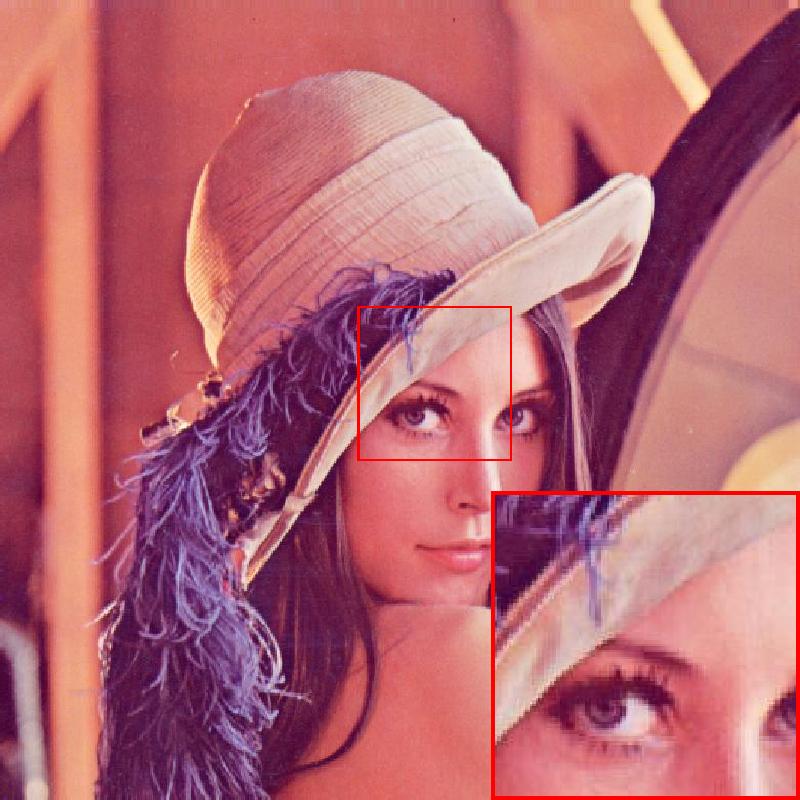}
  \end{tabular}
  \caption{The recovered color images by SNN \cite{liu2013tensor},  TNN  \cite{zhang2014novel}, TNN-3DTV \cite{jiang2018anisotropic}, and the proposed method with the sampling rate $30\%$.}
  \label{fig:result}
\end{figure}

For visualization, we exhibit the recovered images with the sampling rate $ 30\%$ in Fig.\thinspace\ref{fig:result}. From the images illustrated in Fig.\thinspace\ref{fig:result}, we can find that the SNN-based method and TNN-based method are able to recover the general structure of the color images, while the details are unclear and results remain artifacts, which can be seen in the enlarged red boxes. The results by TNN-3DTV are relatively better. However, we can observe the over-smoothed phenomena in the enlarged areas. The recovered results by the proposed method are of high quality, being highly clear and very similar to the original images.

\begin{figure}[htbp]
\footnotesize
\setlength{\tabcolsep}{1.2pt}
\centering\renewcommand\arraystretch{0.9}
\begin{tabular}{ccccccc}
Observed & SNN & TNN & TNN-3DTV  & DP3LRTC & Ground truth\\

\includegraphics[width=0.16\textwidth]{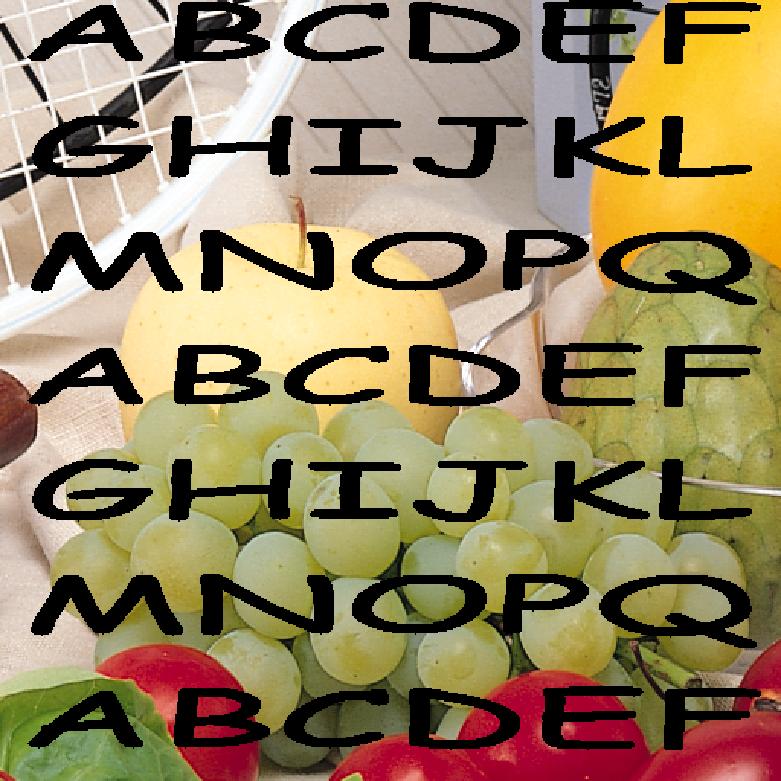}&
\includegraphics[width=0.16\textwidth]{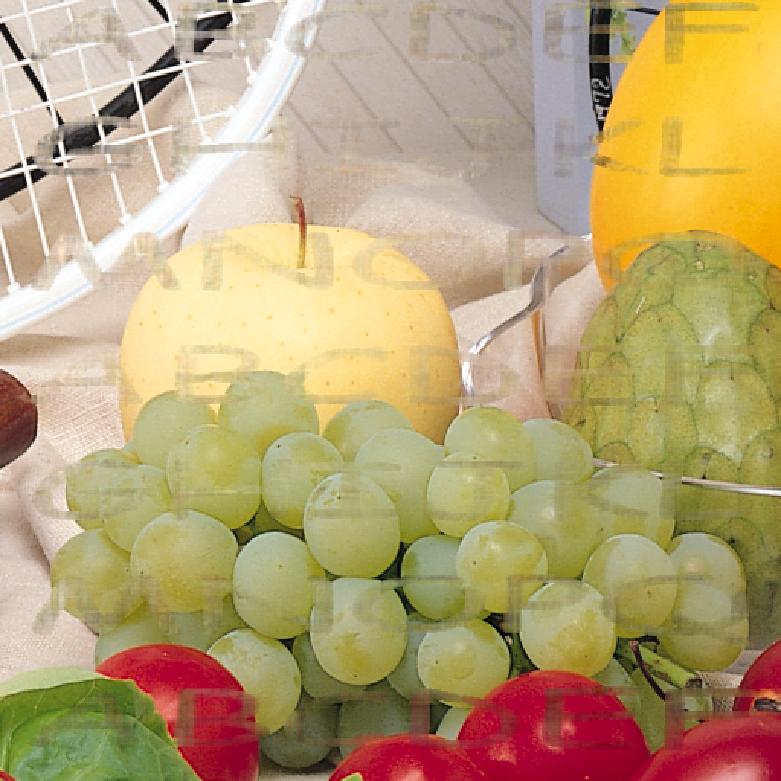}&
\includegraphics[width=0.16\textwidth]{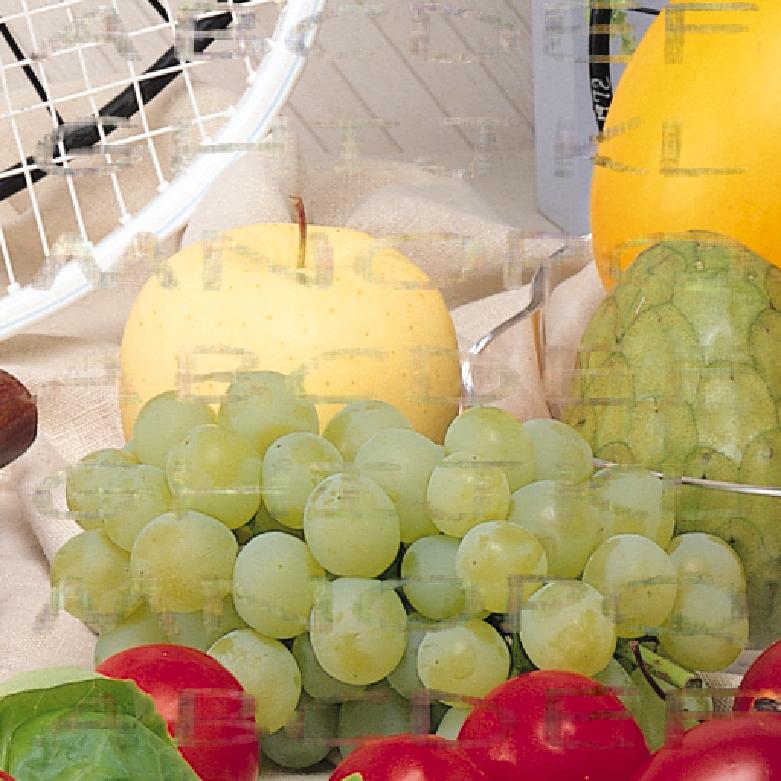}&
\includegraphics[width=0.16\textwidth]{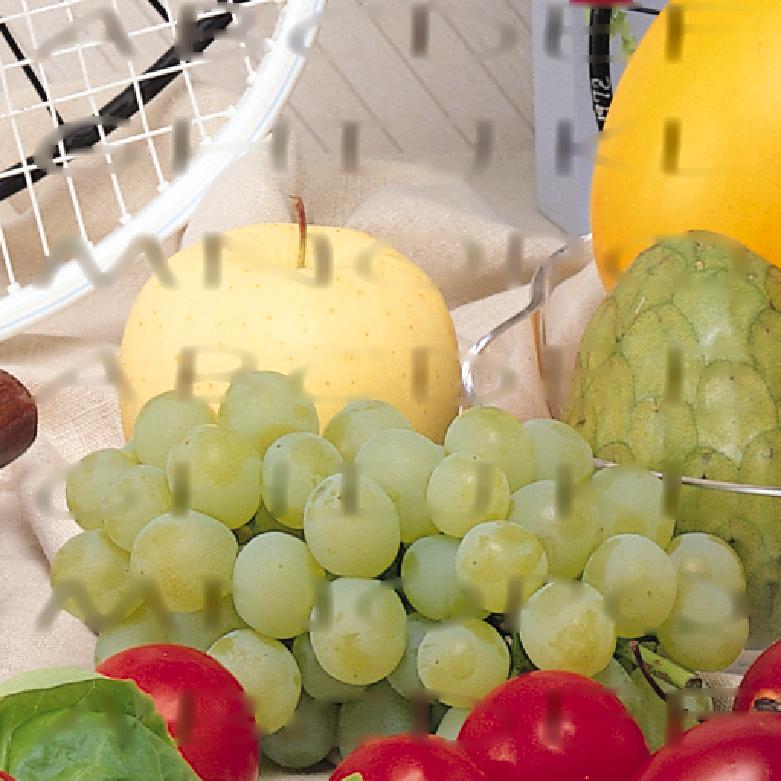}&
\includegraphics[width=0.16\textwidth]{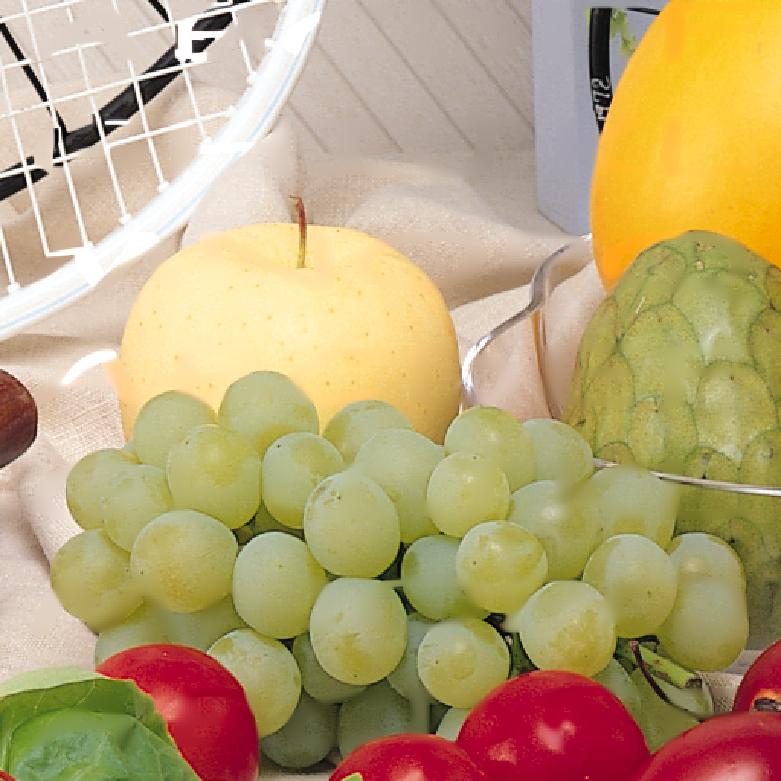}&
\includegraphics[width=0.16\textwidth]{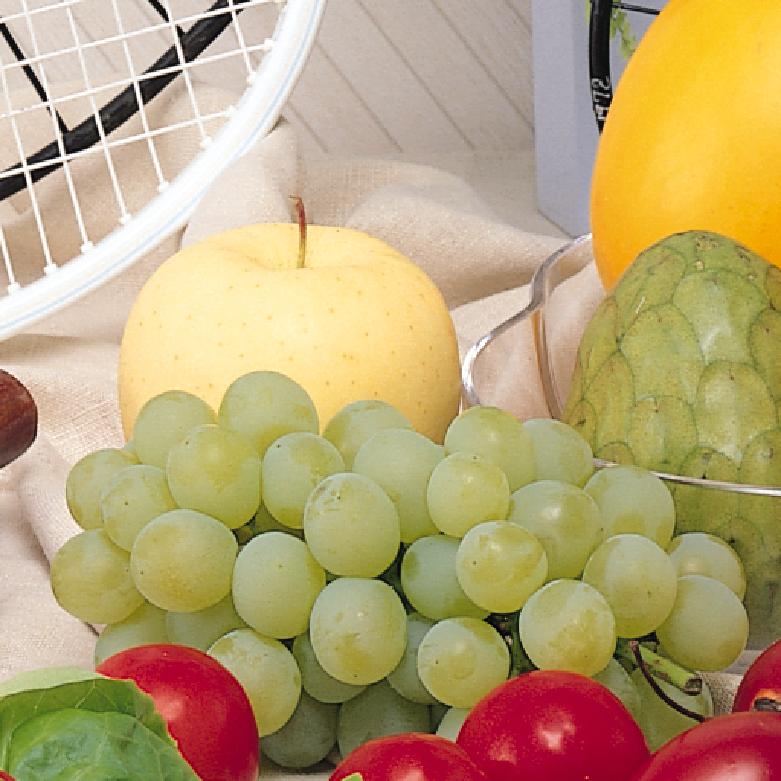}\\
 PSNR 13.17 dB & PSNR 26.58 dB & PSNR 27.24 dB & PSNR 28.68 dB  & PSNR 34.57 dB & PSNR inf \\
\includegraphics[width=0.16\textwidth]{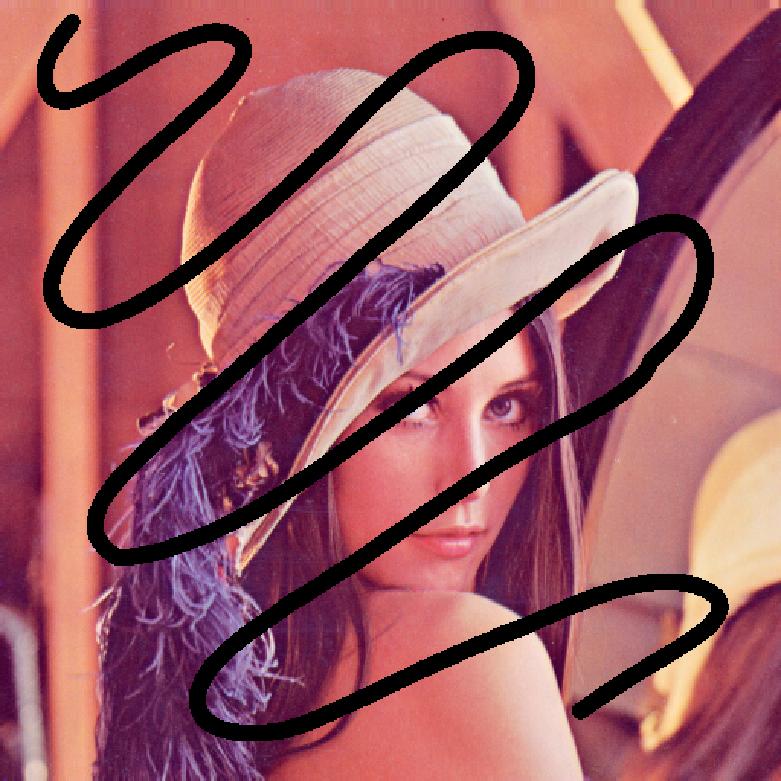}&
\includegraphics[width=0.16\textwidth]{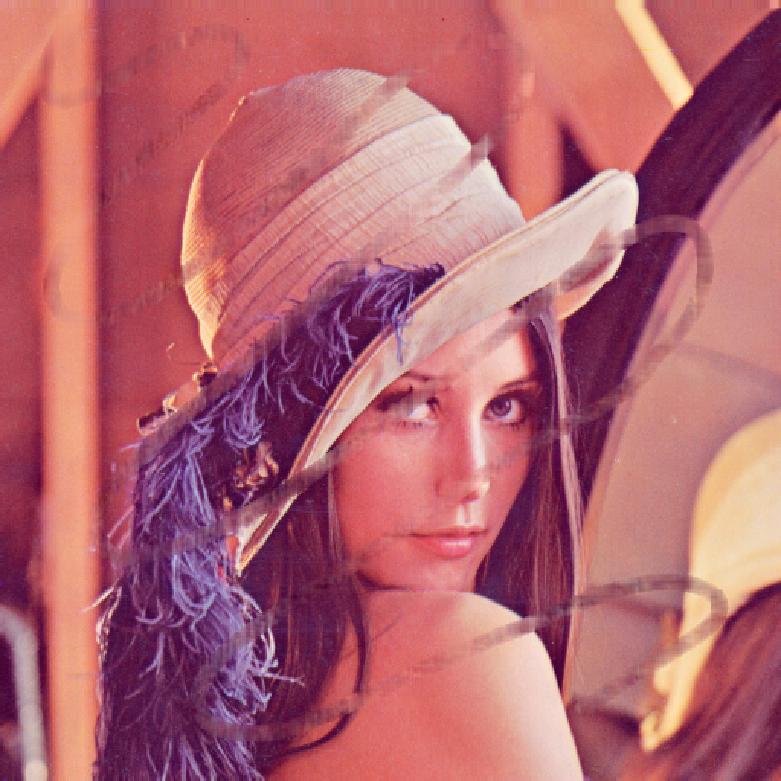}&
\includegraphics[width=0.16\textwidth]{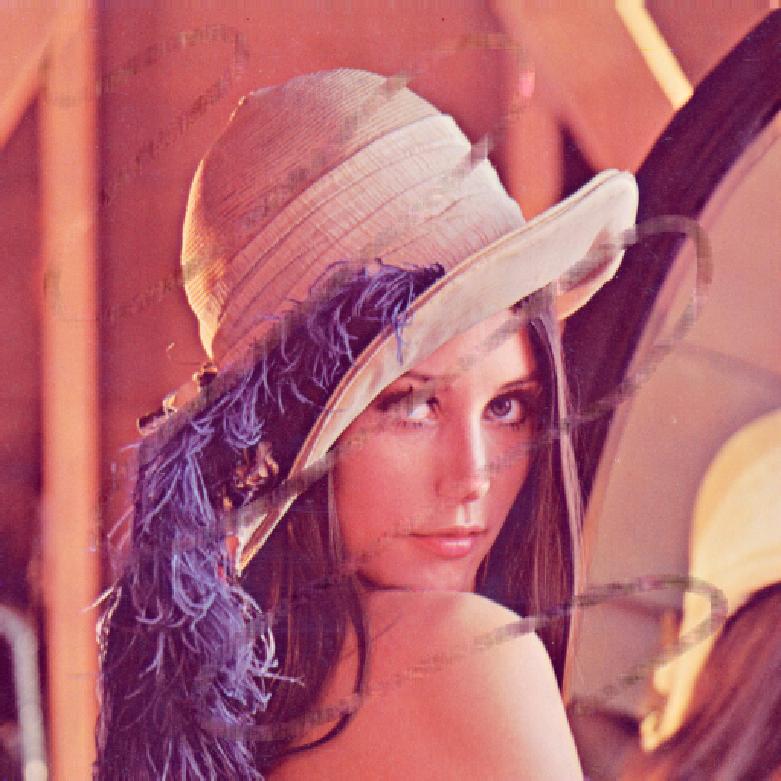}&
\includegraphics[width=0.16\textwidth]{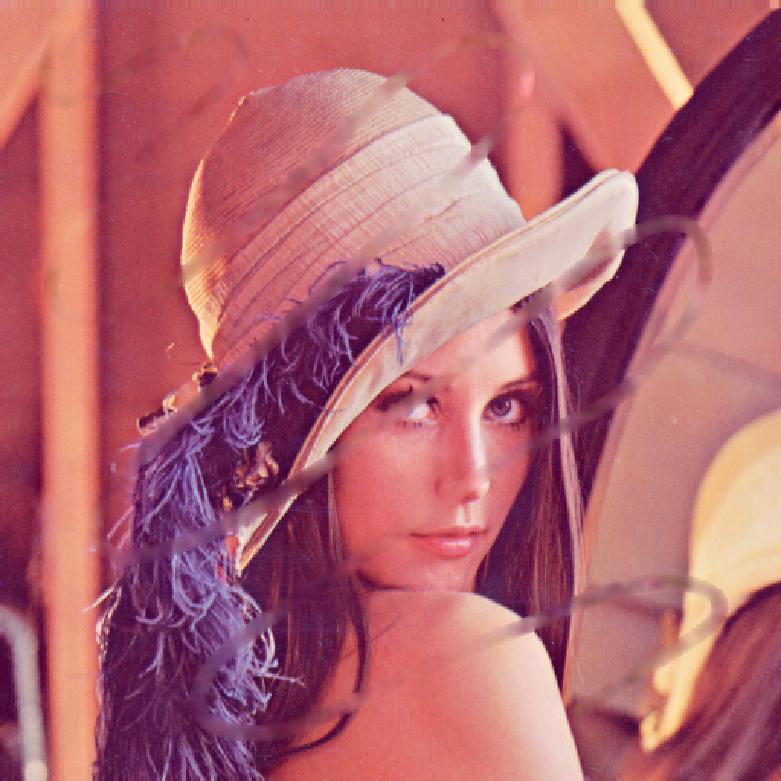}&
\includegraphics[width=0.16\textwidth]{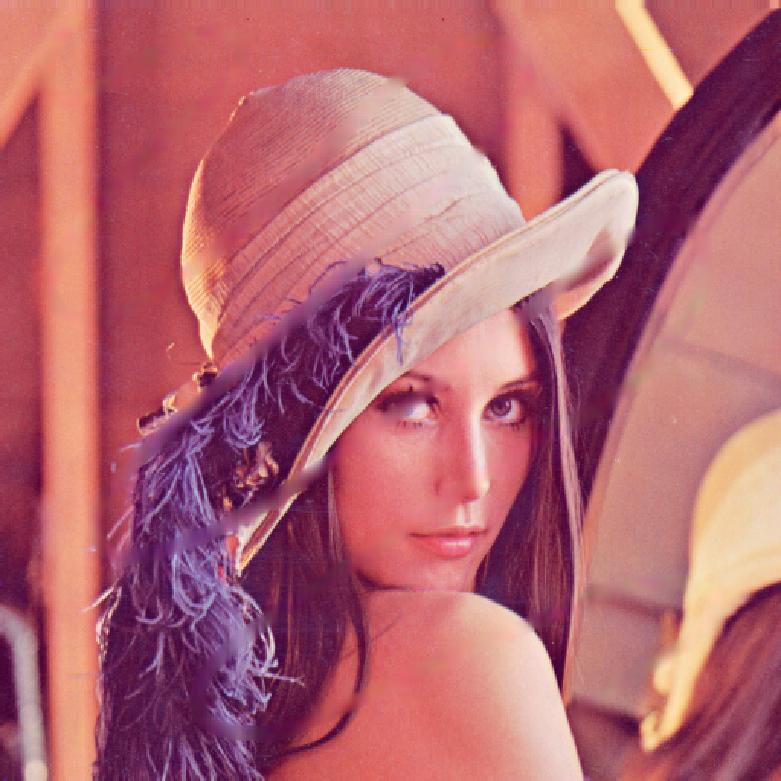}&
\includegraphics[width=0.16\textwidth]{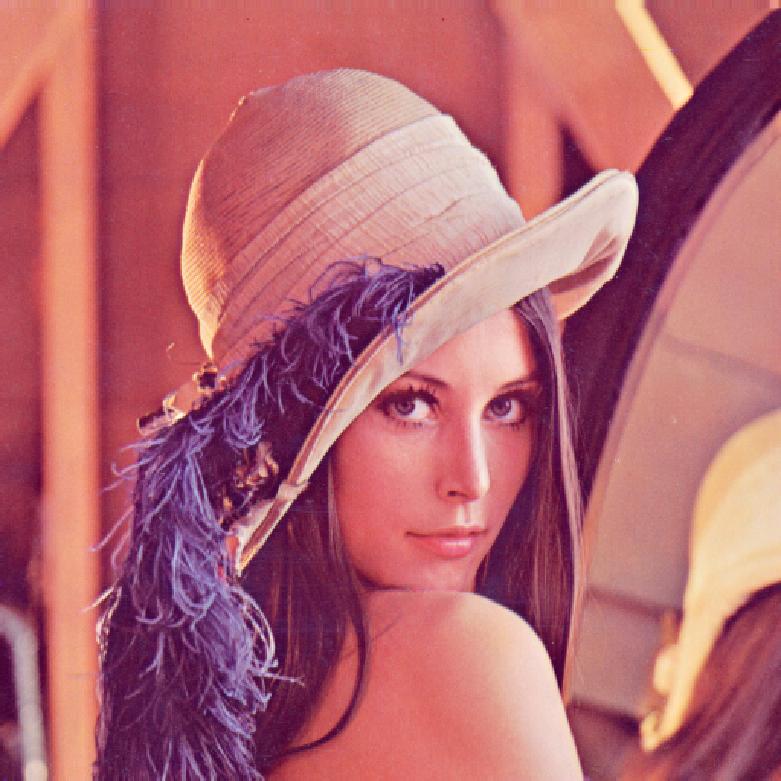}\\
 PSNR 14.78 dB & PSNR 29.21 dB & PSNR 29.12 dB & PSNR 30.18 dB  & PSNR 32.91 dB & PSNR inf  \\
\includegraphics[width=0.16\textwidth]{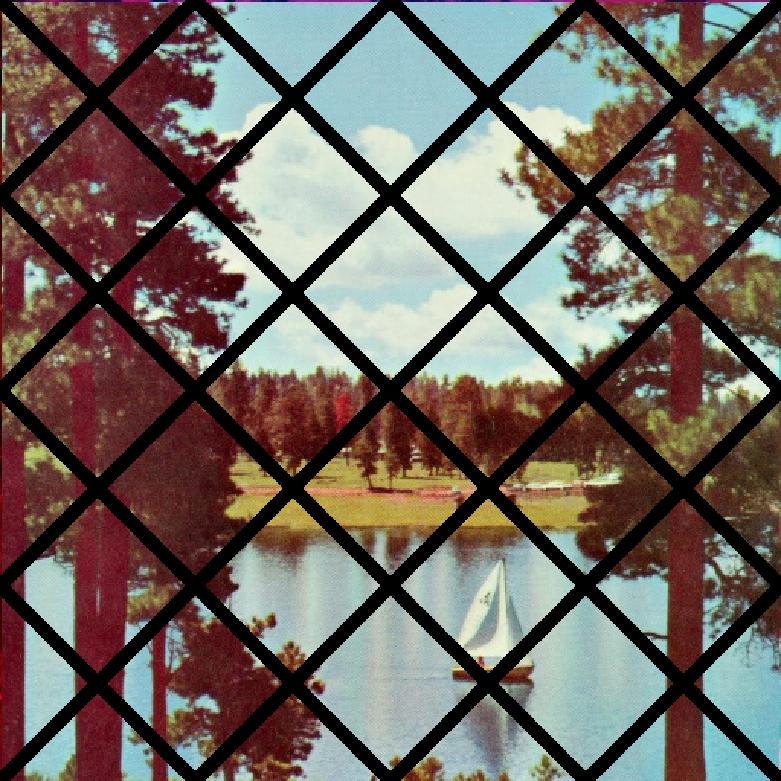}&
\includegraphics[width=0.16\textwidth]{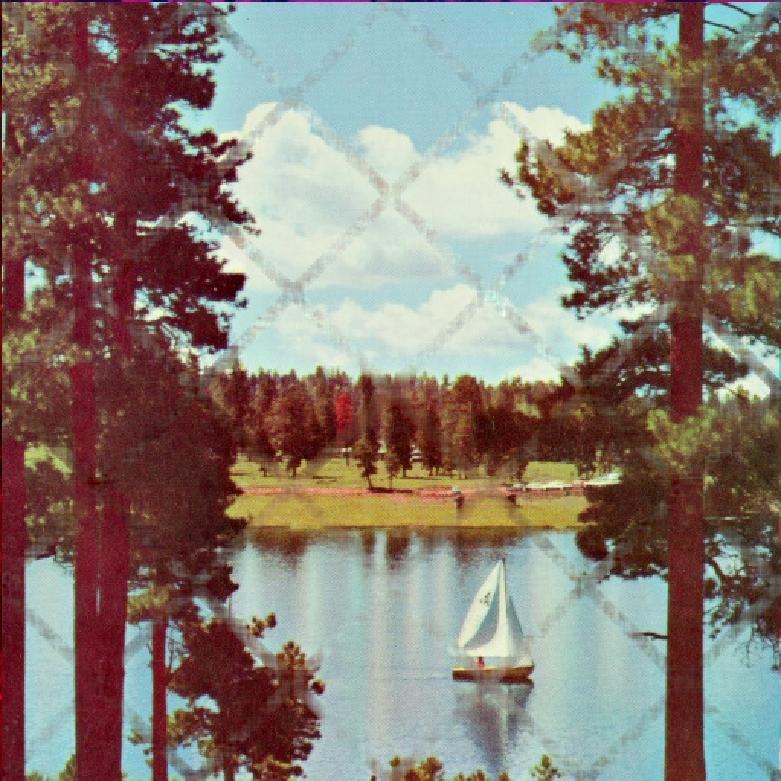}&
\includegraphics[width=0.16\textwidth]{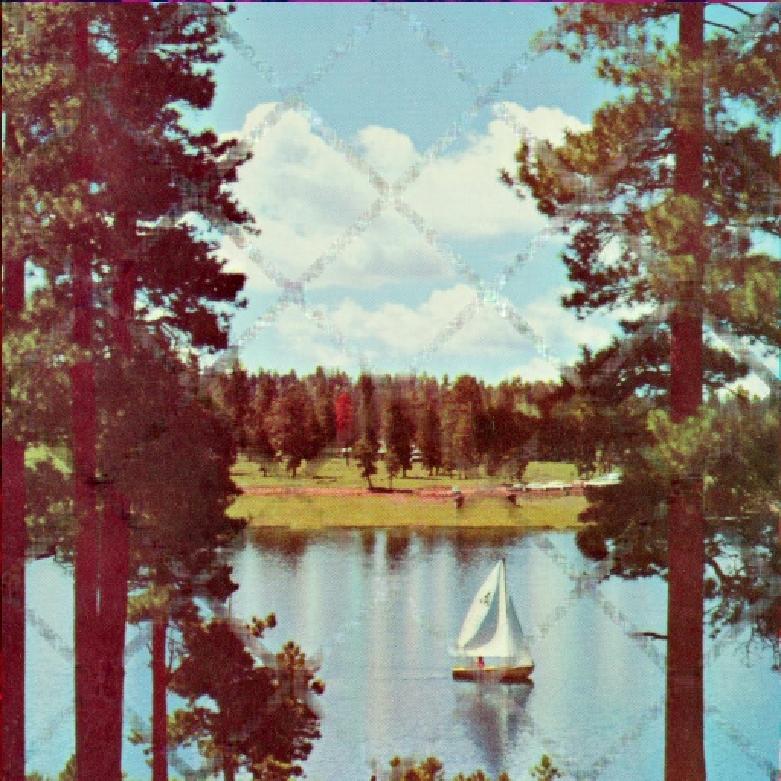}&
\includegraphics[width=0.16\textwidth]{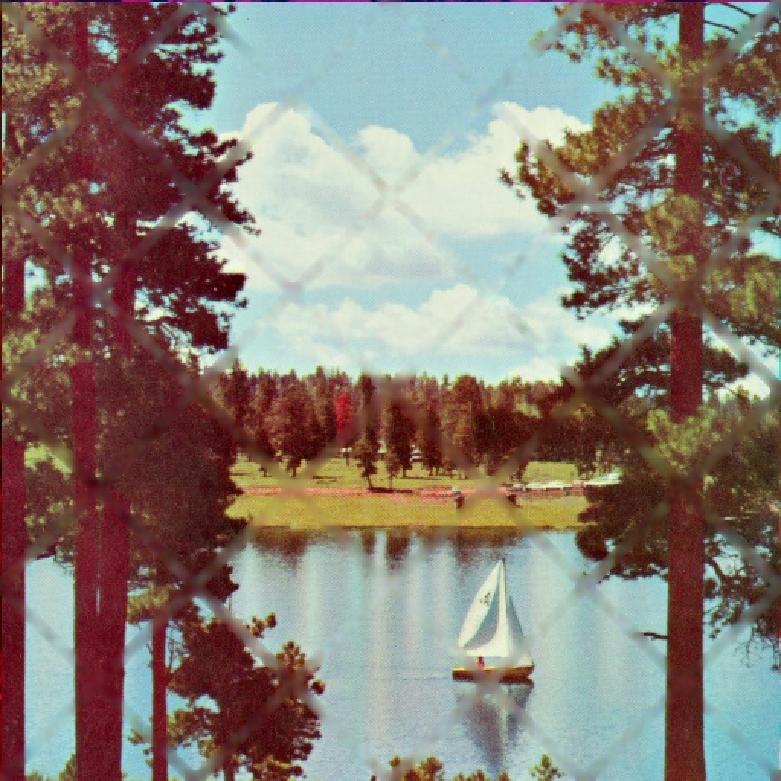}&
\includegraphics[width=0.16\textwidth]{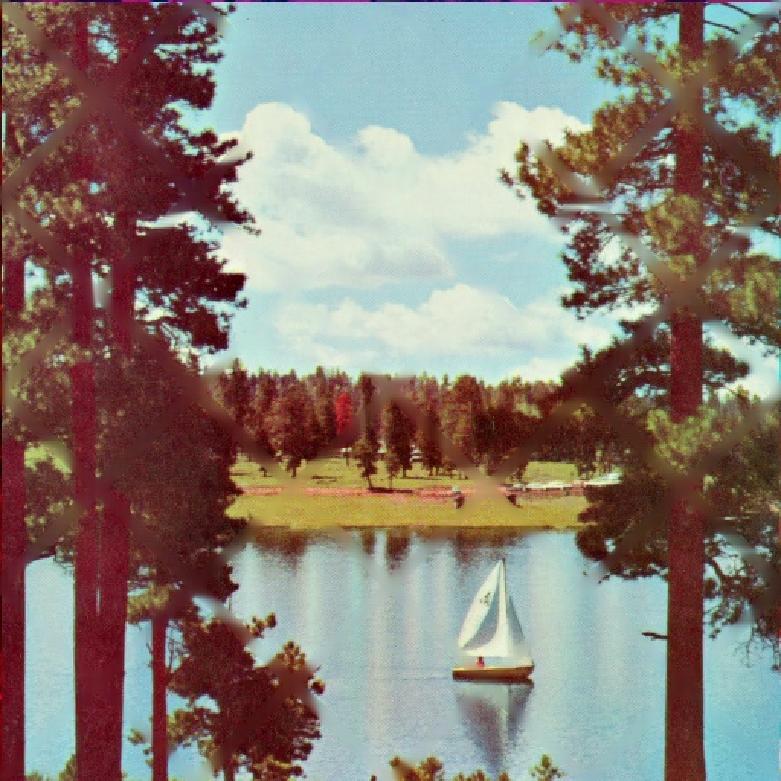}&
\includegraphics[width=0.16\textwidth]{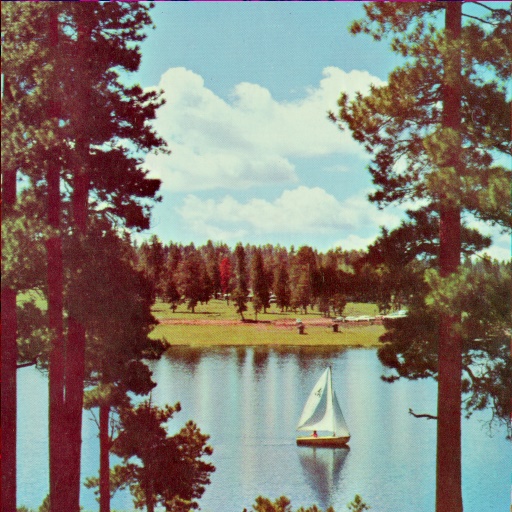}\\
  PSNR 12.18 dB & PSNR 26.03 dB & PSNR 25.83 dB & PSNR 26.99 dB  & PSNR 28.47 dB & PSNR inf \\
\includegraphics[width=0.16\textwidth]{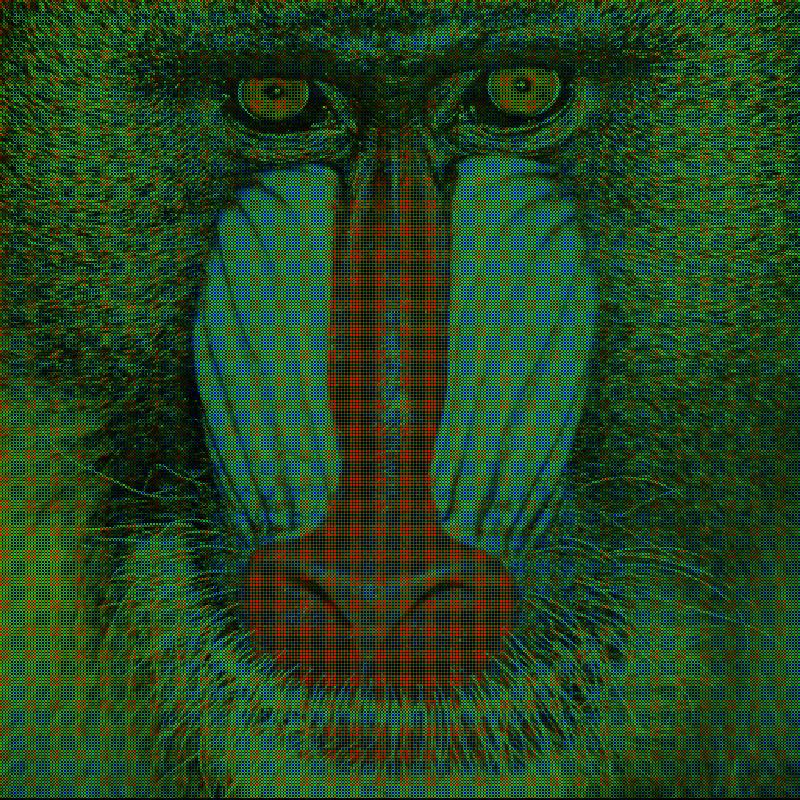}&
\includegraphics[width=0.16\textwidth]{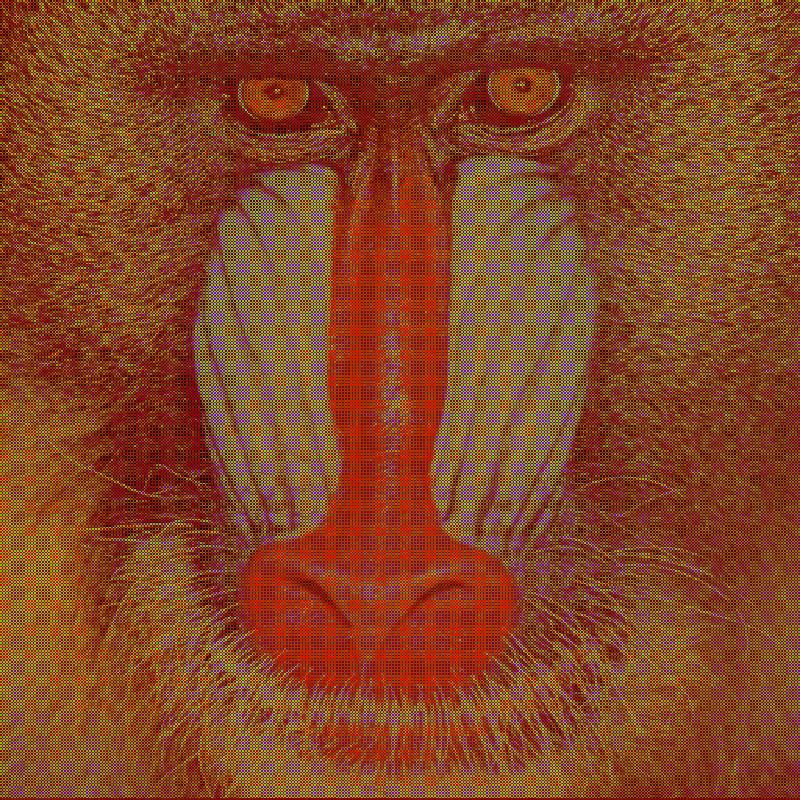}&
\includegraphics[width=0.16\textwidth]{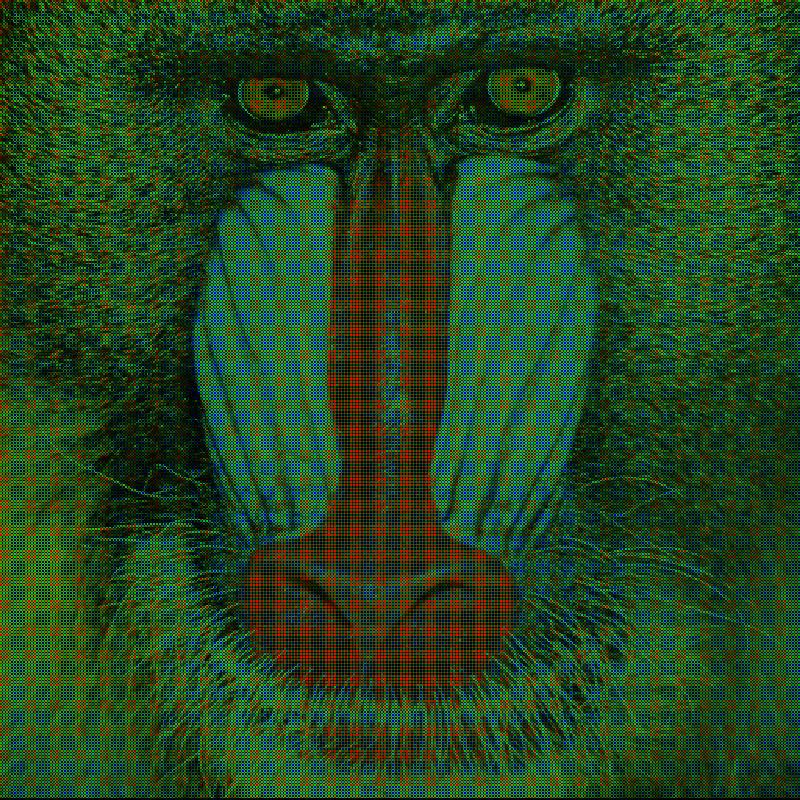}&
\includegraphics[width=0.16\textwidth]{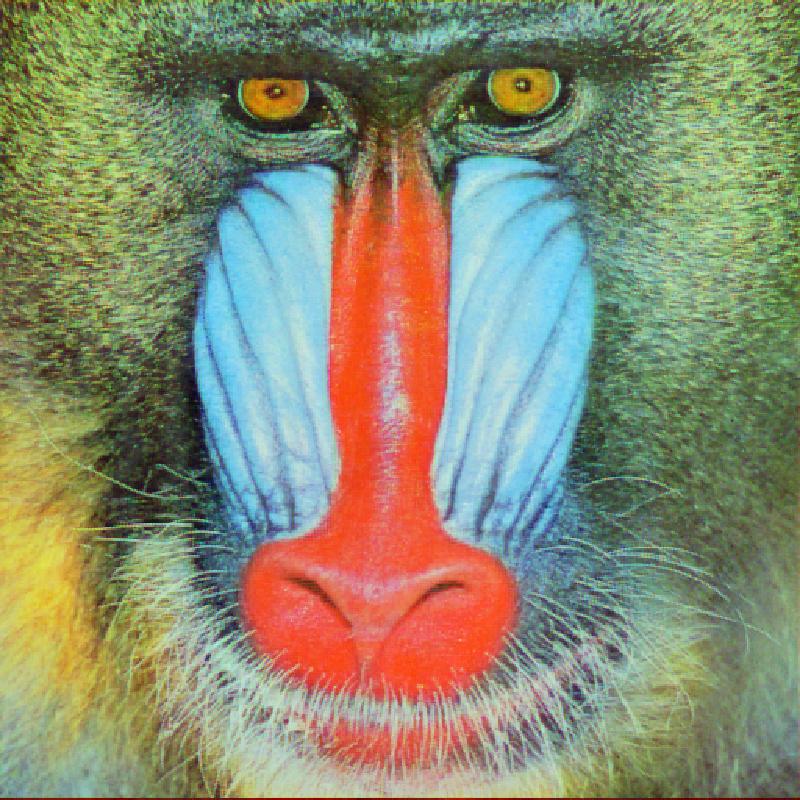}&
\includegraphics[width=0.16\textwidth]{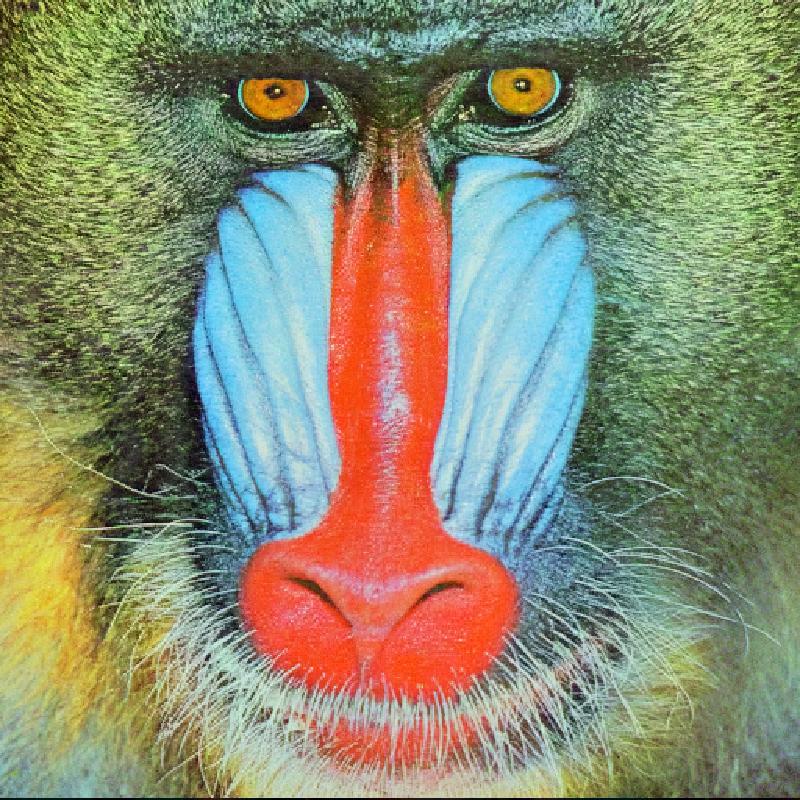}&
\includegraphics[width=0.16\textwidth]{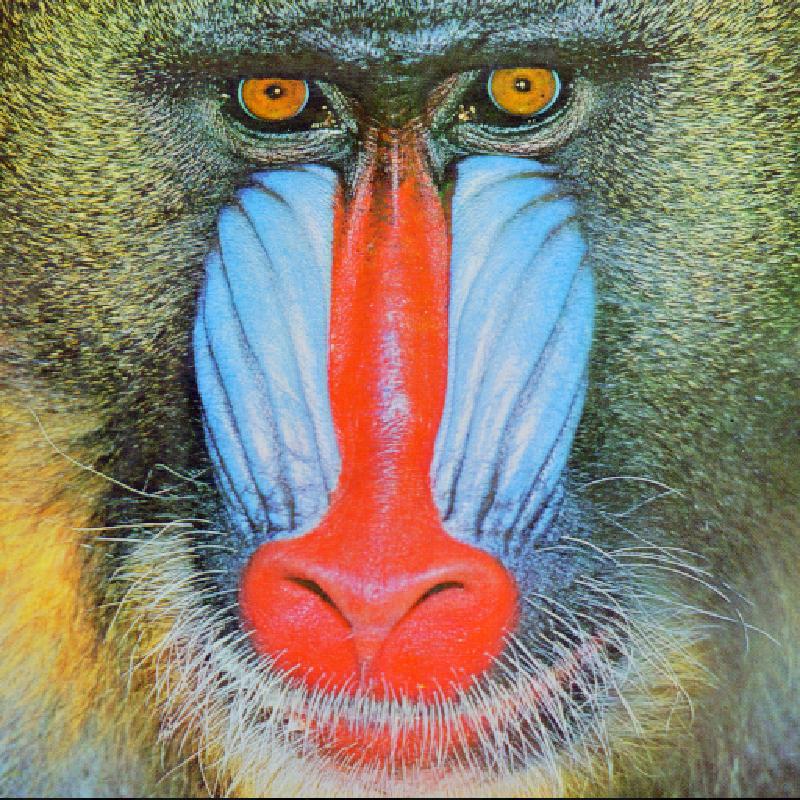}\\

PSNR 6.95 dB & PSNR 9.81 dB & PSNR 6.95 dB & PSNR 23.08 dB  & PSNR 26.23 dB & PSNR inf\\
\end{tabular}
\caption{Inpainting (top three rows) and demosaicing (bottom row) results by SNN \cite{liu2013tensor},  TNN  \cite{zhang2014novel}, TNN-3DTV \cite{jiang2018anisotropic}, and the proposed method on color images.}
  \label{fig:inpainting}
\end{figure}

Now, we consider the structured samplings, i.e., inpainting and demosaicing, which are more difficult than the  element-wise sampling. In image inpainting and demosaicing, we consider the tubal sampling (i.e.,  entries are  sampled along all channels for a spatial location) and the Bayer pattern sampling (i.e., each two-by-two cell contains two green, one blue, and one red), respectively. In inpainting, there are three different kinds of masks: letters, graffiti, and grid, respectively.
Fig.\thinspace\ref{fig:inpainting} shows the inpainting and demosaicing results by different methods on color images: \textit{Fruits}, \textit{Lena},  \textit{Sailboat},  and \textit{Baboon}.  It is easy to observe that the proposed method is superior to competing methods in  inpainting and demosaicing experiments.

\subsection{Video Completion}\label{subsec:vc}

In this subsection, 5 grayscale videos\footnote{Available at \url{http://trace.eas.asu.edu/yuv/}.} with different sizes  are selected. For gray videos,  we feed the spatial slices into  the FFDNet denoiser  trained for gray images as the
PnP denoiser.
Tab.\thinspace\ref{tab:vidc} lists the PSNR and SSIM values of the recovered video results by different methods with different sampling rates. It is easy to observe that DP3LRTC obtains the results with the highest performance evaluation indices. For visualization, we show the 20-th frame of the recovered videos with the sampling rate $20\%$ in Fig.\thinspace\ref{fig:video}. We can see that the results recovered by DP3LRTC are of higher quality than the results obtained by compared methods.
\begin{table}[htbp]
\footnotesize
\selectfont
\setlength{\tabcolsep}{1.5pt}
\renewcommand\arraystretch{0.8}
\caption{Quantitative comparison of the results by SNN \cite{liu2013tensor},  TNN  \cite{zhang2014novel}, TNN-3DTV \cite{jiang2018anisotropic}, and the proposed method on videos. The \textbf{best} and \underline{second} best values are highlighted in bold and underlined, respectively.}
\centering
\begin{tabular}{cccccccccccccccc}
\toprule
\multicolumn{1}{c}{\multirow{2}[4]{*}{Video}}& \multicolumn{1}{c}{\multirow{2}[4]{*}{SR}} & \multicolumn{4}{c}{PSNR}     & & \multicolumn{4}{c}{SSIM}& & \multicolumn{4}{c}{Time (s)}\\
\cmidrule{3-16}\multicolumn{2}{c}{} & {\scriptsize SNN}   & {\scriptsize TNN}   & {\scriptsize TNN-3DTV} & {\scriptsize  DP3LRTC} && {\scriptsize SNN}   & {\scriptsize TNN}   & {\scriptsize TNN-3DTV} & {\scriptsize DP3LRTC} && {\scriptsize SNN}   & {\scriptsize TNN}   & {\scriptsize TNN-3DTV} & {\scriptsize DP3LRTC}\\\midrule

\multirow{3}[2]{*}{\shortstack{\textit{Akiyo}\\ \\$144\times176\times30$}}
& 5\%   & 20.18  & 28.33  & \underline{28.76}  & \textbf{30.59} && 0.5976  & 0.8630  & \underline{0.8949}  & \textbf{0.9294} && \bf 5 & \underline{20}&  56 &  54 \\
& 10\%  & 23.64  & 31.16  & \underline{31.97}  & \textbf{33.33} && 0.7340  & 0.9272  & \underline{0.9452}  & \textbf{0.9623} && \bf 3 & \underline{20}&  56 &  51 \\
& 20\%  & 27.54  & 34.82  & \underline{36.06}  & \textbf{36.96} && 0.8642  & 0.9674  & \underline{0.9777}  & \textbf{0.9816} && \bf 3 & \underline{20}&  57 &  52 \\
\midrule

\multirow{3}[2]{*}{\shortstack{\textit{Suzie}\\ \\$144\times176\times30$}}
& 5\%   & 20.84  & 26.37  & \underline{27.39}  & \textbf{29.37} && 0.5944  & 0.7203  & \underline{0.7989}  & \textbf{0.8373} && \bf 4 & \underline{19}&  56 &  51 \\
& 10\%  & 24.40  & 28.41  & \underline{29.27}  & \textbf{31.80} && 0.7046  & 0.7976  & \underline{0.8513}  & \textbf{0.8899} && \bf 3 & \underline{18}&  56 &  51 \\
& 20\%  & 28.12  & 31.14  & \underline{31.95}  & \textbf{34.75} && 0.8189  & 0.8739  & \underline{0.9123}  & \textbf{0.9358} && \bf 2 & \underline{18}&  57 &  52 \\
\midrule

\multirow{3}[2]{*}{\shortstack{\textit{Container}\\ \\$144\times176\times30$}}
& 5\%   & 19.81  & 26.27  & \underline{26.63}  & \textbf{26.87} && 0.6446  & 0.8305  & \underline{0.8611}  & \textbf{0.8695} && \bf 4 & \underline{19}&  56 &  51 \\
& 10\%  & 22.26  & 29.53  & \underline{30.05}  & \textbf{30.44} && 0.7413  & 0.9023  & \underline{0.9272}  & \textbf{0.9230} && \bf 3 & \underline{19}&  56 &  51 \\
& 20\%  & 25.56  & 33.65  & \underline{34.69}  & \textbf{34.82} && 0.8495  & 0.9553  & \underline{0.9677}  & \textbf{0.9654} && \bf 2 & \underline{19}&  57 &  52 \\
\midrule

\multirow{3}[2]{*}{\shortstack{\textit{News}\\ \\$144\times176\times30$}}
& 5\%   & 18.34  & 26.65  & \underline{27.20}  & \textbf{28.29} && 0.5610  & 0.8182  & \underline{0.8738}  & \textbf{0.9023} && \bf 4 & \underline{20}&  56 &  51 \\
& 10\%  & 21.47  & 30.31  & \underline{30.56}  & \textbf{31.96} && 0.6955  & 0.9106  & \underline{0.9285}  & \textbf{0.9473} && \bf 3 & \underline{20}&  57 &  51 \\
& 20\%  & 25.00  & 33.82  & \underline{34.21}  & \textbf{35.55} && 0.8256  & 0.9554  & \underline{0.9665}  & \textbf{0.9740} && \bf 3 & \underline{21}&  57 &  52 \\
\midrule

\multirow{3}[2]{*}{\shortstack{\textit{Bus}\\ \\$256\times256\times30$}}
& 5\%   & 15.79  & 18.30  & \underline{18.36}  & \textbf{20.78} && 0.3300  & 0.3331  & \underline{0.3973}  & \textbf{0.5932} && \bf 11 & \underline{52}&  185 &  145 \\
& 10\%  & 17.70  & 19.61  & \underline{19.66}  & \textbf{22.83} && 0.4174  & 0.4421  & \underline{0.5083}  & \textbf{0.7222} && \bf 9 & \underline{50}&  185 &  146 \\
& 20\%  & 19.86  & 21.74  & \underline{21.91}  & \textbf{25.48} && 0.5560  & 0.5993  & \underline{0.6778}  & \textbf{0.8359} && \bf 5 & \underline{50}&  187 &  148 \\
\midrule

\multirow{3}[2]{*}{Average}
& 5\%   & 18.99  & 25.18  & \underline{25.67}  & \textbf{27.18} && 0.5455  & 0.7130  & \underline{0.7652}  & \textbf{0.8263} && \bf 6 & \underline{26}&  82 &  70 \\
& 10\%  & 21.89  & 27.80  & \underline{28.30}  & \textbf{30.07} && 0.6586  & 0.7960  & \underline{0.8321}  & \textbf{0.8889} && \bf 4 & \underline{25}&  82 &  70 \\
& 20\%  & 25.22  & 31.03  & \underline{31.76}  & \textbf{33.51} && 0.7828  & 0.8703  & \underline{0.9004}  & \textbf{0.9385} && \bf 3 & \underline{26}&  83 &  71 \\
\bottomrule
\end{tabular}%
\label{tab:vidc}
\end{table}

\begin{figure*}[htbp]
\footnotesize
\setlength{\tabcolsep}{0.9pt}
\centering
\begin{tabular}{cccccc}
Observed & SNN & TNN & TNN-3DTV & DP3LRTC & Ground truth\\

\includegraphics[width=0.16\textwidth]{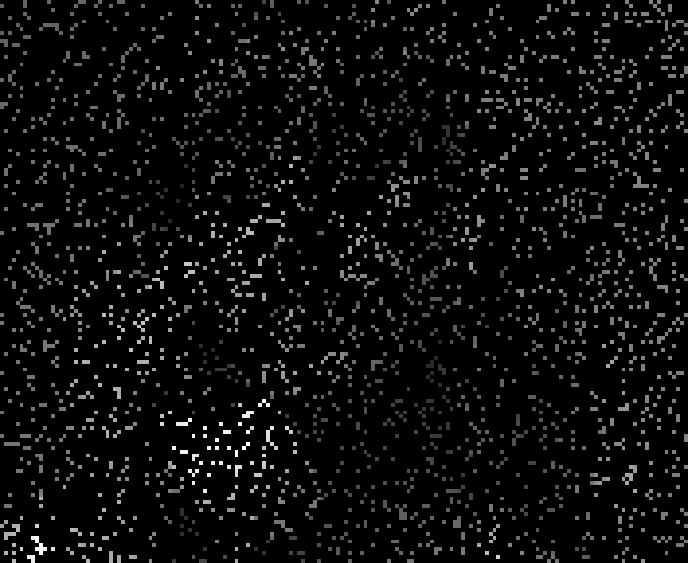}&
\includegraphics[width=0.16\textwidth]{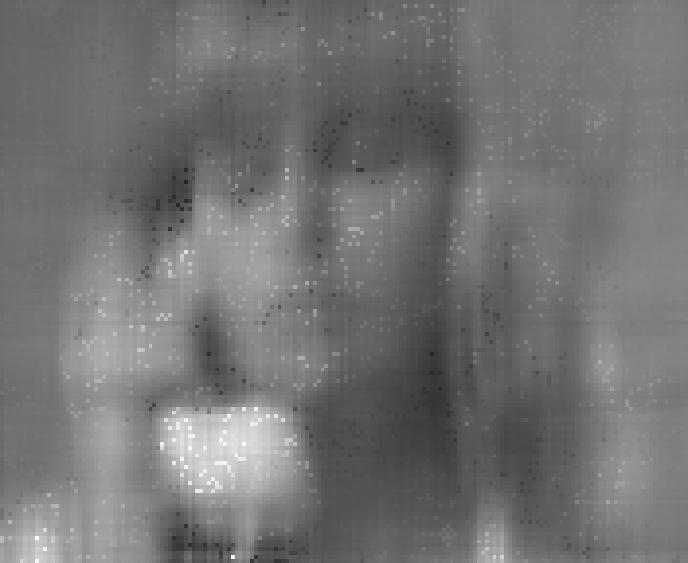}&
\includegraphics[width=0.16\textwidth]{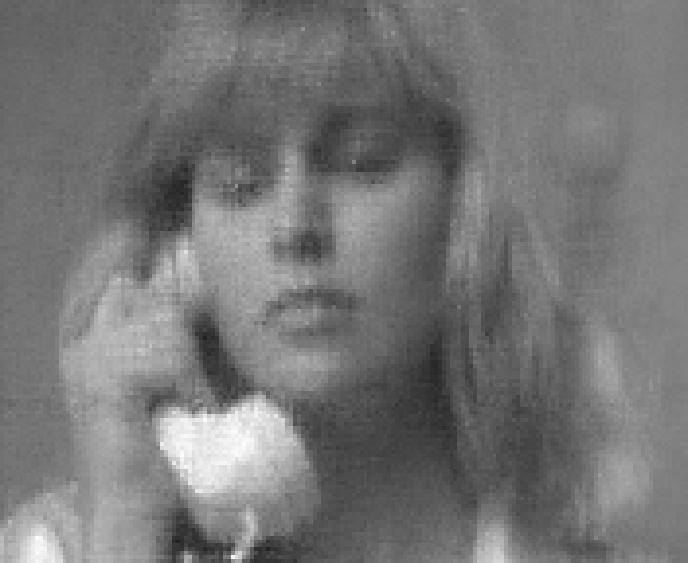}&
\includegraphics[width=0.16\textwidth]{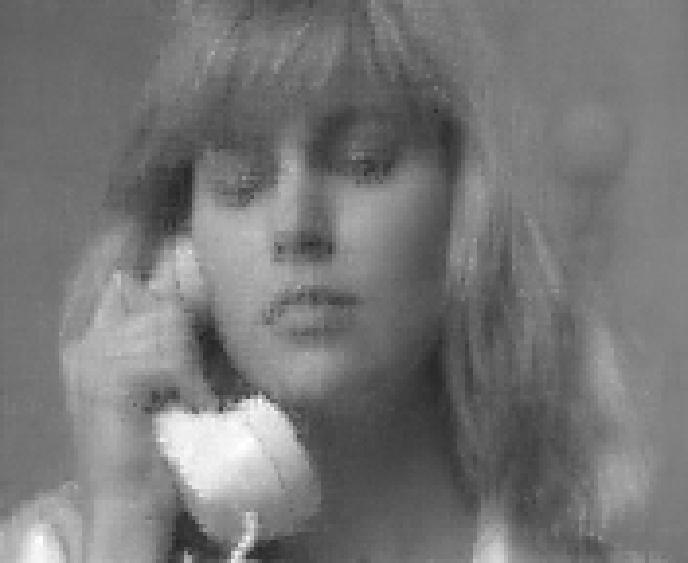}&
\includegraphics[width=0.16\textwidth]{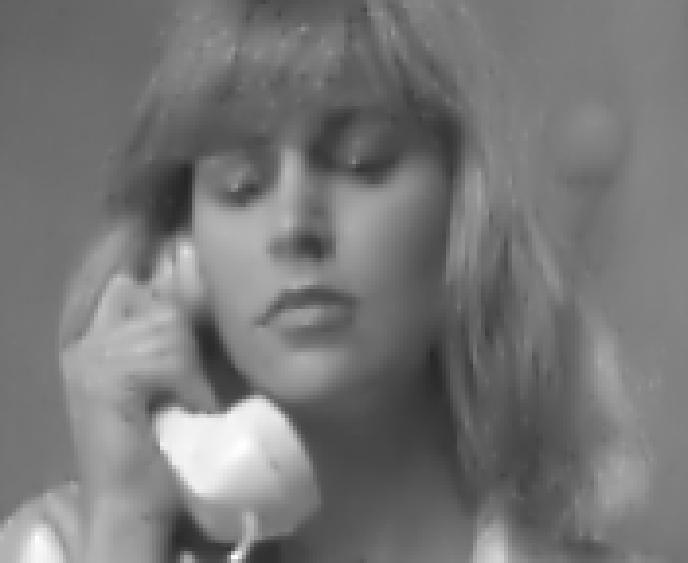}&
\includegraphics[width=0.16\textwidth]{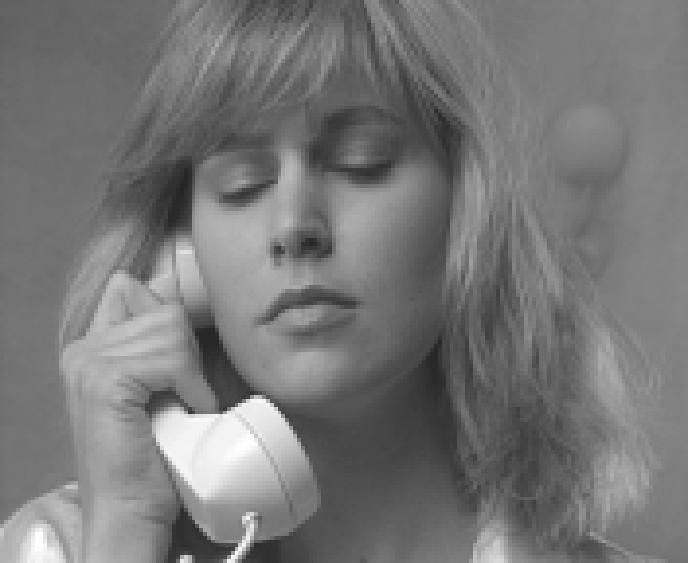}\\

\includegraphics[width=0.16\textwidth]{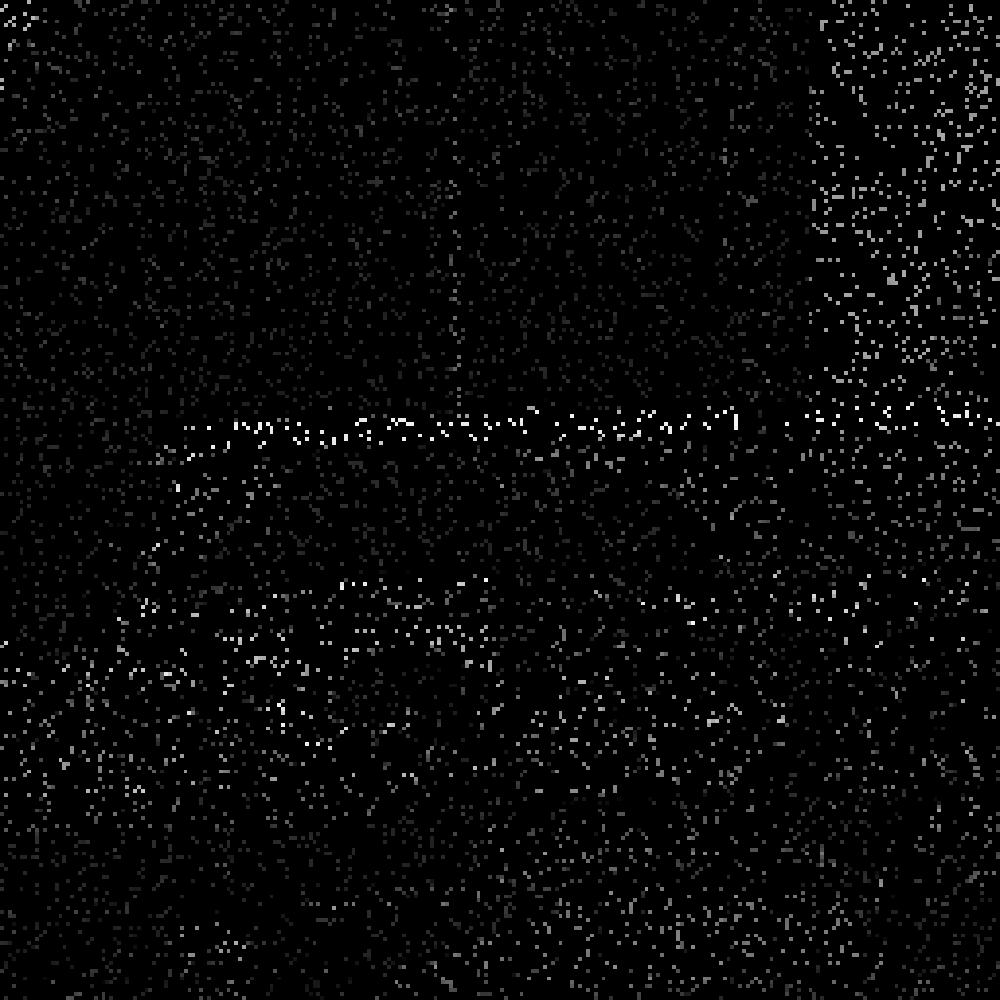}&
\includegraphics[width=0.16\textwidth]{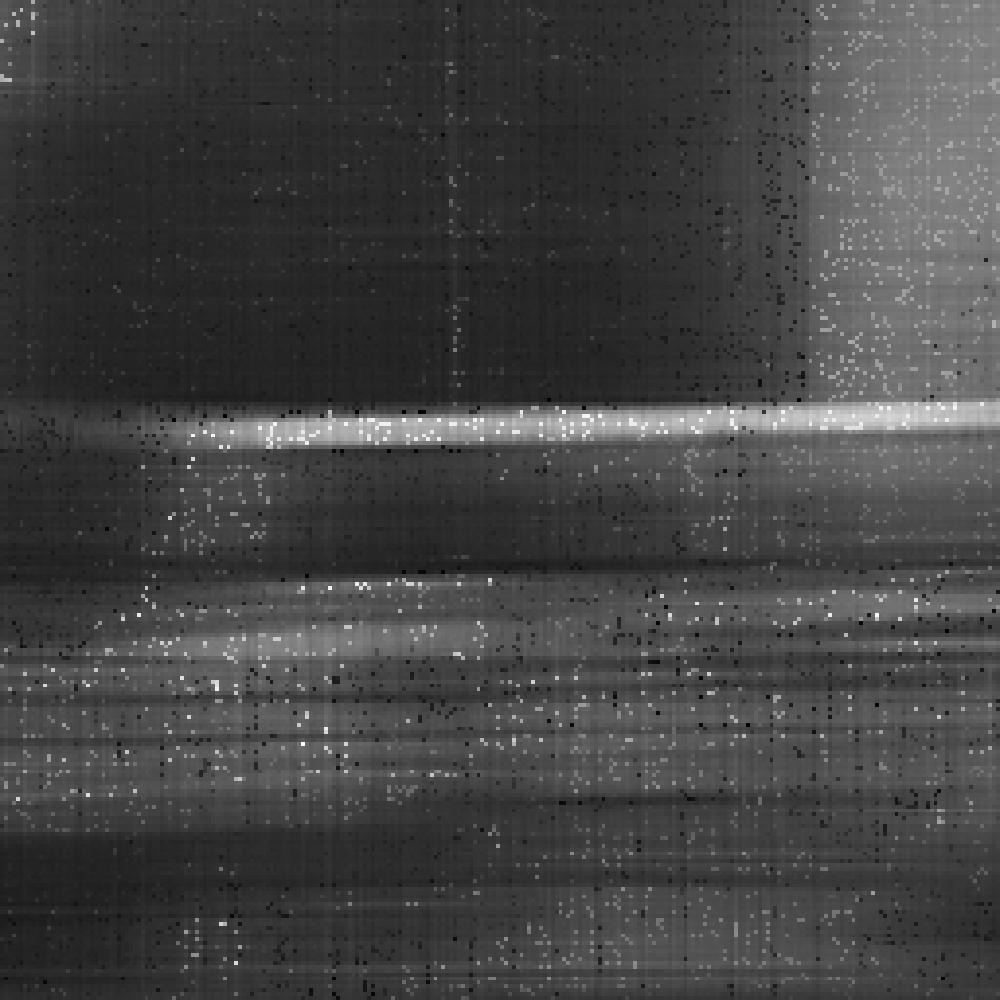}&
\includegraphics[width=0.16\textwidth]{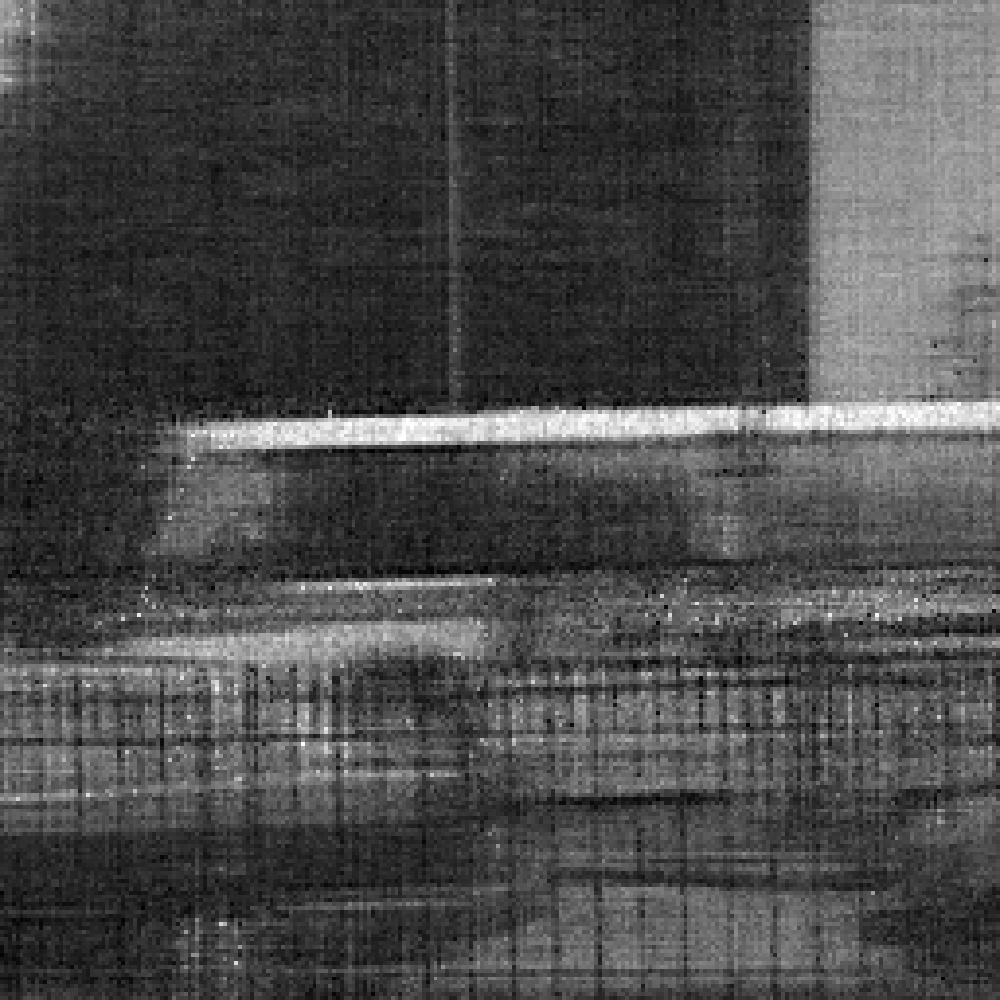}&
\includegraphics[width=0.16\textwidth]{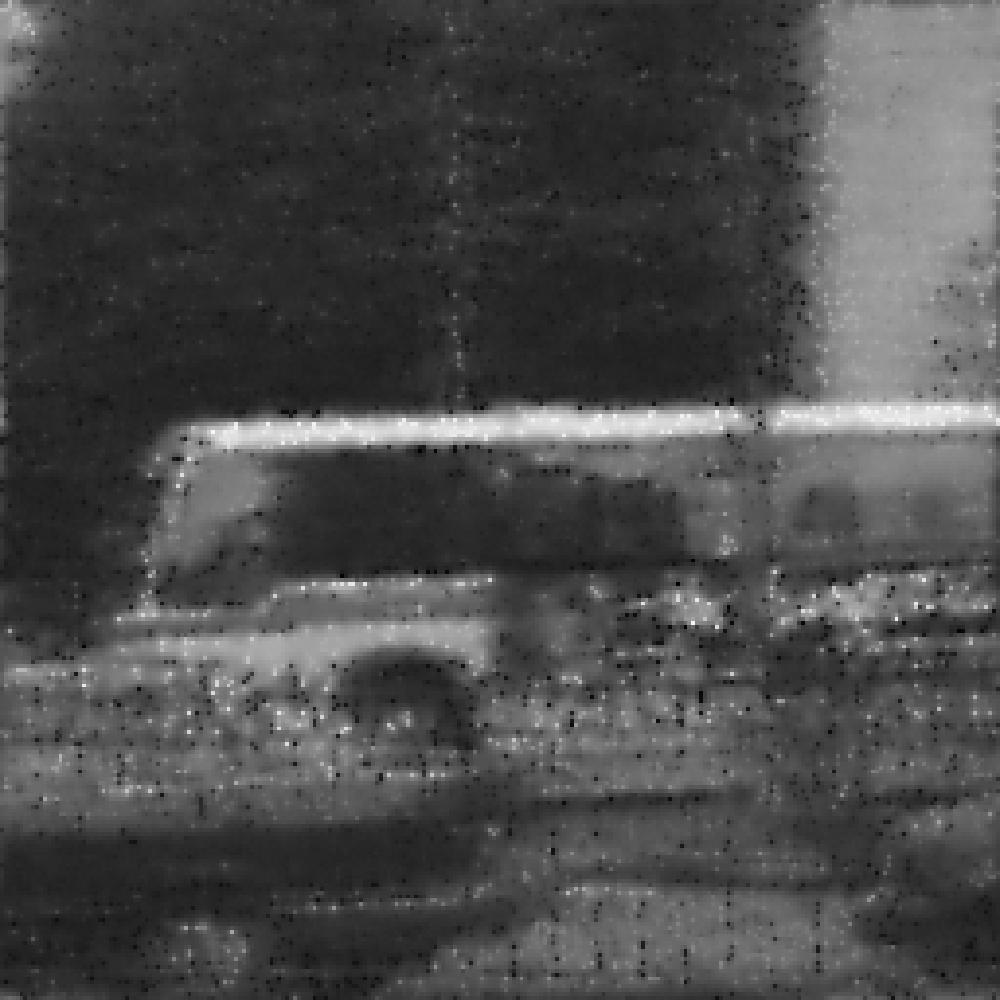}&
\includegraphics[width=0.16\textwidth]{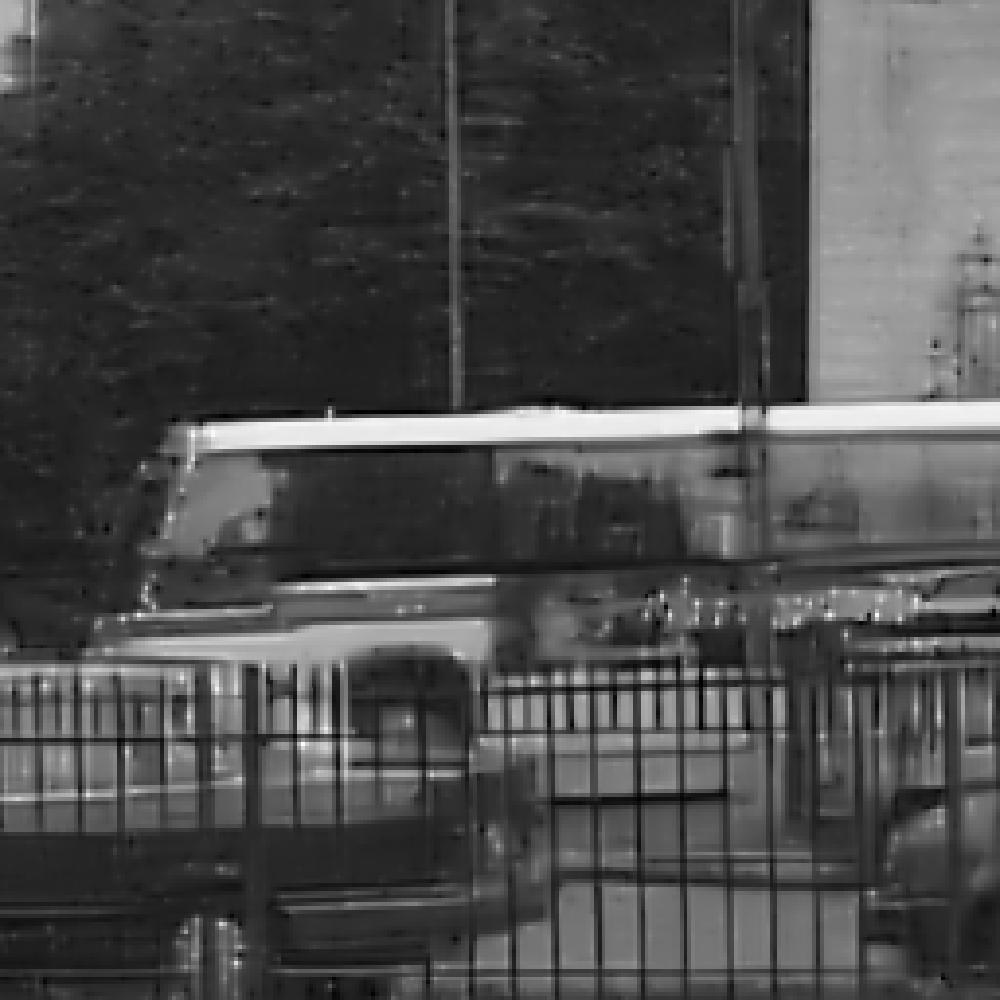}&
\includegraphics[width=0.16\textwidth]{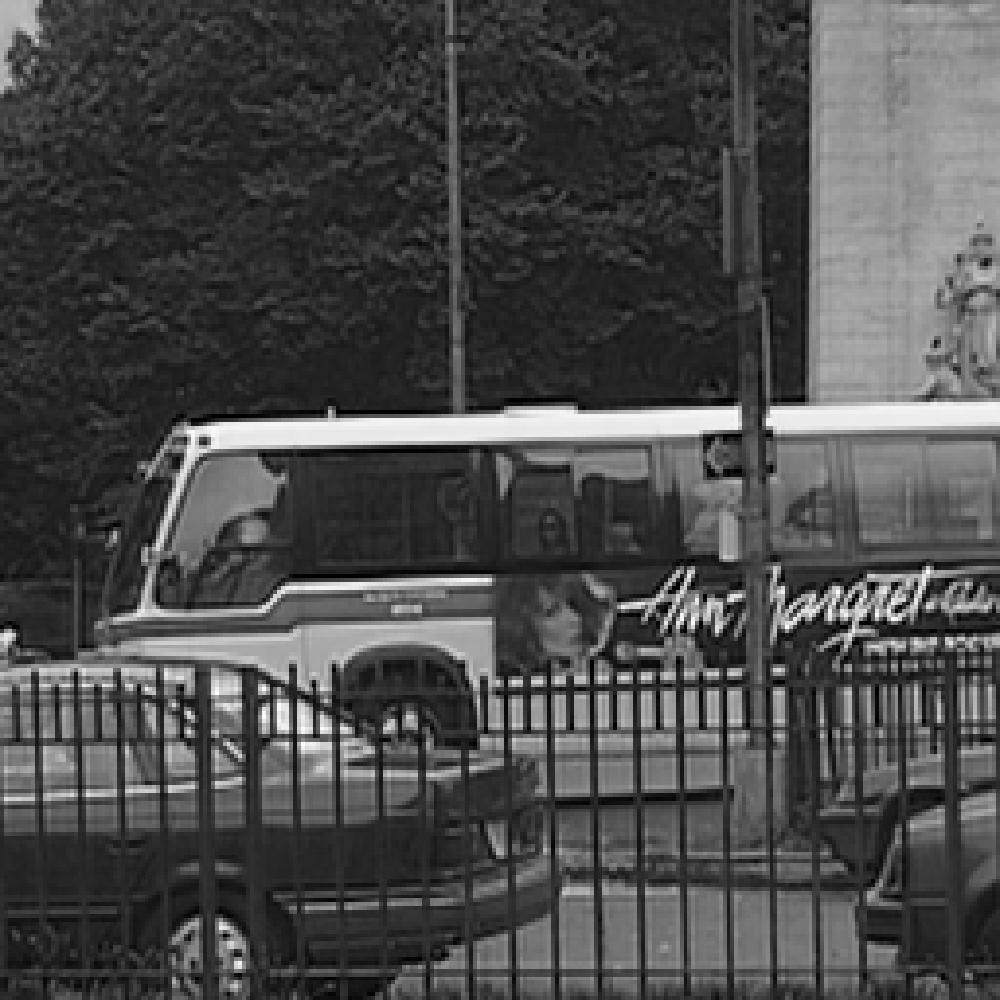}\\

  \end{tabular}
  \caption{The $20$-th frame of the videos recovered by SNN, TNN, TNN-3DTV, and DP3LRTC with the sampling rate $10\%$.}
  \label{fig:video}
\end{figure*}
\subsection{MSI Completion}\label{subsec:msic}
In this subsection, we test 7  MSIs  from the \textit{CAVE} dataset\footnote{Available at \url{http://www.cs.columbia.edu/CAVE/databases/multispectral/}.} which are of size $256 \times 256 \times 31$ with the wavelengths in the range of $400\sim700$ nm at an interval of 10 nm. For MSIs,   we feed the spatial slices into  the FFDNet denoiser  trained for gray images as the
PnP denoiser.

\begin{table}[htbp]
\footnotesize
\selectfont
\setlength{\tabcolsep}{1.5pt}
\renewcommand\arraystretch{0.8}
\caption{Quantitative comparison of the results by SNN \cite{liu2013tensor},  TNN  \cite{zhang2014novel}, TNN-3DTV \cite{jiang2018anisotropic}, and the proposed method on MSIs. The \textbf{best} and \underline{second} best values are highlighted in bold and underlined, respectively.}
\centering
\begin{tabular}{cccccccccccccccc}
\toprule
\multicolumn{1}{c}{\multirow{2}[4]{*}{MSI}}& \multicolumn{1}{c}{\multirow{2}[4]{*}{SR}} & \multicolumn{4}{c}{PSNR}     & & \multicolumn{4}{c}{SSIM}& & \multicolumn{4}{c}{Time (s)}\\
\cmidrule{3-16}\multicolumn{2}{c}{} & {\scriptsize SNN}   & {\scriptsize TNN}   & {\scriptsize TNN-3DTV} & {\scriptsize  DP3LRTC} && {\scriptsize SNN}   & {\scriptsize TNN}   & {\scriptsize TNN-3DTV} & {\scriptsize DP3LRTC} && {\scriptsize SNN}   & {\scriptsize TNN}   & {\scriptsize TNN-3DTV} & {\scriptsize DP3LRTC}\\\midrule

\multirow{3}[2]{*}{\shortstack{\textit{Balloons}\\ \\$256\times256\times31$}}
& 5\%   & 24.29  & 31.35  & \underline{32.66}  & \textbf{39.04} && 0.8232  & 0.8670  & \underline{0.9381}  & \textbf{0.9843} && \bf 8 & \underline{58}&  191 &  144 \\
& 10\%  & 31.27  & 35.63  & \underline{37.66}  & \textbf{42.98} && 0.9242  & 0.9380  & \underline{0.9735}  & \textbf{0.9937} && \bf 7 & \underline{61}&  200 &  155 \\
& 20\%  & 37.34  & 41.11  & \underline{42.77}  & \textbf{47.54} && 0.9737  & 0.9803  & \underline{0.9904}  & \textbf{0.9975} && \bf 7 & \underline{59}&  192 &  147 \\
\midrule

\multirow{3}[2]{*}{\shortstack{\textit{Beads}\\ \\$256\times256\times31$}}
& 5\%   & 16.50  & 19.86  & \underline{21.27}  & \textbf{22.53} && 0.3259  & 0.4569  & \underline{0.6506}  & \textbf{0.6912} && \bf 9 & \underline{60}&  199 &  147 \\
& 10\%  & 17.99  & 22.92  & \underline{23.95}  & \textbf{26.08} && 0.4191  & 0.6512  & \underline{0.7907}  & \textbf{0.8369} && \bf 7 & \underline{61}&  198 &  153 \\
& 20\%  & 21.34  & 27.61  & \underline{27.78}  & \textbf{30.20} && 0.6404  & 0.8420  & \underline{0.9024}  & \textbf{0.9319} && \bf 7 & \underline{58}&  197 &  152 \\
\midrule

\multirow{3}[2]{*}{\shortstack{\textit{Cd}\\ \\$256\times256\times31$}}
& 5\%   & 22.19  & 26.51  & \underline{29.35}  & \textbf{30.22} && 0.8497  & 0.8029  & \underline{0.9251}  & \textbf{0.9597} && \bf 10 & \underline{62}&  191 &  144 \\
& 10\%  & 25.98  & 29.54  & \underline{31.92}  & \textbf{35.30} && 0.9018  & 0.8871  & \underline{0.9550}  & \textbf{0.9790} && \bf 10 & \underline{69}&  196 &  157 \\
& 20\%  & 30.85  & 33.22  & \underline{36.04}  & \textbf{39.35} && 0.9498  & 0.9447  & \underline{0.9791}  & \textbf{0.9890} && \bf 8  & \underline{64}&  193 &  150 \\
\midrule

\multirow{3}[2]{*}{\shortstack{\textit{Clay}\\ \\$256\times256\times31$}}
& 5\%   & 28.83  & 33.62  & \underline{34.25}  & \textbf{39.89} && 0.9224  & 0.9052  & \underline{0.9487}  & \textbf{0.9680} && \bf 10 & \underline{59}&  194 &  145 \\
& 10\%  & 35.78  & 38.22  & \underline{39.66}  & \textbf{43.80} && 0.9726  & 0.9657  & \underline{0.9819}  & \textbf{0.9907} && \bf 9  & \underline{62}&  193 &  144 \\
& 20\%  & 41.58  & 43.69  & \underline{44.69}  & \textbf{48.02} && 0.9895  & 0.9813  & \underline{0.9890}  & \textbf{0.9952} && \bf 9  & \underline{62}&  191 &  147 \\
\midrule

\multirow{3}[2]{*}{\shortstack{\textit{Face}\\ \\$256\times256\times31$}}
& 5\%   & 25.64  & 32.32  & \underline{32.70}  & \textbf{36.82} && 0.8346  & 0.8981  & \underline{0.9381}  & \textbf{0.9663} && \bf 9 & \underline{61}&  198 &  146 \\
& 10\%  & 30.51  & 36.50  & \underline{37.52}  & \textbf{39.66} && 0.9125  & 0.9548  & \underline{0.9736}  & \textbf{0.9829} && \bf 8 & \underline{60}&  192 &  145 \\
& 20\%  & 36.20  & 41.33  & \underline{42.00}  & \textbf{42.96} && 0.9664  & 0.9842  & \underline{0.9778}  & \textbf{0.9916} && \bf 8 & \underline{61}&  192 &  147 \\
\midrule

\multirow{3}[2]{*}{\shortstack{\textit{Feathers}\\ \\$256\times256\times31$}}
& 5\%   & 20.45  & 27.55  & \underline{28.23}  & \textbf{31.14} && 0.6565  & 0.7694  & \underline{0.8712}  & \textbf{0.9375} && \bf 7 & \underline{58}&  196 &  151 \\
& 10\%  & 24.58  & 31.19  & \underline{31.83}  & \textbf{34.59} && 0.7925  & 0.8717  & \underline{0.9308}  & \textbf{0.9656} && \bf 8 & \underline{58}&  193 &  147 \\
& 20\%  & 29.30  & 35.86  & \underline{36.73}  & \textbf{38.68} && 0.8995  & 0.9481  & \underline{0.9726}  & \textbf{0.9838} && \bf 8 & \underline{60}&  193 &  150 \\
\midrule

\multirow{3}[2]{*}{\shortstack{\textit{Flowers}\\ \\$256\times256\times31$}}
& 5\%   & 20.47  & 26.75  & \underline{27.90}  & \textbf{30.34} && 0.6293  & 0.7153  & \underline{0.8288}  & \textbf{0.8894} && \bf 10 & \underline{60}&  196 &  148 \\
& 10\%  & 24.88  & 30.36  & \underline{31.37}  & \textbf{33.70} && 0.7605  & 0.8357  & \underline{0.9025}  & \textbf{0.9419} && \bf 7 & \underline{58}&  192 &  147 \\
& 20\%  & 29.49  & 35.33  & \underline{36.25}  & \textbf{37.66} && 0.8804  & 0.9342  & \underline{0.9621}  & \textbf{0.9745} && \bf 7 & \underline{60}&  194 &  149 \\
\midrule

\multirow{3}[2]{*}{Average}
& 5\%   & 22.62  & 28.28  & \underline{29.48}  & \textbf{32.85} && 0.7202  & 0.7735  & \underline{0.8715}  & \textbf{0.9138} && \bf 9 & \underline{60}&  195 &  146 \\
& 10\%  & 27.28  & 32.05  & \underline{33.42}  & \textbf{36.59} && 0.8119  & 0.8720  & \underline{0.9297}  & \textbf{0.9558} && \bf 8 & \underline{61}&  195 &  150 \\
& 20\%  & 32.30  & 36.88  & \underline{38.04}  & \textbf{40.63} && 0.9000  & 0.9450  & \underline{0.9676}  & \textbf{0.9805} && \bf 8 & \underline{60}&  193 &  149 \\
\bottomrule
\end{tabular}%
\label{tab:msic}
\end{table}

\begin{figure}[htbp]
\small
\setlength{\tabcolsep}{0.9pt}
\centering
\begin{tabular}{cccccc}
Observed & SNN & TNN & TNN-3DTV & DP3LRTC & Ground truth\\

\includegraphics[width=0.16\textwidth]{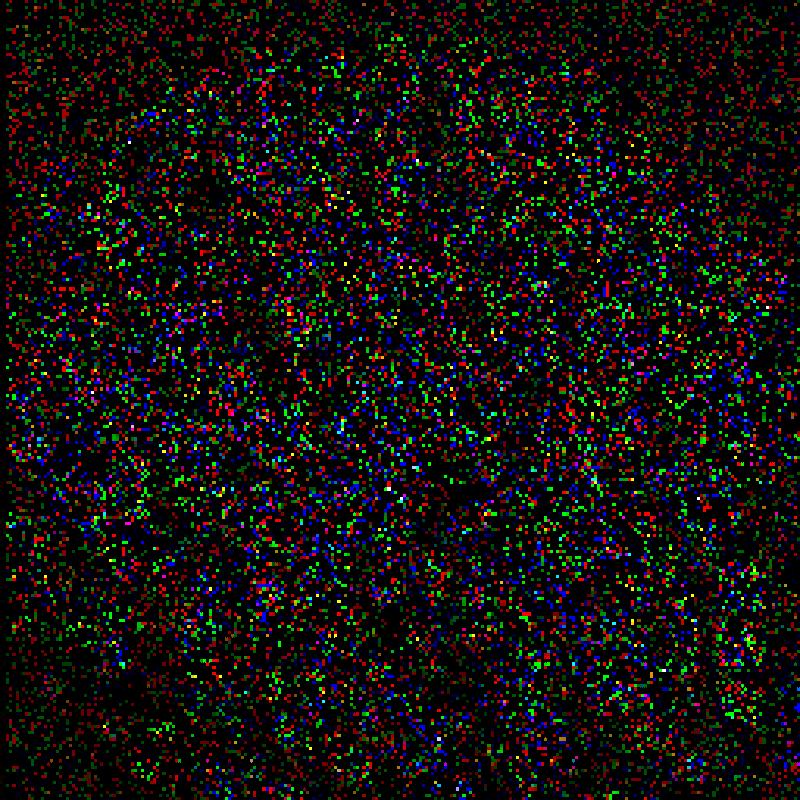}&
\includegraphics[width=0.16\textwidth]{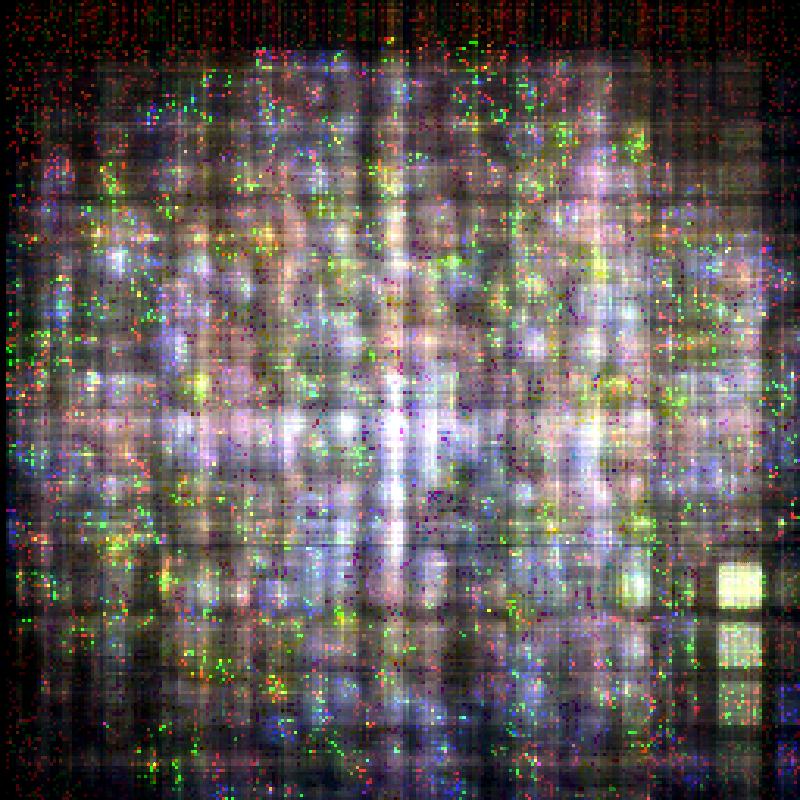}&
\includegraphics[width=0.16\textwidth]{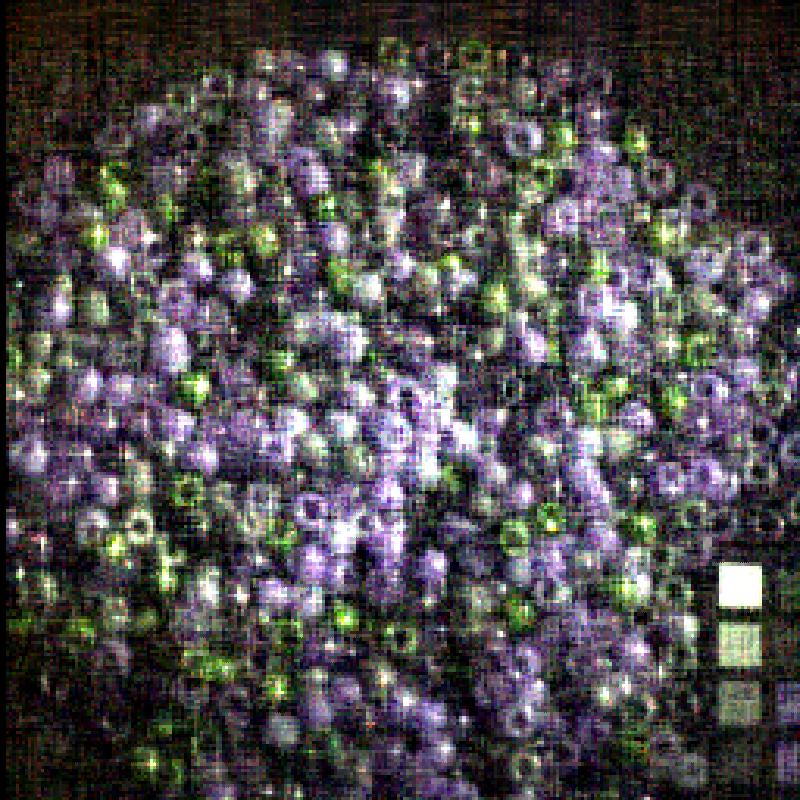}&
\includegraphics[width=0.16\textwidth]{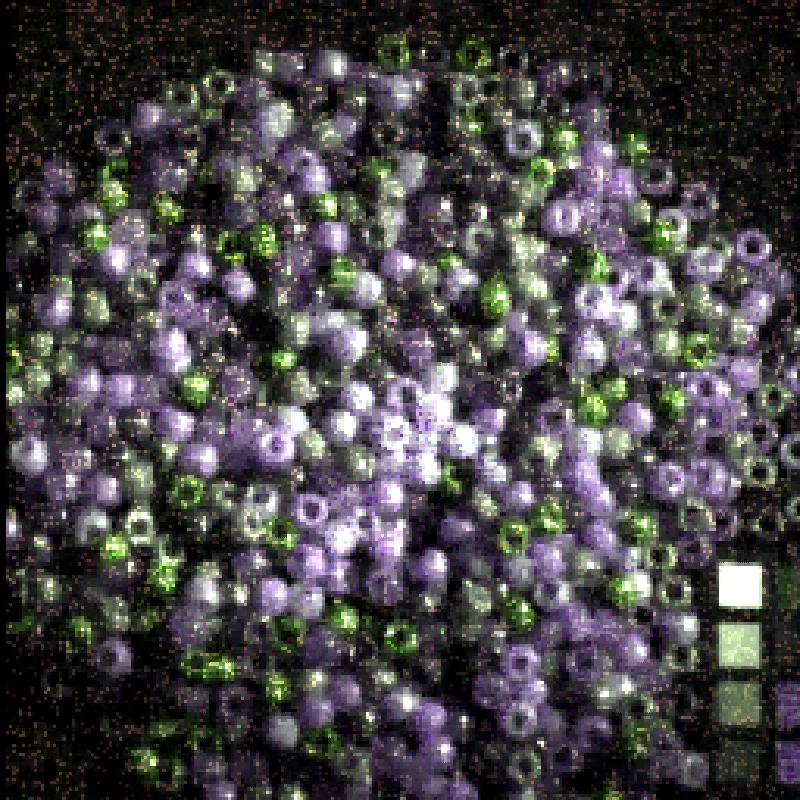}&
\includegraphics[width=0.16\textwidth]{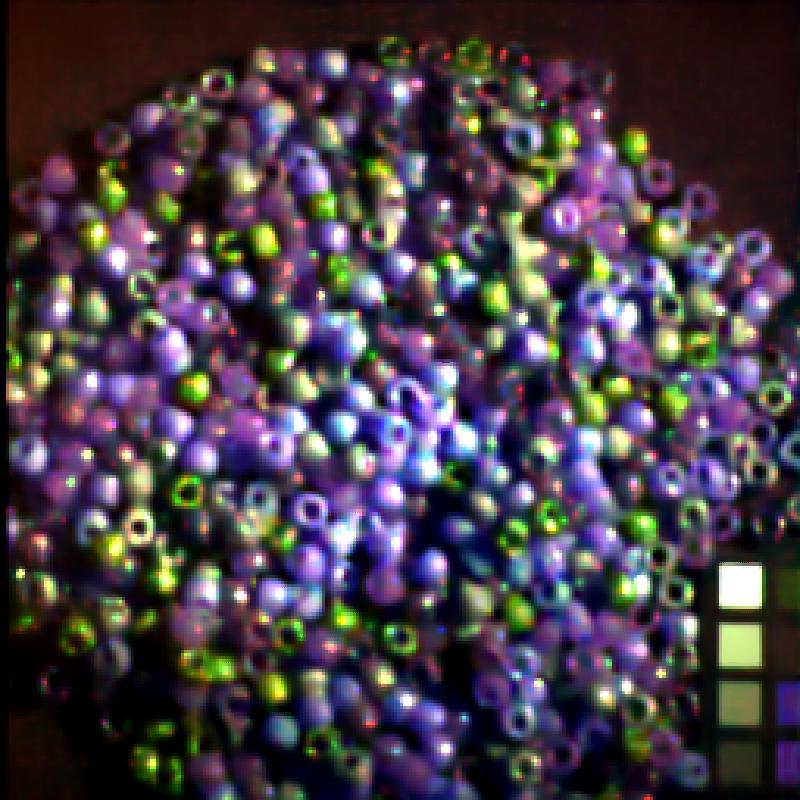}&
\includegraphics[width=0.16\textwidth]{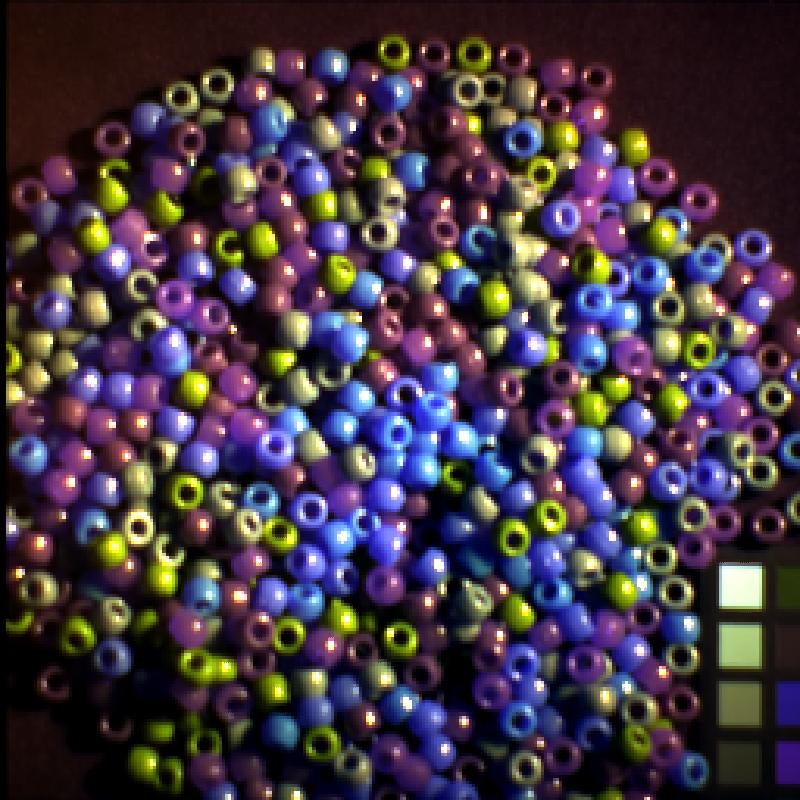}\\

\includegraphics[width=0.16\textwidth]{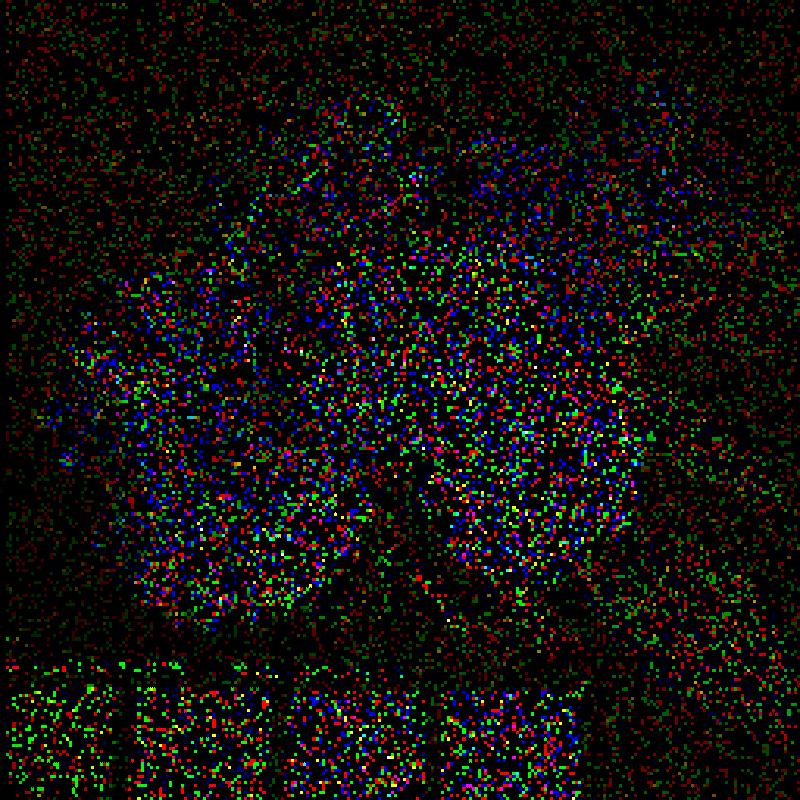}&
\includegraphics[width=0.16\textwidth]{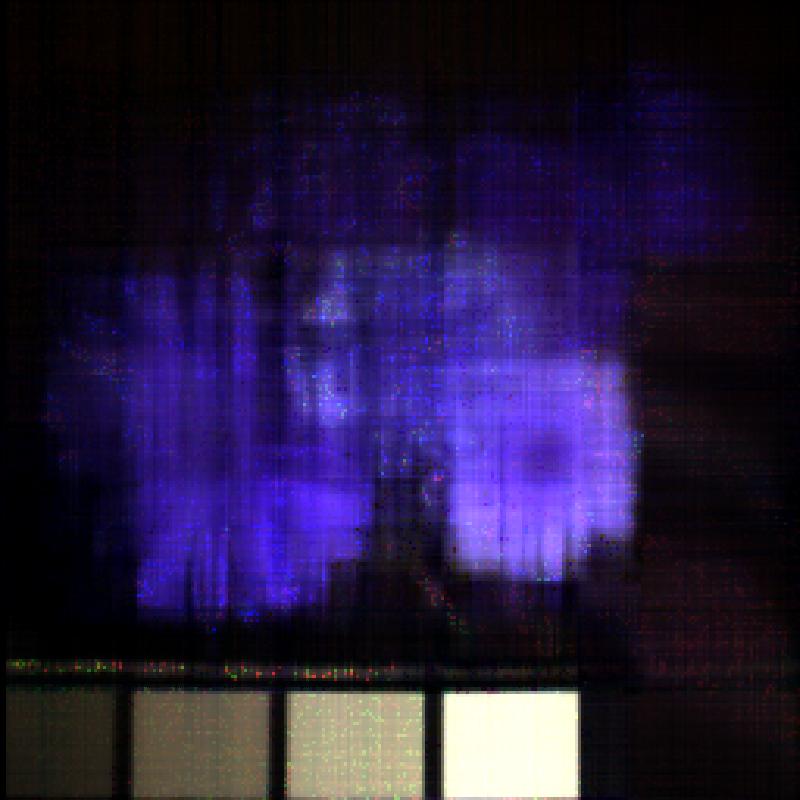}&
\includegraphics[width=0.16\textwidth]{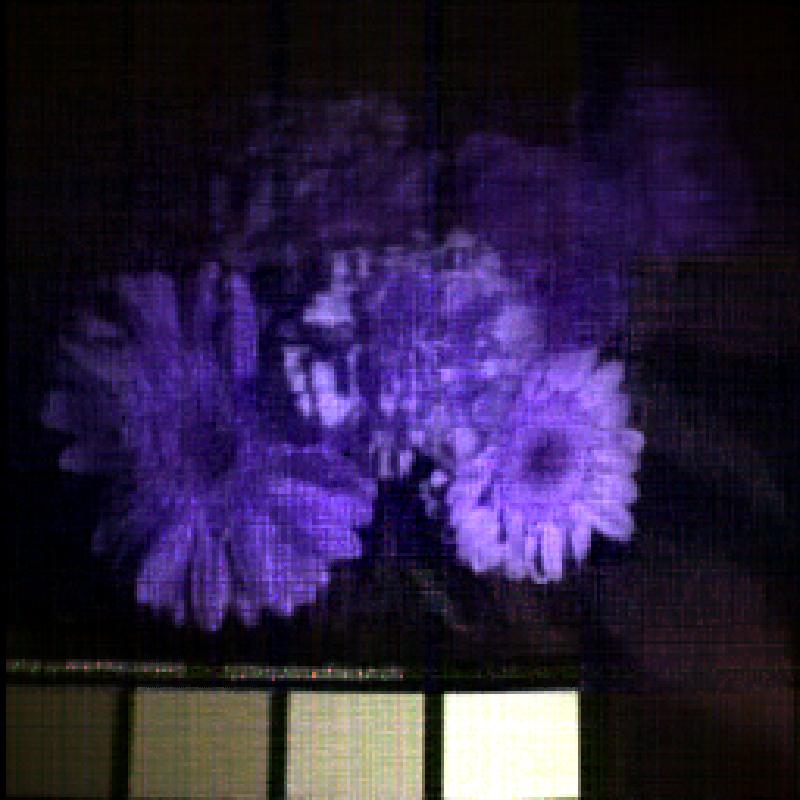}&
\includegraphics[width=0.16\textwidth]{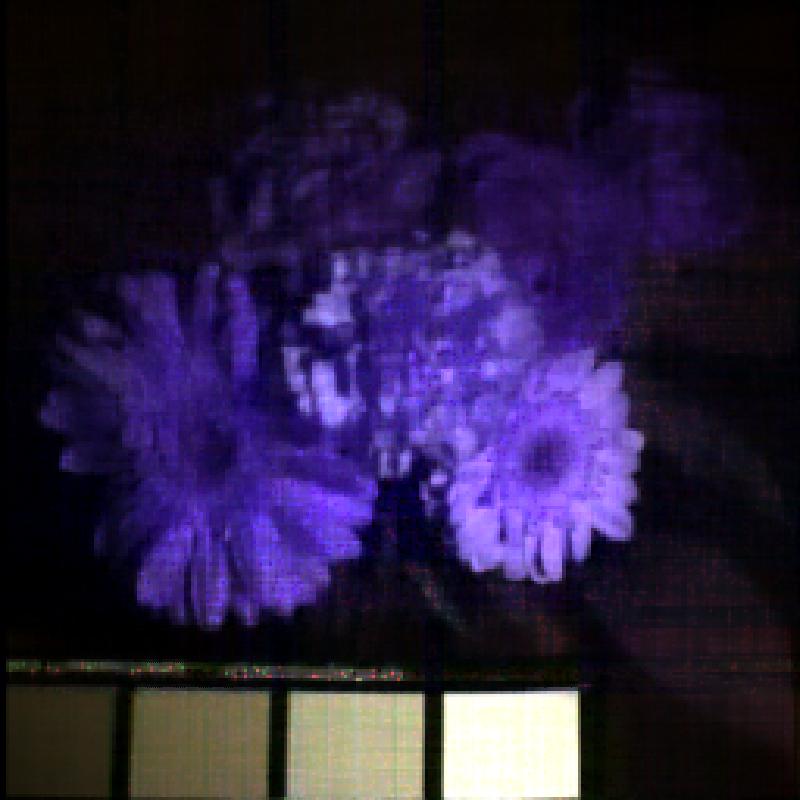}&
\includegraphics[width=0.16\textwidth]{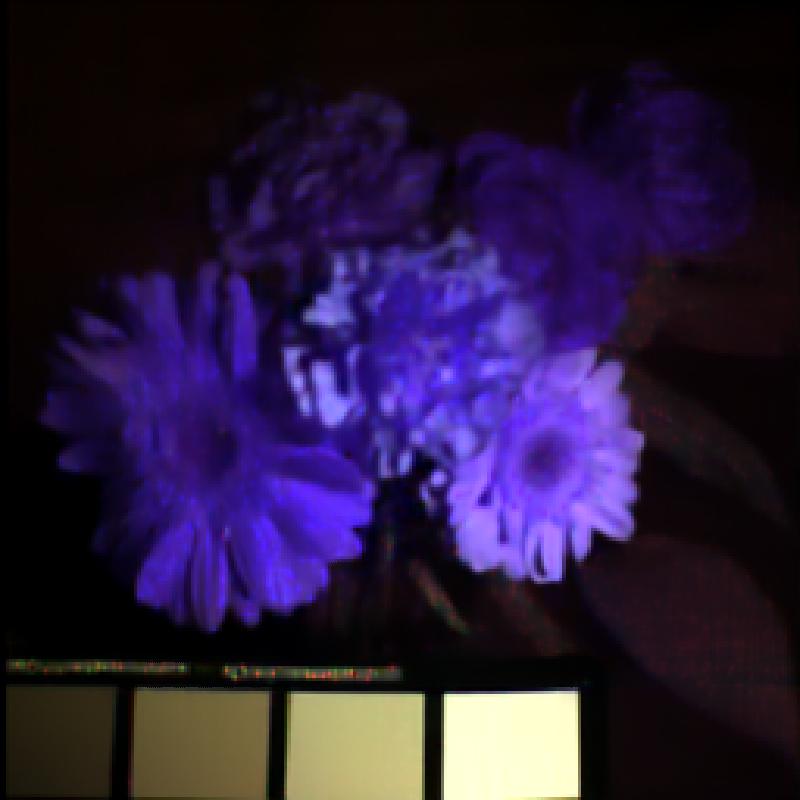}&
\includegraphics[width=0.16\textwidth]{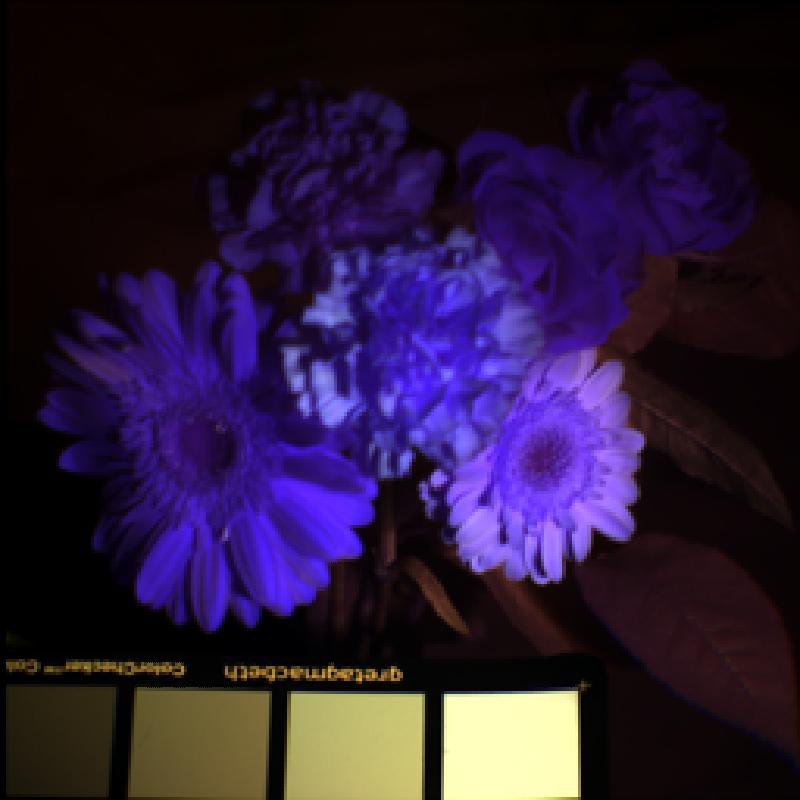}\\

  \end{tabular}
  \caption{The MSIs (pseudo-color images composed of the 1st, 2nd, and 31-st bands) recovered by SNN, TNN, TNN-3DTV, and DP3LRTC  with the sampling rate $10\%$.}
  \label{fig:msi}
\end{figure}

\begin{figure}[htbp]
\centering
\includegraphics[width=0.65\textwidth]{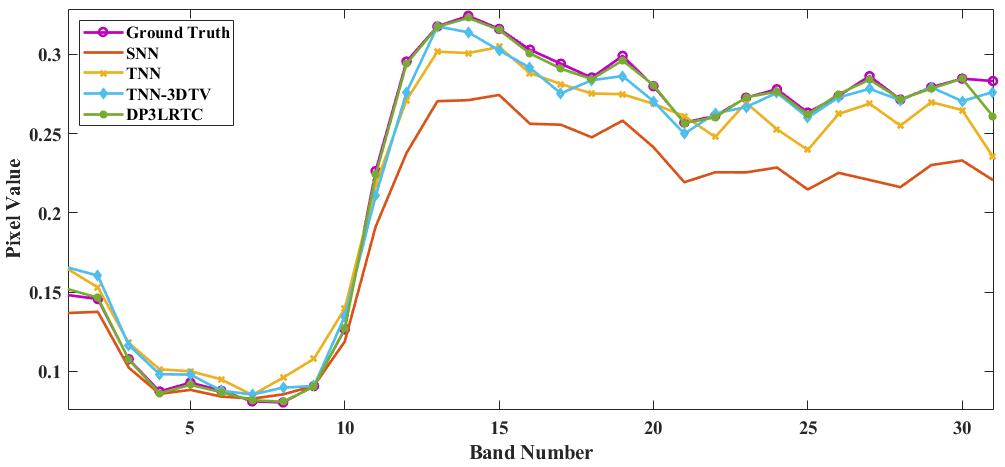}

\caption{The selected tubes of results recovered by by SNN \cite{liu2013tensor},  TNN  \cite{zhang2014novel}, TNN-3DTV \cite{jiang2018anisotropic}, and the proposed method on the MSI \textit{Balloons} with the sampling rate 10\%.}
\label{fig:tube}
\end{figure}

Tab.\thinspace\ref{tab:msic} exhibits the PSNR and SSIM values of all the results by different methods. The DP3LRTC achieves the highest PSNR and SSIM values, while the TNN-3DTV gets the second best. Fig.\thinspace\ref{fig:msi} displays the recovered MSIs (pseudo-color images composed of the 1st, 2nd, and 31-st bands)  by different methods with the sampling rate $10\%$.
From Fig.\thinspace\ref{fig:msi}, it can be observed that the results recovered by DP3LRTC are of the highest quality.
In Fig.\thinspace\ref{fig:tube}, we display one selected tube of results recovered  by SNN, TNN, TNN-3DTV, and DP3LRTC on MSI \textit{Balloons} with the sampling rate 10\%. We can observe that the selected tube of the MSI recovered by DP3LRTC is closer to the ground truth as compared with those by SNN, TNN, and TNN-3DTV.

\subsection{Discussions} \label{sec:Dis}
\subsubsection{Convergence Behaviour}\label{subsec:conv}

\begin{figure}[htbp]
\centering
\subfigure{%
    \includegraphics[width=0.65\textwidth]{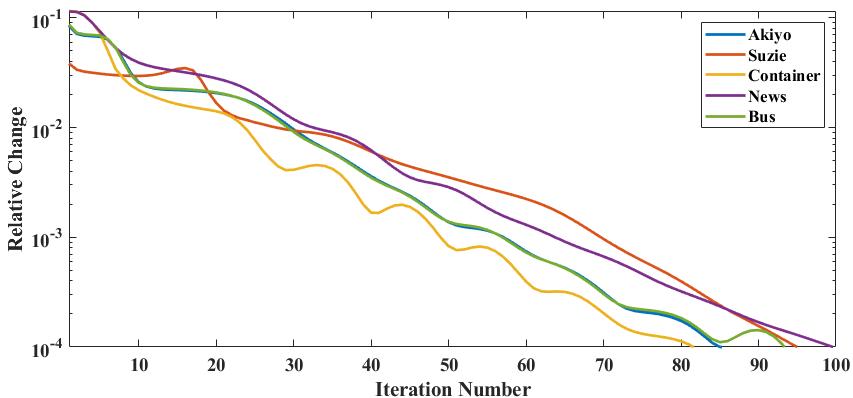}}
 \caption{The convergence curves of the ADMM algorithm on videos \textit{Akiyo}, \textit{Suzie}, \textit{Container}, \textit{News}, and \textit{Bus} with the sampling rate $10\%$.}
  \label{fig:relcha}
\end{figure}

The numerical experiments have shown the  great empirical
success of DP3LRTC. However, it is still an open question whether the ADMM algorithm under the PnP framework
 has a good convergence behavior.
In Fig.\thinspace\ref{fig:relcha}, we displayed the relative change curves of the ADMM algorithm (in the logarithmic scale)  with respect to the iteration on different videos. We can evidently observe the  numerical convergence of the ADMM algorithm.

\subsubsection{Contributions of Different Terms}
\begin{figure}[htbp]
\small
\setlength{\tabcolsep}{0.9pt}
\centering
\begin{tabular}{ccccc}
Observed & TNN & DPR & DP3LRTC & Ground truth\\

\includegraphics[width=0.18\textwidth]{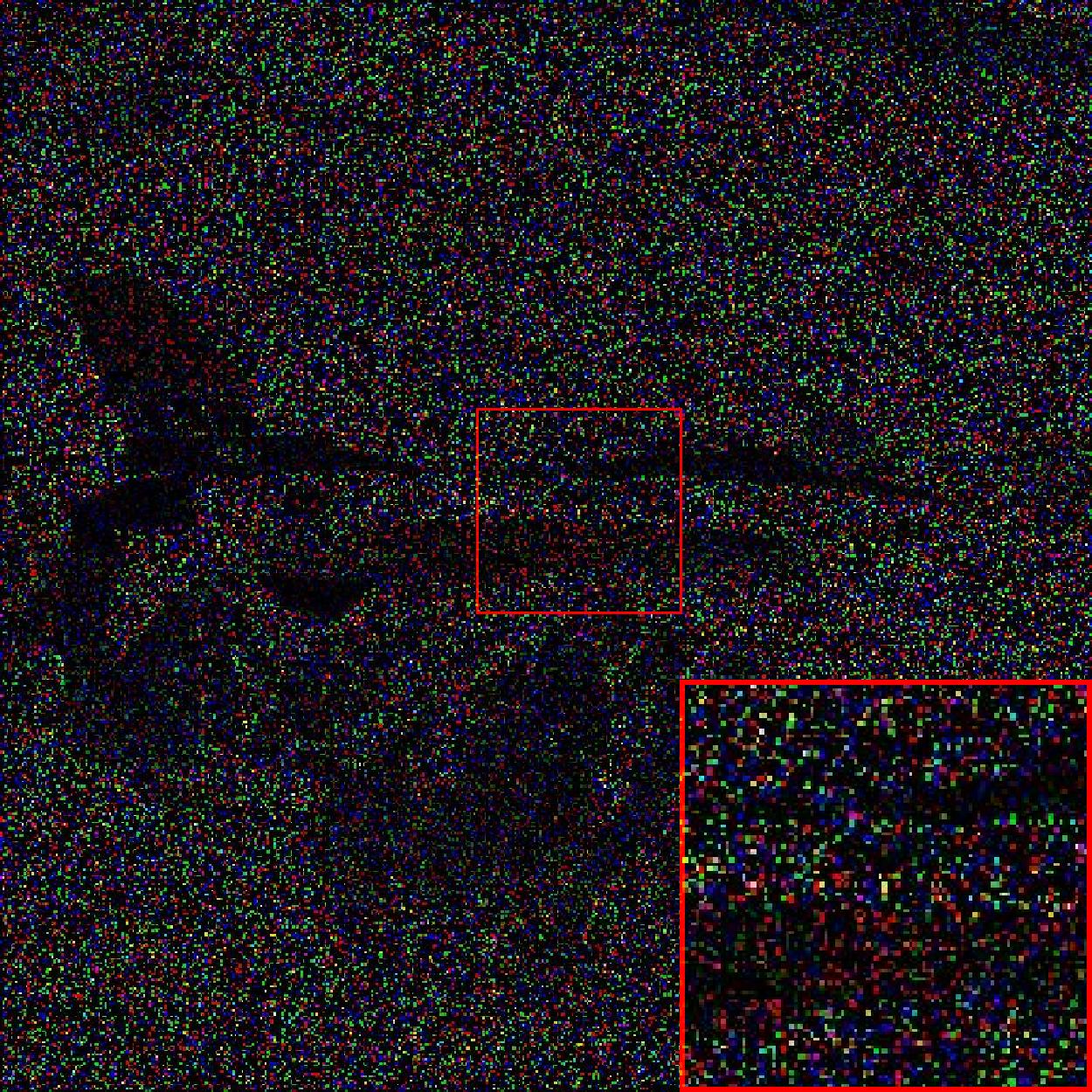}&
\includegraphics[width=0.18\textwidth]{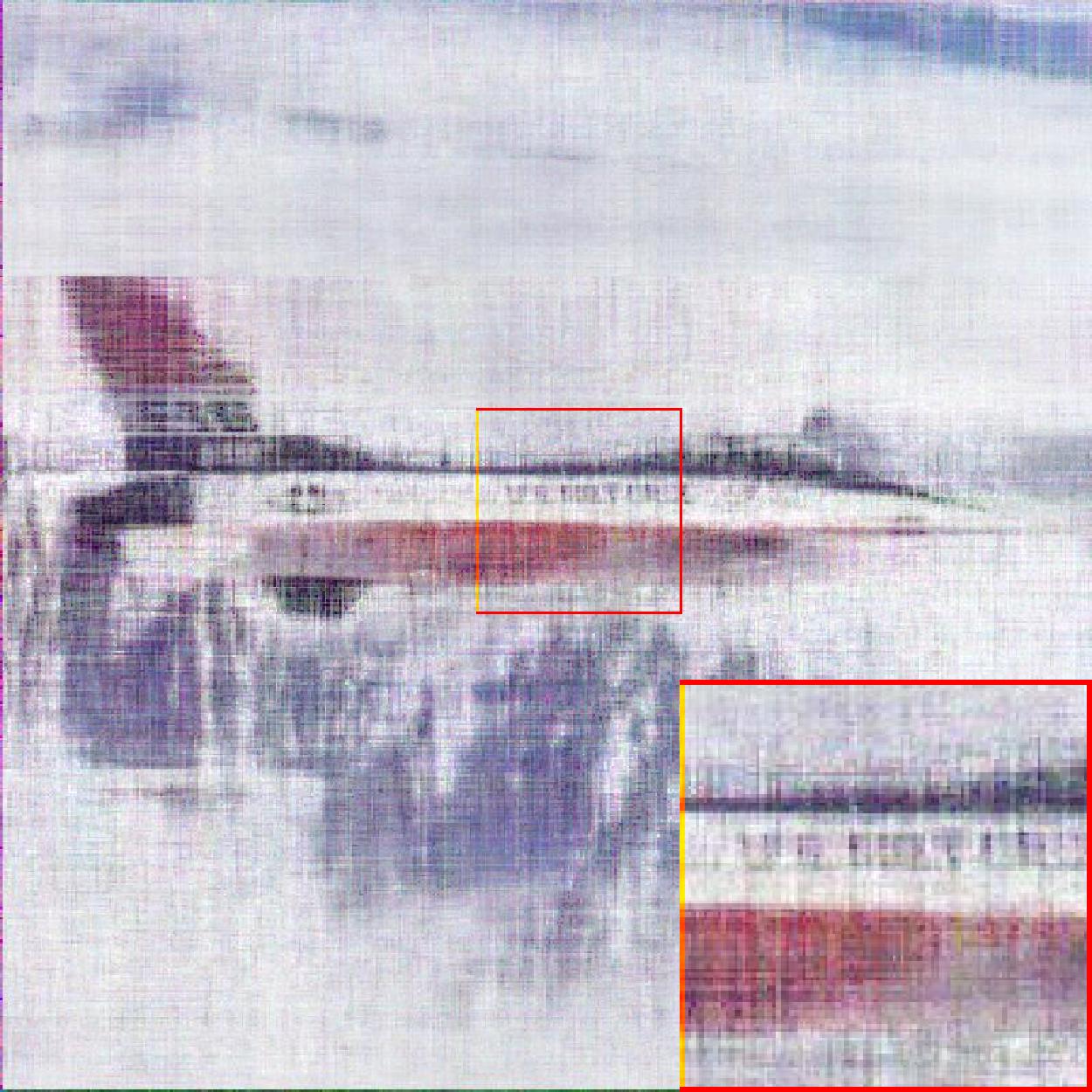}&
\includegraphics[width=0.18\textwidth]{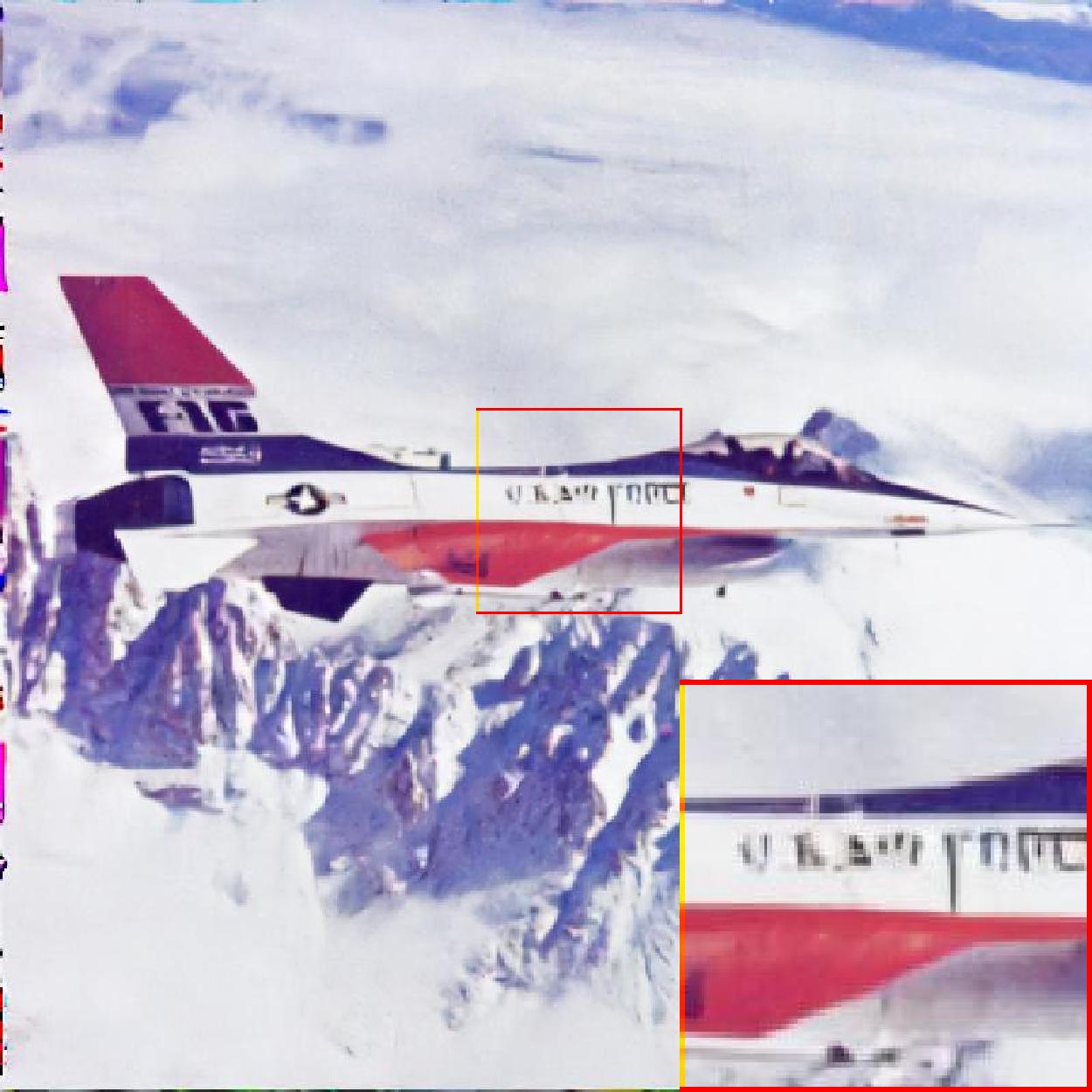}&
\includegraphics[width=0.18\textwidth]{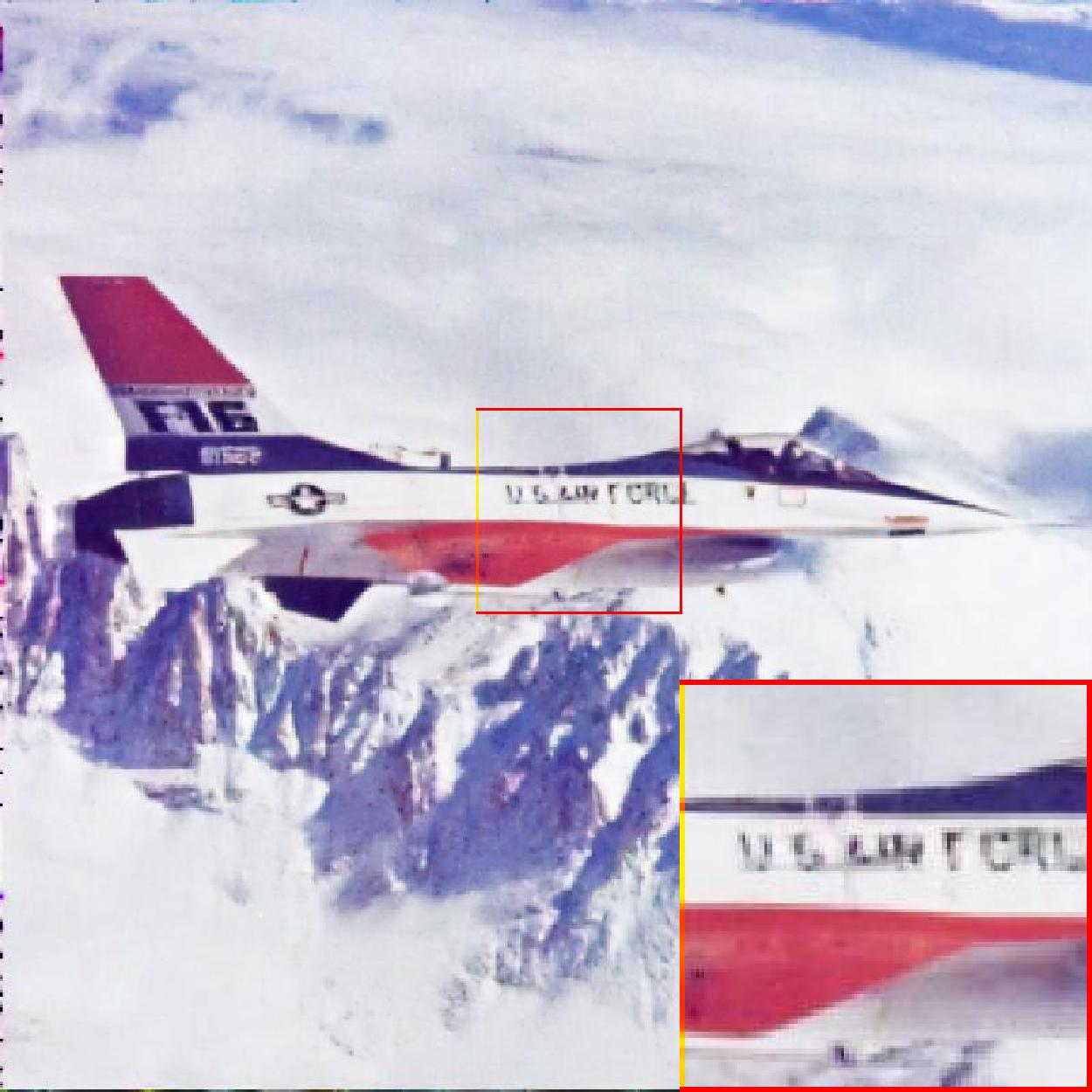}&
\includegraphics[width=0.18\textwidth]{figs/airplane.jpg}\\

PSNR 2.27 dB & PSNR 21.94 dB & PSNR 26.29 dB & PSNR 28.48 dB & PSNR inf\\

\includegraphics[width=0.18\textwidth]{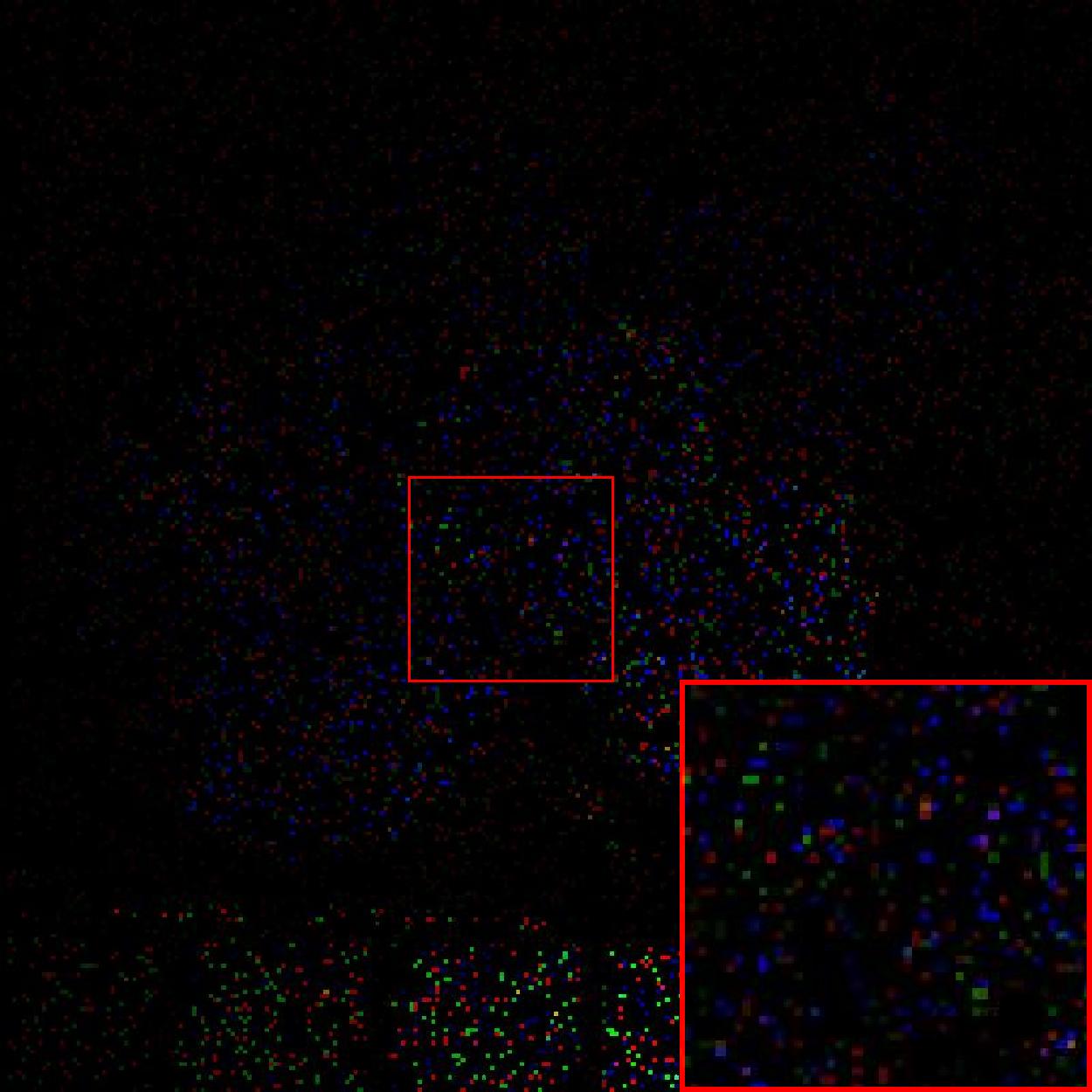}&
\includegraphics[width=0.18\textwidth]{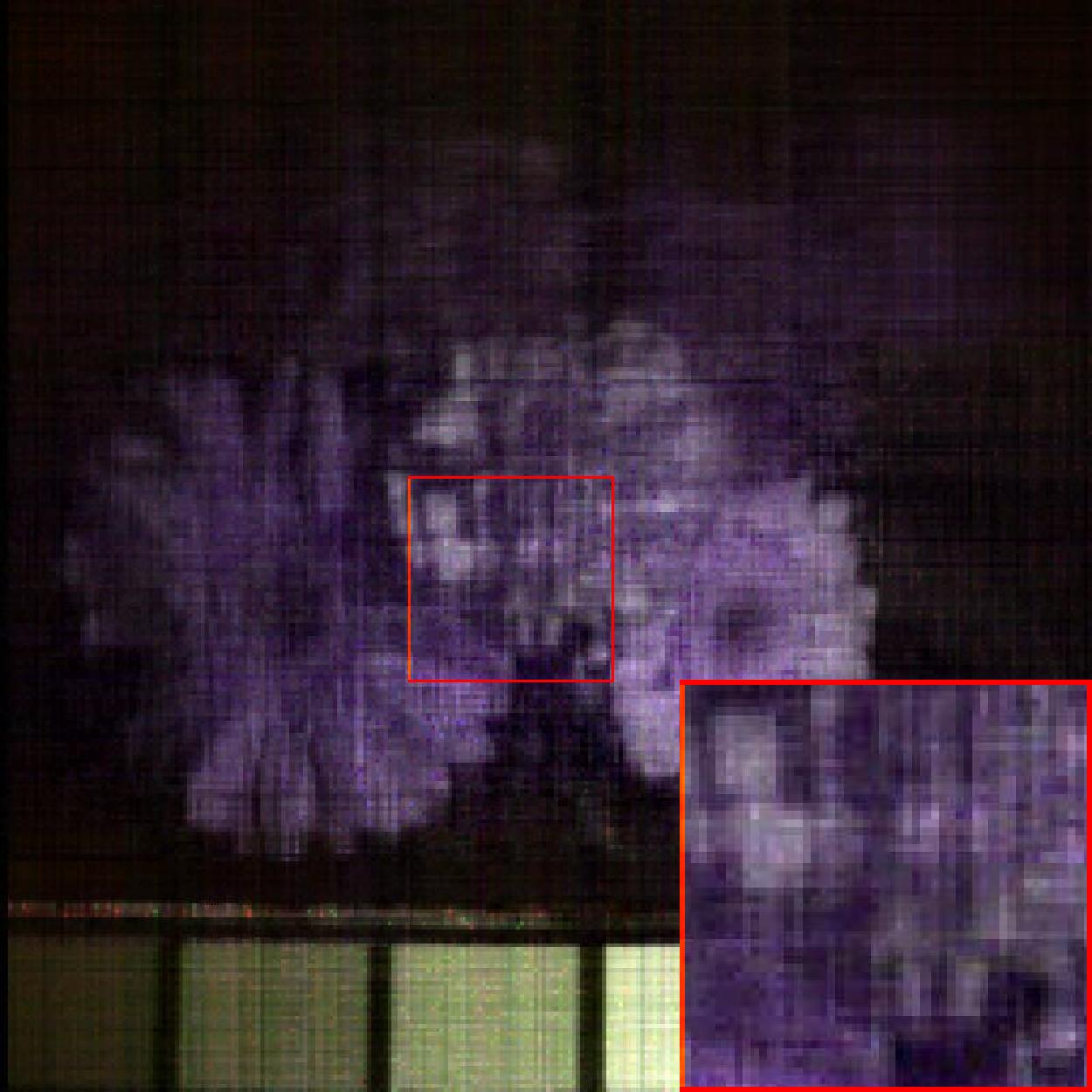}&
\includegraphics[width=0.18\textwidth]{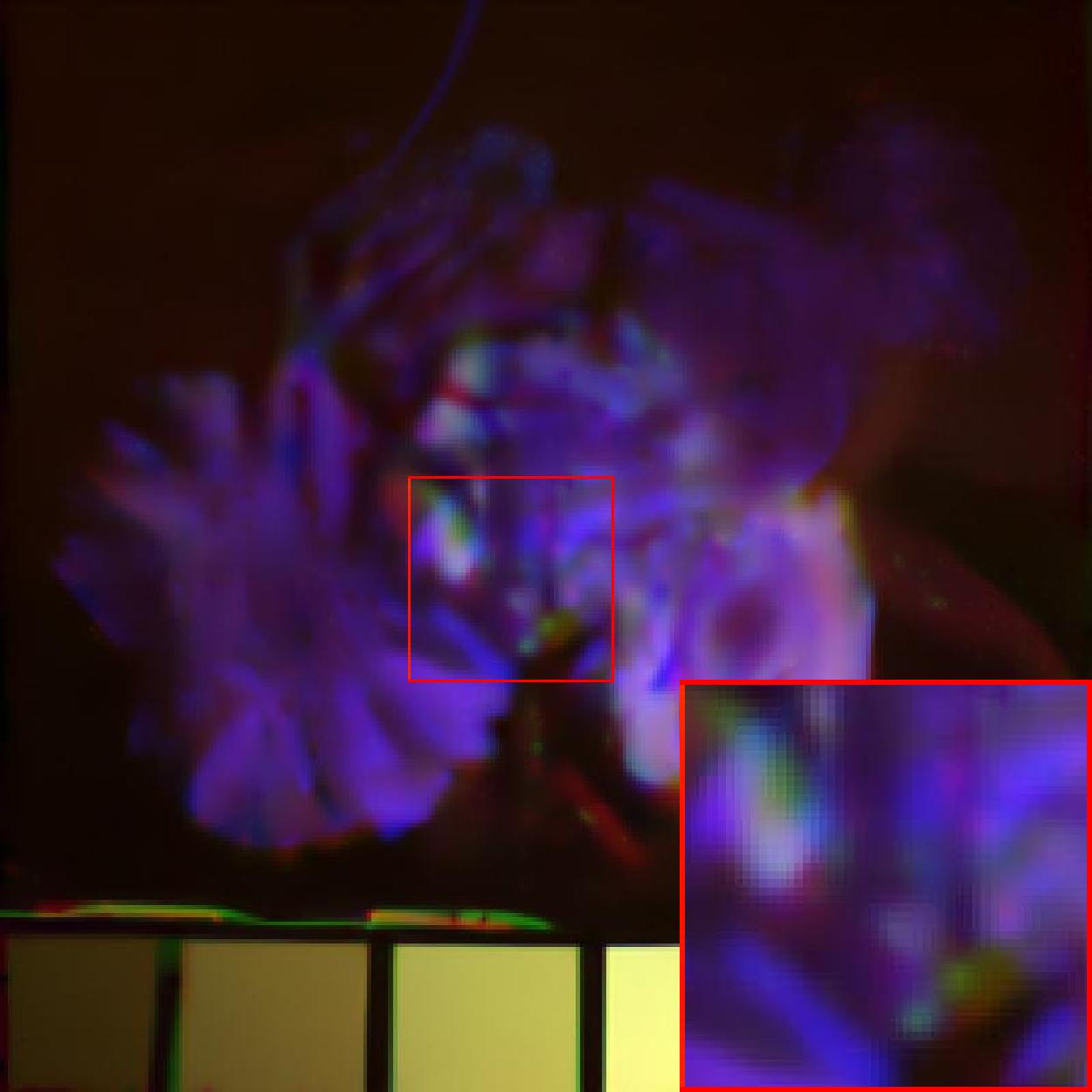}&
\includegraphics[width=0.18\textwidth]{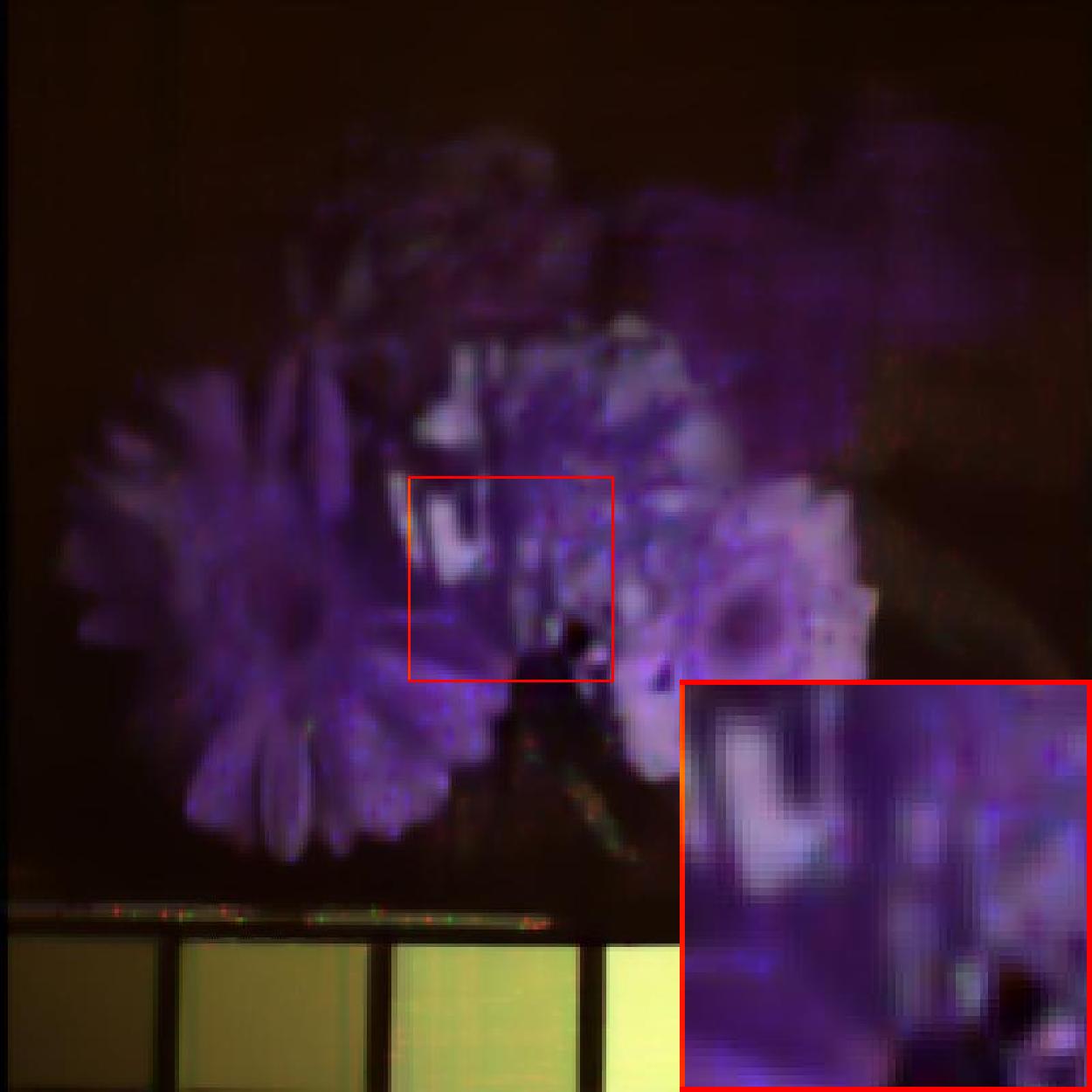}&
\includegraphics[width=0.18\textwidth]{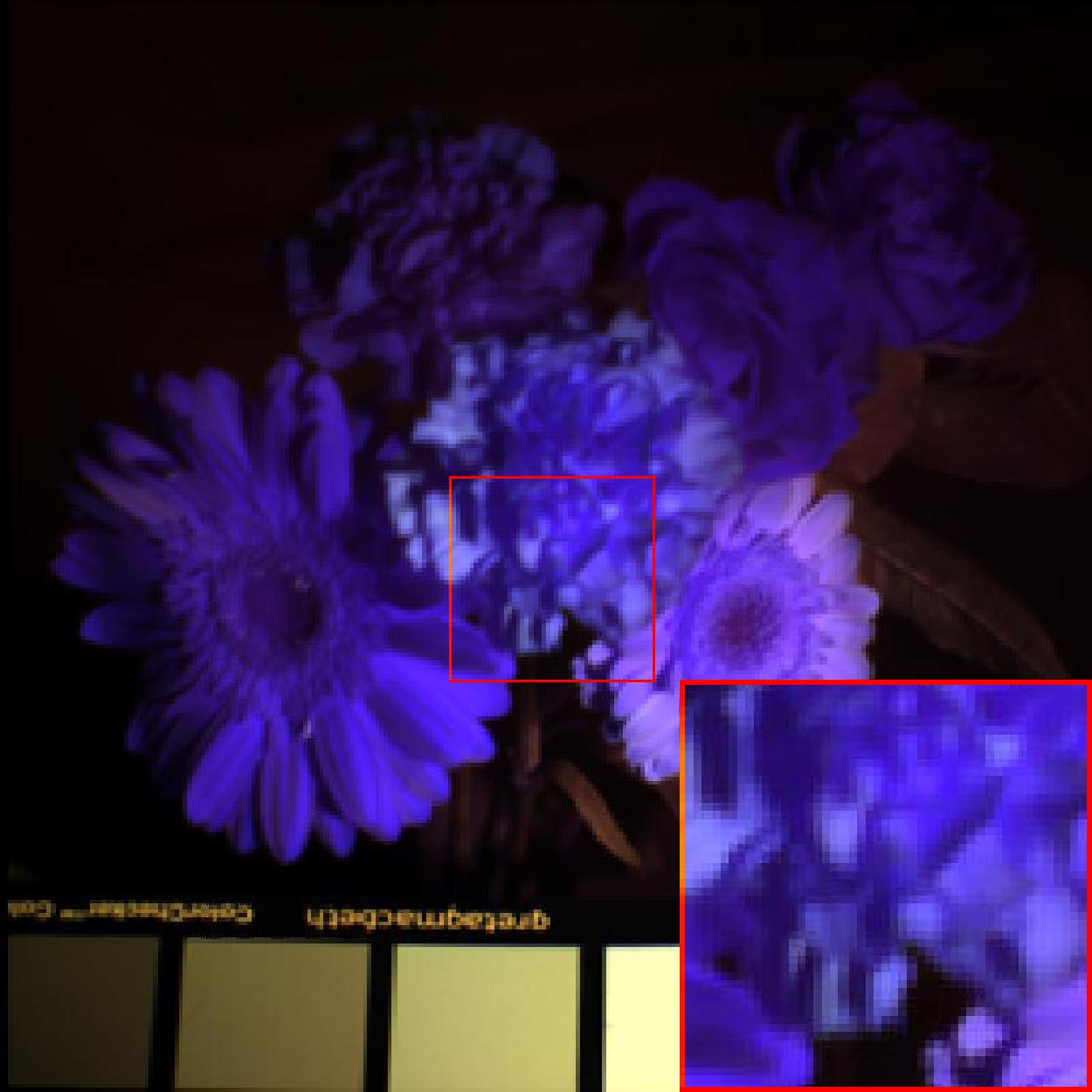}\\

PSNR 14.05 dB & PSNR 26.75 dB & PSNR 24.08 dB & PSNR 30.34 dB & PSNR inf\\

  \end{tabular}
  \caption{First row:  the recovered results by TNN, DPR, and DP3LRTC on  color image  \textit{Airplane} with the sampling rate 10\%. Second row: the recovered results (pseudo-color images composed of the 1st, 2nd, and 31-th bands) by TNN, DPR, and DP3LRTC on  MSI \textit{Flowers} with the sampling rate 5\%.}
  \label{fig:dpr}
\end{figure}

To evaluate the distinct contributions of the low-rankness prior and deep image prior, we test the performance of 3 methods: TNN (only the TNN regularizer), DPR (only the implicit deep  regularizer), and DP3LRTC (both the TNN and implicit deep  regularizers). In Fig.\thinspace\ref{fig:dpr}, we exhibit the recovered results by TNN, DPR, and DP3LRTC on color image  \textit{Airplane} and MSI \textit{Flowers}, together with PSNR values. In Fig.\thinspace\ref{fig:dpr}, the results by DPR are clearer than those by TNN. This shows the  remarkable ability of CNN denoiser to capture the spatial structures.
Meanwhile, we can observe that the performance of DPR is comparable to DP3LRTC on color images, since that color images only consists of 3 color channels. However, from the enlarged areas, we can see obvious artifact in the result by DPR. When reconstructing the MSI, which consists of 31 spectral bands, TNN and DP3LRTC outperform DPR. This shows that the TNN regularizer is good at utilizing the spectral redundancy.
Therefore, we can confirm that two regularizers are indispensable to DP3LRTC and  benefit from each other. Together, they  contribute to
the superior performance of the proposed DP3LRTC.

To take a little more in-depth look at the mechanism of each term, we conduct experiments on all 8 MSIs in Sec.\thinspace\ref{subsec:msic} with different sampling rates. The averaged PSNR and SSIM values are reported in Tab.\thinspace\ref{tab:msic}. Generally, the metric SSIM evaluates the local structure similarity \cite{wang2004image} and the metric PSNR reflects the overall recovery quality. The PSNR values of results by TNN are higher than those by DPR, while DPR achieves higher SSIM values. This phenomenon also reveals that the TNN regularizer is more suitable for characterising the global information and the CNN denoiser is appropriate for recovering the fine structures. As the sampling rate increases, the SSIM margin between TNN and DP3LRTC decreases from 0.140 to 0.036 while the PSNR margin is always around 4dB. This shows that in the low sampling cases, recovering the local structure becomes more difficult than capturing the global information.
\begin{table*}[htbp]
\small\renewcommand\arraystretch{0.8}
\setlength{\tabcolsep}{6pt}
\centering
\caption{Quantitative comparison of the results by different methods on MSIs. The \textbf{best} are highlighted in bold.}
\begin{tabular}{cccccccc}
\toprule
\multicolumn{1}{c}{\multirow{2}[4]{*}{MSI}}&\multicolumn{1}{c}{\multirow{2}[4]{*}{SR}} & \multicolumn{3}{c}{PSNR} & \multicolumn{3}{c}{SSIM} \\
\cmidrule{3-8}\multicolumn{2}{c}{} & TNN   & DPR   & DP3LRTC & TNN   & DPR   & DP3LRTC\\
\midrule
\multirow{3}[2]{*}{Average} & 5\%   & 28.28  & 24.80  & \textbf{32.85} & 0.7735  & 0.8631  & \textbf{0.9138}\\
      & 10\%  & 32.05  & 30.64  & \textbf{36.59} & 0.8720  & 0.9111  & \textbf{0.9558} \\
      & 20\%  & 36.88  & 36.24  & \textbf{40.63} & 0.9443  & 0.9516  & \textbf{0.9805}\\
\bottomrule
\end{tabular}%
\label{tab:msic}
\end{table*}

\begin{figure}[htbp]
\scriptsize
\setlength{\tabcolsep}{5pt}
\centering
\begin{tabular}{cc}
\includegraphics[width=0.4\textwidth]{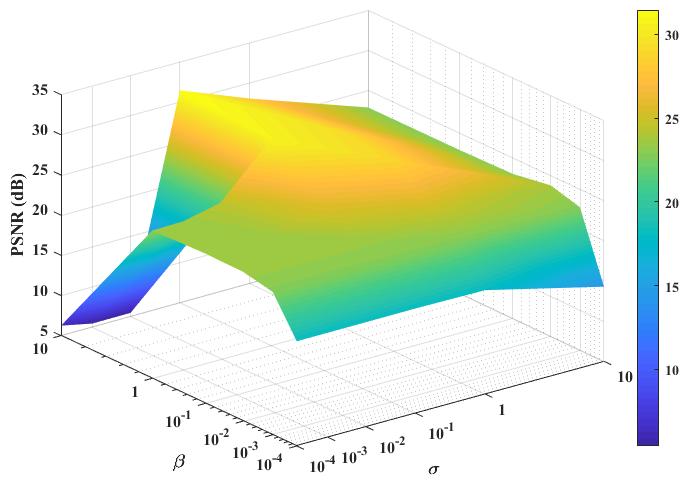}&\includegraphics[width=0.4\textwidth]{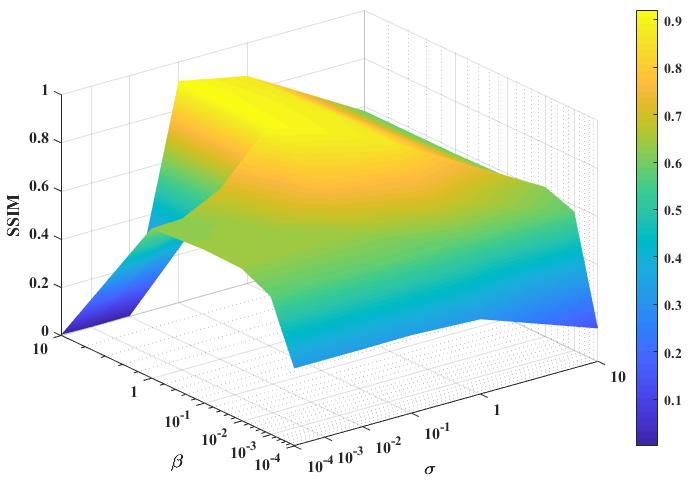}

  \end{tabular}
  \caption{The PSNR and SSIM values of the results recovered by DP3LRTC with respect to $\beta$ and $\sigma$ on color image \textit{Starfish} with the sampling rate $20\%$.}
  \label{fig:par}
\end{figure}

Moreover, we study the influence of  parameters $\beta$ and $\sigma$.
In Fig.\thinspace\ref{fig:par}, we show the PSNR and SSIM values of the results recovered by DP3LRTC on color image \textit{Starfish} with respect to    $\beta$ and $\sigma$. We can see that DP3LRTC delivers the best performance when $\beta \in \{10^{-1}, 1\}$ and $\sigma \in \{10^{-2}, 10^{-1}, 1\}$.

\subsubsection{Low-matrix-rank and Low-tensor-rank}
To illustrate the superiority of the tensor-rank-based model, we reported the quantitative comparison of the recovered results by our DP3LRTC and DP3LRMC, in which the TNN is replaced by the matrix nuclear norm of the matricization of multi-dimensional images (where the rows represent the vectorized spatial images and the columns the spectral channels), on color images and multi-spectral images (MSIs) in Tabs.\thinspace\ref{tab:mnn1}.
The setup of experiments on color images and the multispectral images are the same as Sec. \ref{subsec:cic} and Sec. \ref{subsec:msic}, respectively. Similarly, we only report the average indexes of the results.
For better visual inspection, we further displayed the recovered results by DP3LRMC and DP3LRTC in Fig.\thinspace\ref{fig:msi2}.
From   Tab.\thinspace\ref{tab:mnn1} and Fig.\thinspace\ref{fig:msi2}, we confirmed that the tensor-rank-based model DP3LRTC is evidently better than the matrix-rank-based model DP3LRMC.
\begin{table}[htbp]
\small
\setlength{\tabcolsep}{5pt}
\renewcommand\arraystretch{0.8}
\caption{Quantitative comparison of the results by different methods on color images. The \textbf{best} are highlighted in bold.}
\centering
%\begin{tabular}{cc}
\begin{tabular}[t]{cccccc}
\toprule
 &{\multirow{2}[4]{*}{SR}} & \multicolumn{2}{c}{PSNR} & \multicolumn{2}{c}{SSIM}\\
\cmidrule{3-6}\multicolumn{2}{c}{} & DP3LRMC & DP3LRTC & DP3LRMC & DP3LRTC \\
\midrule
\multirow{3}[2]{*}{Color images (average)} & 10\%  & 23.89  & \textbf{28.83} & 0.7630  & \textbf{0.9061}\\
      & 20\%  & 26.76  & \textbf{31.70} & 0.8719  & \textbf{0.9415} \\
      & 30\%  & 29.07  & \textbf{33.40} & 0.9308  & \textbf{0.9668} \\\midrule
\multirow{3}[2]{*}{MSIs (average)} & 5\%   & 25.76  & \textbf{32.85} & 0.6948  & \textbf{0.9138} \\
      & 10\%  & 28.75  & \textbf{36.59} & 0.7863  & \textbf{0.9558} \\
      & 20\%  & 32.27  & \textbf{40.63} & 0.8706  & \textbf{0.9805} \\
\bottomrule
\end{tabular}
\label{tab:mnn1}
\end{table}

\begin{figure}[h]
\small
\setlength{\tabcolsep}{1pt}
\centering
\begin{tabular}{ccccc}
Observed & DP3LRMC& DP3LRTC & Ground truth\\

\includegraphics[width=0.16\textwidth]{figs/airplane-10-b.jpg}&
\includegraphics[width=0.16\textwidth]{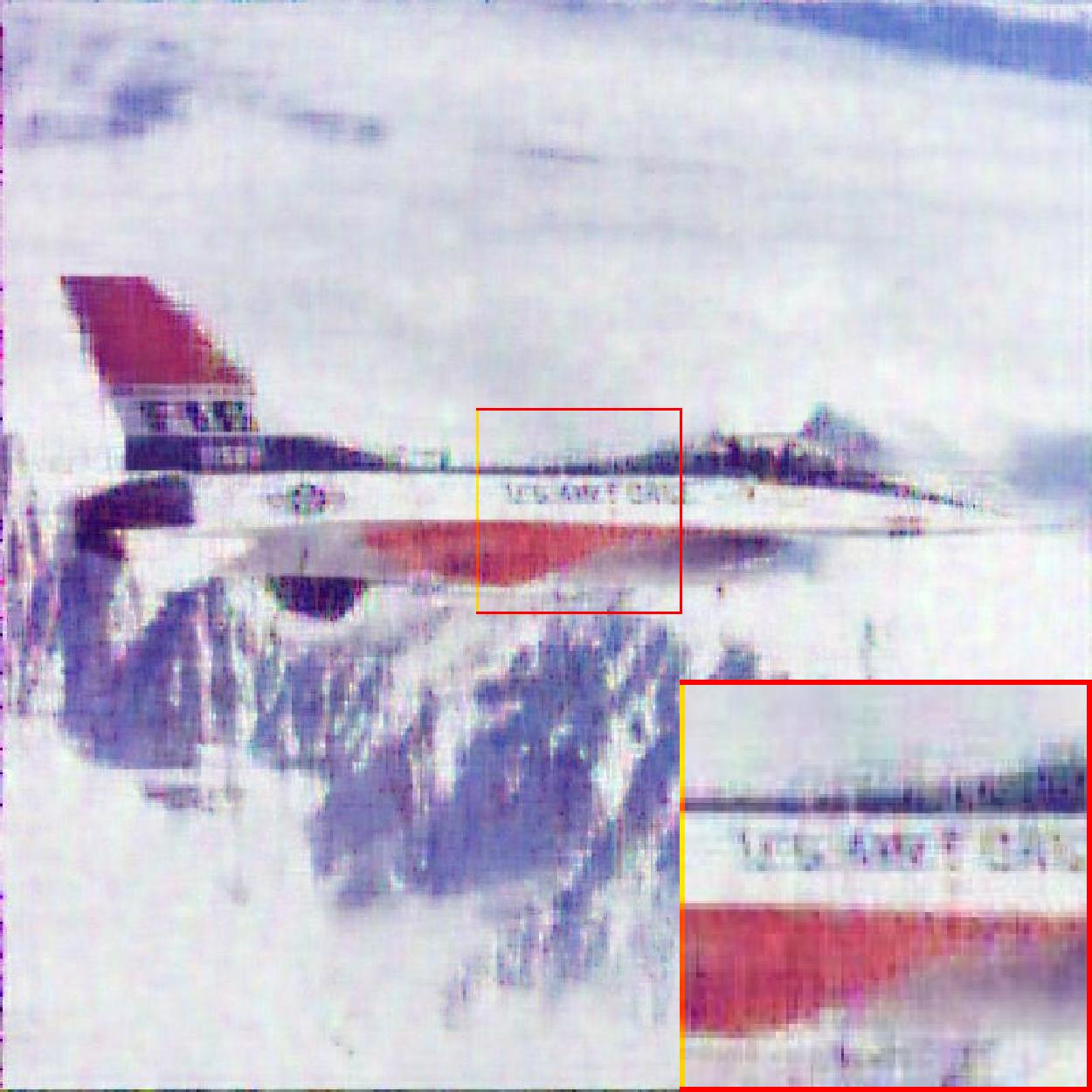}&
\includegraphics[width=0.16\textwidth]{figs/airplane-DP3-10-b.jpg}&
\includegraphics[width=0.16\textwidth]{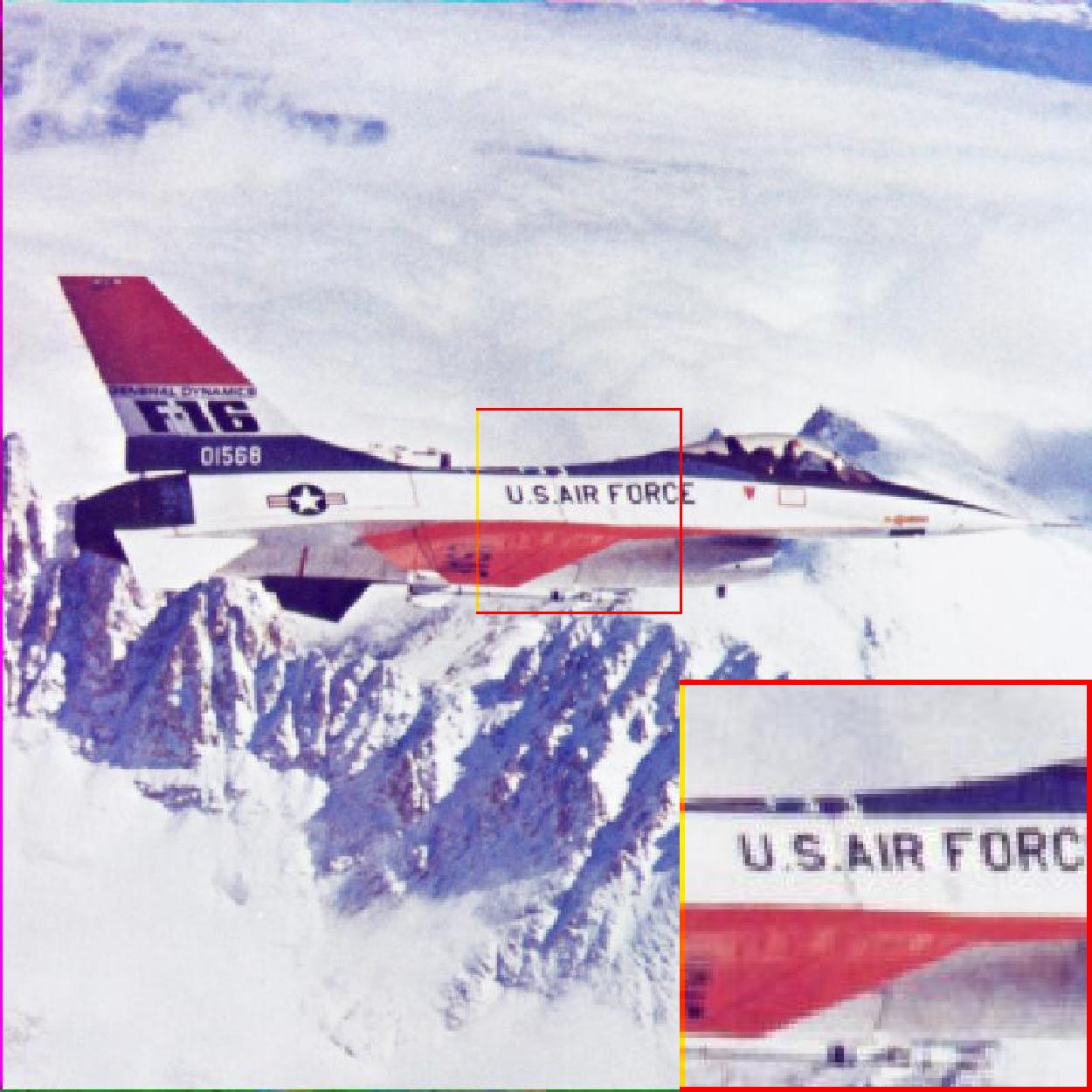}\\

PSNR 2.27 dB & PSNR 24.24 dB & PSNR 28.48 dB & PSNR inf \\

\includegraphics[width=0.16\textwidth]{figs/flowers-5-b.jpg}&
\includegraphics[width=0.16\textwidth]{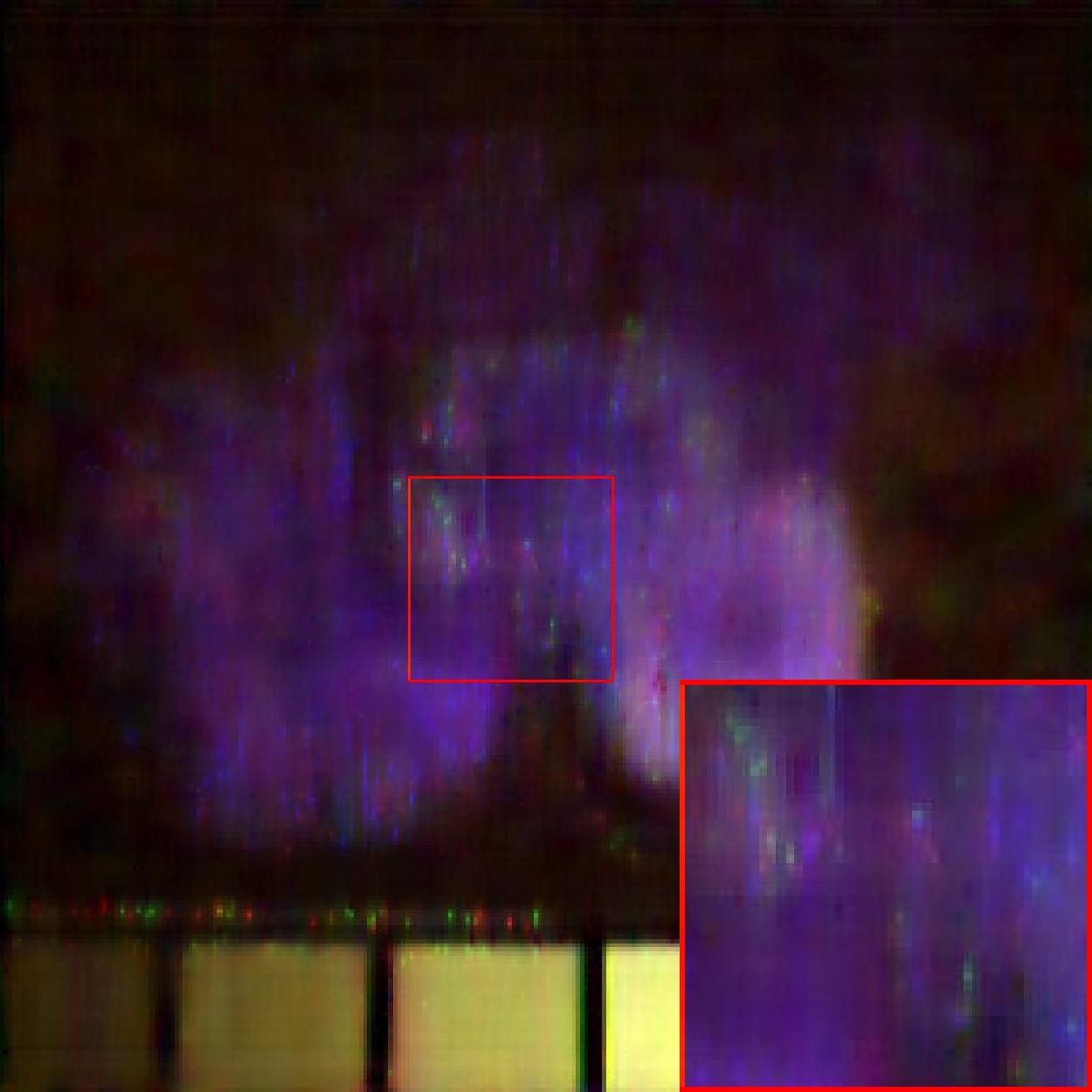}&
\includegraphics[width=0.16\textwidth]{figs/flowers-DP3-5-b.jpg}&
\includegraphics[width=0.16\textwidth]{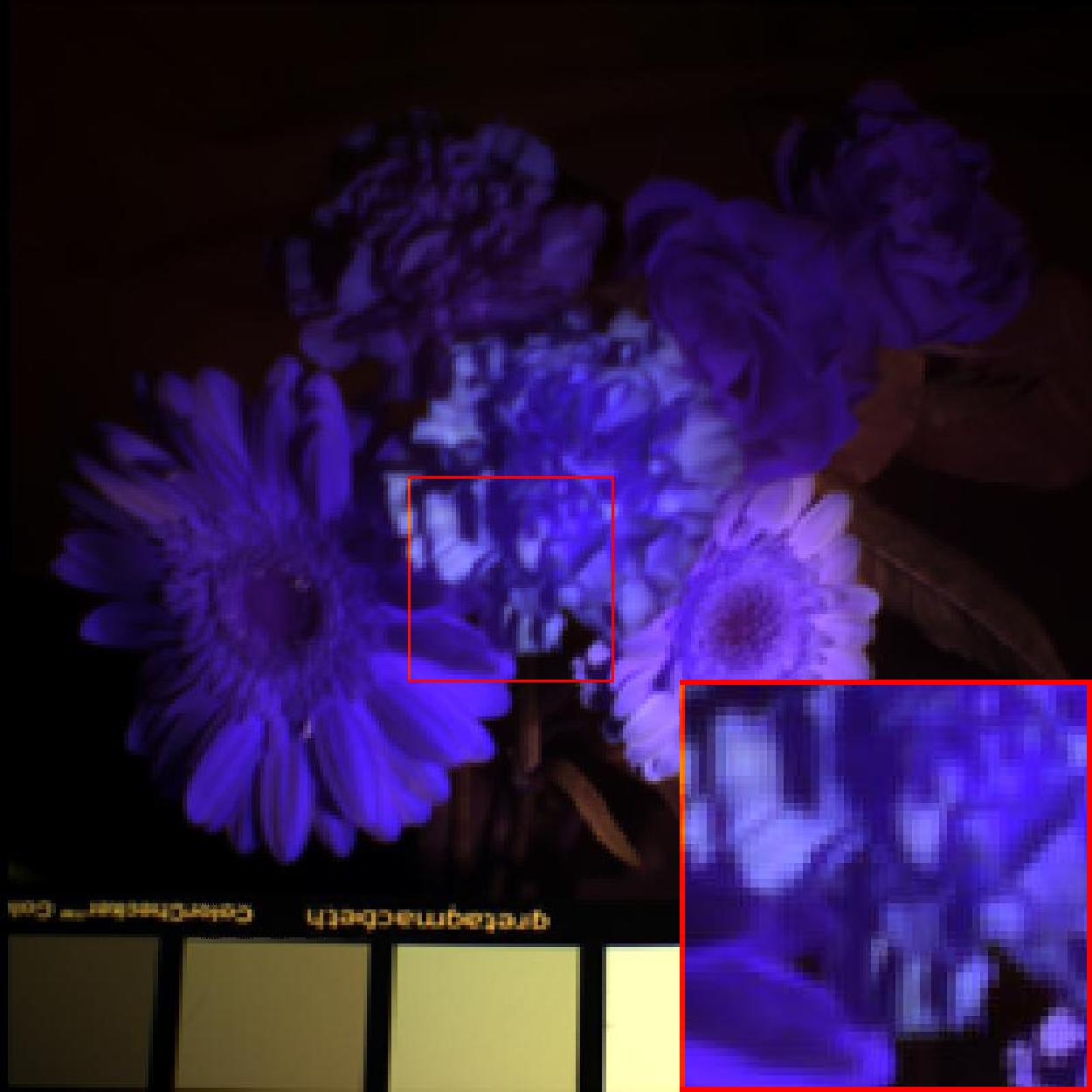}\\

PSNR 14.05 dB & PSNR 24.54 dB & PSNR 30.34 dB & PSNR inf\\

  \end{tabular}
  \caption{First row:  the recovered results  by DP3LRMC and DP3LRTC on color image  \textit{Airplane} with the sampling rate 10\%. Second row: the recovered results (pseudo-color images composed of the 1st, 2nd, and 31-th bands) by DP3LRMC and DP3LRTC on MSI \textit{Flowers} with the sampling rate 5\%.}\vspace{-0.2cm}
  \label{fig:msi2}
\end{figure}
\subsubsection{Beyond TNN, FFDNet, and Third-order Tensors}\label{subsec:btt}
Actually, in this work, we suggested a user-friendly LRTC framework which integrates low-rankness prior and the denoising prior.
Tab.\thinspace\ref{tab:imgc} shows the quantitative comparison of the results obtained by different combinations of low-rankness metrics and denoisers. The setup is the same as the color image completion with random missing pixels in Sec. \ref{subsec:cic}. For saving  space, we only report the average indexes of the results of 8 color images. In Tab.\thinspace\ref{tab:imgc}, ``SNN-DPR''  denotes the combination of the sum of the nuclear norm (SNN) and the  FFDNet denoiser, while ``TNN-BM3D'' indicates that the tensor nuclear norm (TNN) is combined with the BM3D denoiser \cite{dabov2007video}.
From  Tab.\thinspace\ref{tab:imgc}, we can observe  that the combination of the tensor nuclear norm
and the FFDNet denoiser  is best among all combinations.  Thus, we choose the  the combination of the tensor nuclear norm
and the FFDNet denoiser    as the representative example of DP3LRTC framework.

\begin{table}[htbp]
\small
\setlength{\tabcolsep}{8pt}
\renewcommand\arraystretch{0.8}
\caption{Quantitative comparison of the recovered results by different methods on color images. The \textbf{best} are highlighted in bold.}
\centering
\begin{tabular}{cccccccc}
\toprule
\multirow{2}[4]{*}{Color image} &{\multirow{2}[4]{*}{SR}}& \multicolumn{3}{c}{PSNR} & \multicolumn{3}{c}{SSIM}\\
\cmidrule{3-8}&& SNN-DPR &  TNN-BM3D & DP3LRTC & SNN-DPR &  TNN-BM3D & DP3LRTC\\
\midrule
\multirow{3}[2]{*}{Average} & 10\%  & 28.55  & 27.80  & \textbf{28.83} & 0.9013  & 0.8754  & \textbf{0.9061} \bigstrut[t]\\
      & 20\%  & 31.24  & 30.91  & \textbf{31.70} & \textbf{0.9461} & 0.9382  & 0.9415  \\
      & 30\%  & 32.59  & 33.25  & \textbf{33.40} & 0.9597  & 0.9641  & \textbf{0.9668}\\
\bottomrule
\end{tabular}%
\label{tab:imgc}
\end{table}

{Next, we  extend our DP3LRTC framework to high-order tensors  ($n\geq3$) by replacing the TNN regularizer with its high-order extension, i.e., the WSTNN in  \cite{zheng2018tensor}. We conduct the experiment on a fourth-order tensor, the color video ``Suzie'' \footnote{Available at \url{http://trace.eas.asu.edu/yuv/}} of size $144\times176\times3\times100$.  For color videos, we use the FFDNet network trained on color images as the PnP denoiser. When solving the $\mathcal{Z}$ subproblem in \eqref{equ:zpro2}, we feed each frame of size $144\times176\times3$ into the CNN denoiser in sequence. The quantitative comparison of the recovered results by  SNN, WSTNN, and DP3LRTC on color video \textit{Suzie} is reported in Tab.\thinspace\ref{tab:cv}.
For better visual inspection, we further displayed the recovered results by SNN, WSTNN, and DP3LRTC in Fig.\thinspace\ref{fig:cv}.
We can observe that DP3LRTC can be extended to handle high-order tensors with the help of WSTNN. Also, we can generalize our method for high-order tensor via other low-rankness metrics, such as the ``SNN-DPR'', and obtain remarkable performances.

\begin{table}[htbp]
\small
\setlength{\tabcolsep}{4pt}
\renewcommand\arraystretch{0.8}
\caption{Quantitative comparison of the results by different methods on color video. The \textbf{best} are highlighted in bold.}
\centering
\begin{tabular}{cccccccc}
\toprule
\multirow{2}[4]{*}{Color Video}&{\multirow{2}[4]{*}{SR}}& \multicolumn{3}{c}{PSNR} & \multicolumn{3}{c}{SSIM}\\
\cmidrule{3-8}\multicolumn{2}{c}{} & SNN   & WSTNN & DP3LRTC & SNN   & WSTNN & DP3LRTC\\
\midrule
\multirow{3}[1]{*}{Suzie} & 1\%   & 7.44  & 23.26 & \textbf{24.96} & 0.0372  & 0.6967  & \textbf{0.7371}\\
      & 5\%   & 17.97 & 28.24 & \textbf{30.73} & 0.5148  & 0.8384  & \textbf{0.8891} \\
      & 10\%  & 21.22 & 30.94 & \textbf{33.47} & 0.6139  & 0.8943  & \textbf{0.9285}\\
\bottomrule
\end{tabular}%
\label{tab:cv}
\end{table}

\begin{figure*}[h]
\small
\setlength{\tabcolsep}{0.9pt}
\centering
\begin{tabular}{cccccc}
Observed & SNN & TNN & DP3LRTC & Ground truth\\

\includegraphics[width=0.18\textwidth]{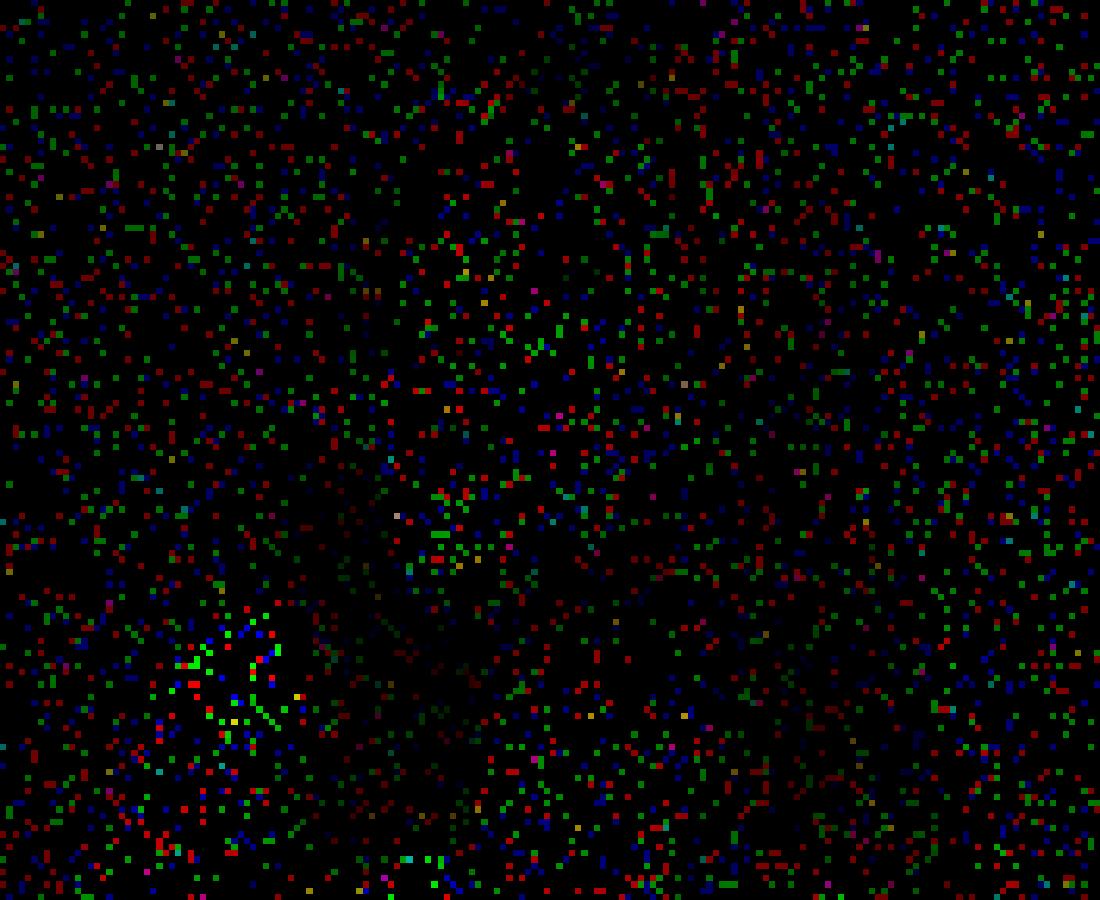}&
\includegraphics[width=0.18\textwidth]{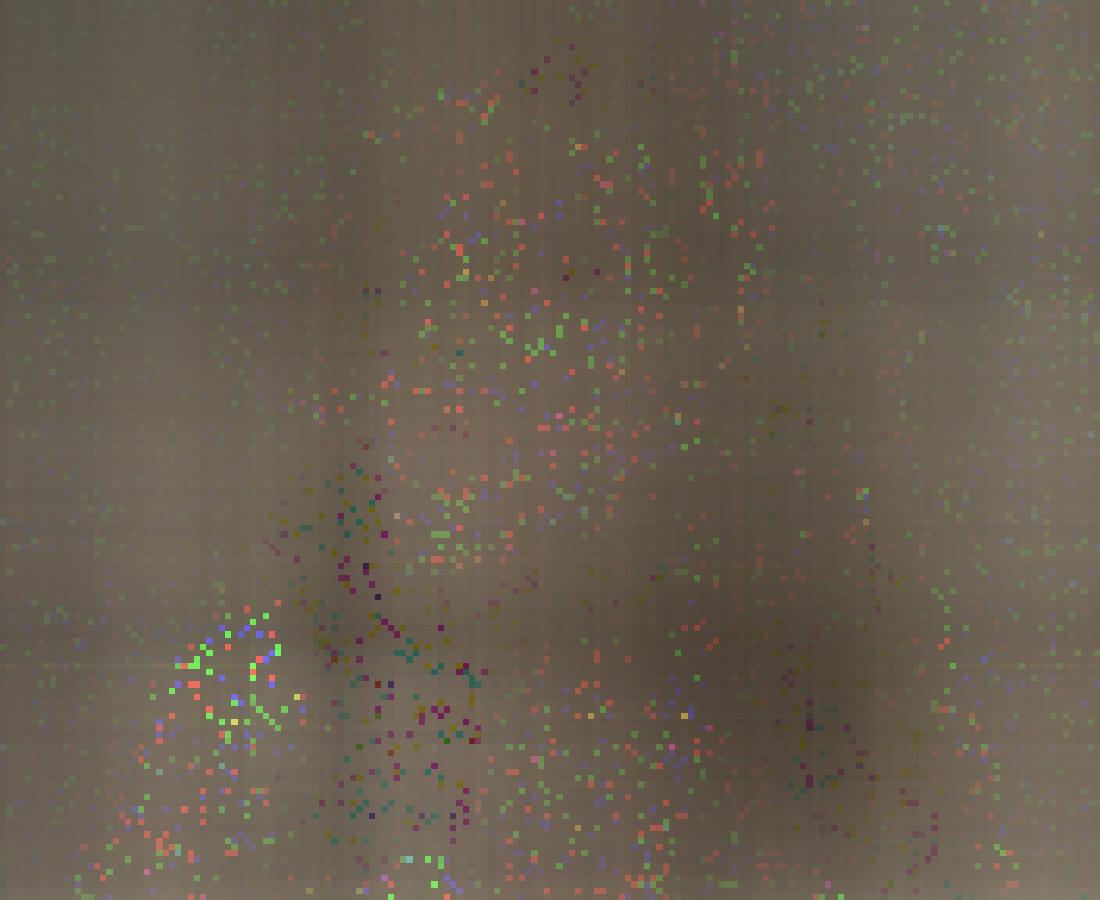}&
\includegraphics[width=0.18\textwidth]{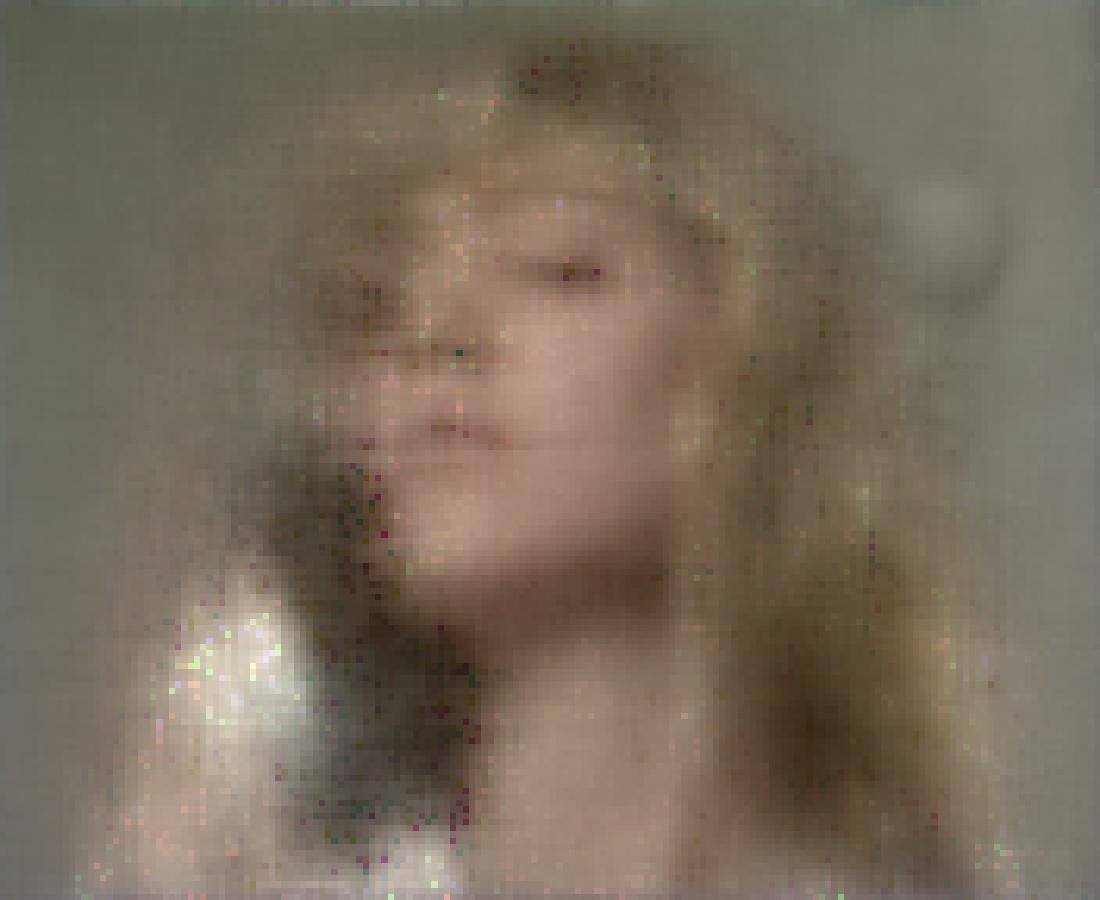}&
\includegraphics[width=0.18\textwidth]{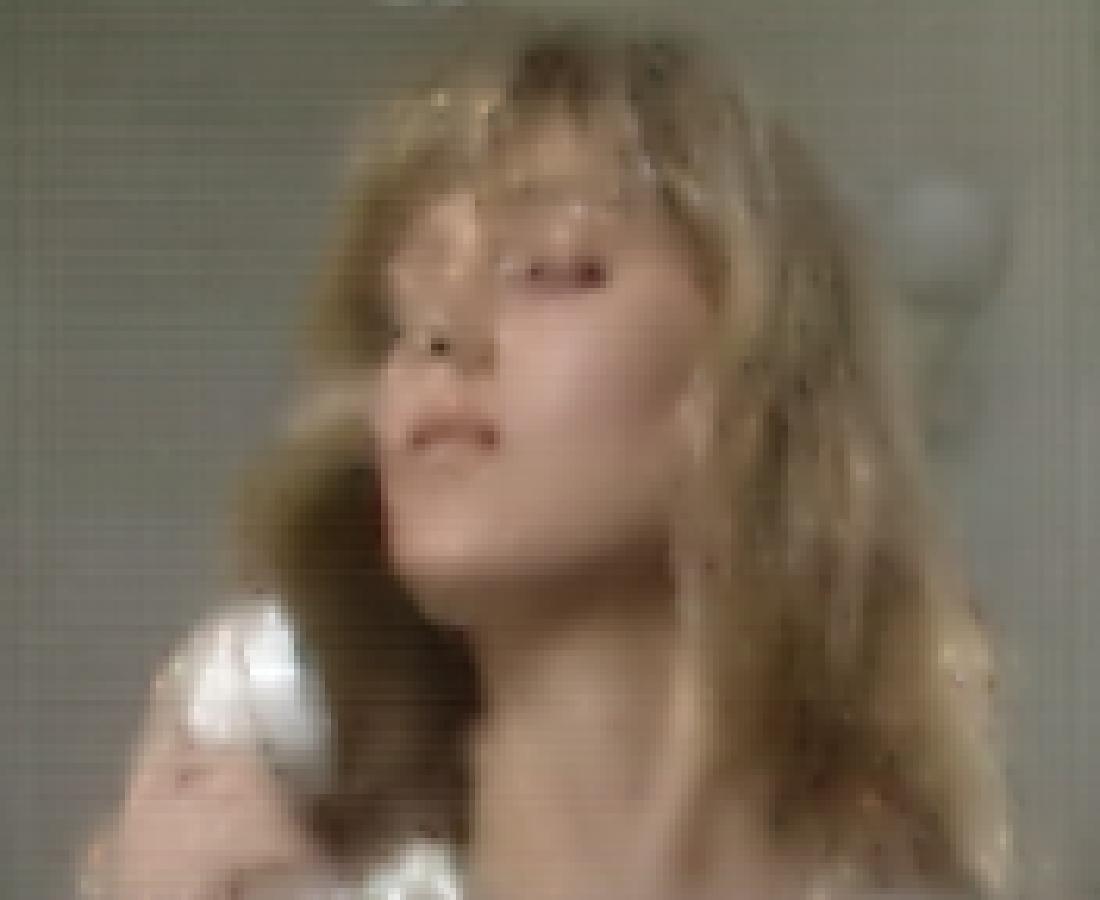}&
\includegraphics[width=0.18\textwidth]{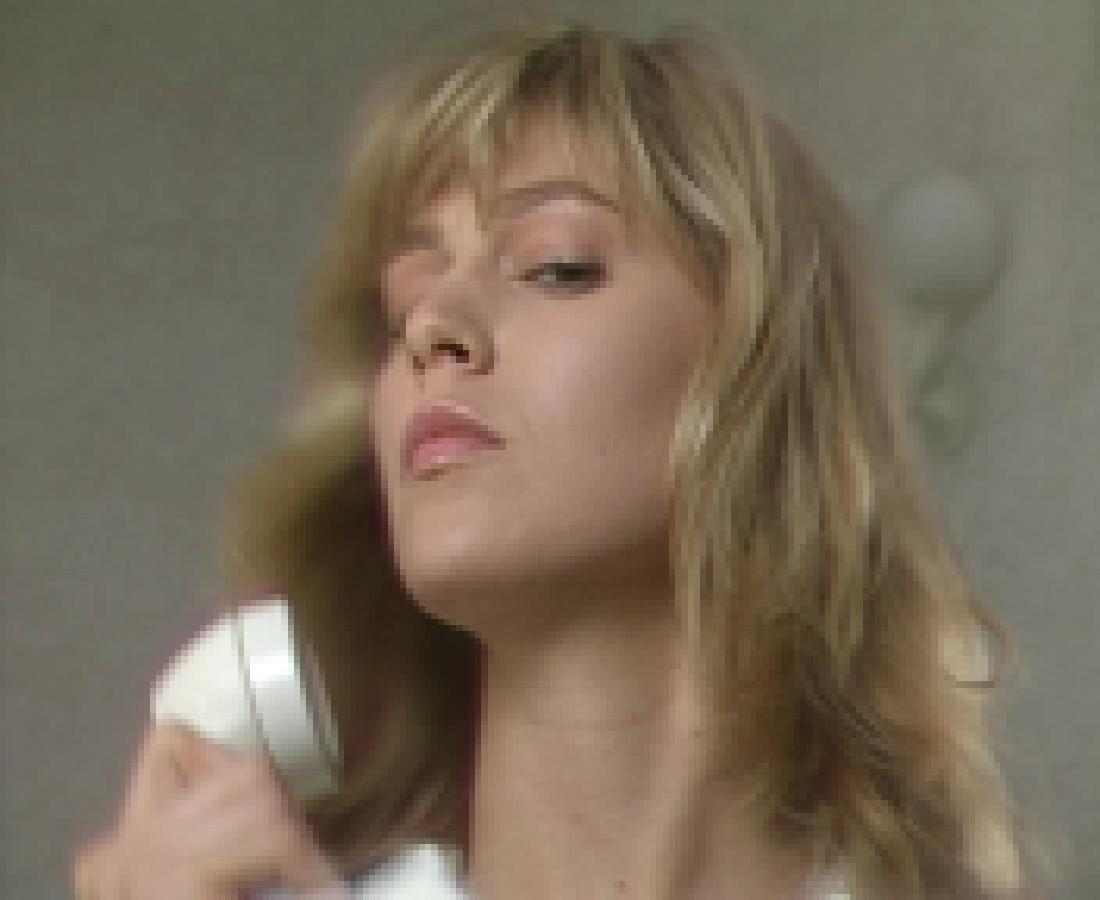}\\
  \end{tabular}
  \caption{The 50-th frame of the recovered color video \textit{Suzie} by SNN, TNN, and DP3LRTC with the sampling rate 5\%.}
  \label{fig:cv}
\end{figure*}

\section{Conclusions} \label{sec:Con}
In this paper, we proposed a hybrid tensor completion model, in which the TNN regularizer is utilized to catch the global information and a data-driven implicit regularizer is used to express the local information. The proposed model simultaneously combines the model-based optimization method with the CNN-based method, in consideration of the global  structure and fine detail preservation. An efficient ADMM is developed to solve the proposed model. Numerical experiments on different types of multi-dimensional images illustrate the superiority of the proposed method on the tensor completion problem.
\section*{References}
{\small
\bibliography{mybibfile}}

\end{document}